\documentclass[journal]{IEEEtran}
\usepackage{amsmath,amsfonts}
\usepackage{algorithmic}
\usepackage{algorithm}
\usepackage{array}
\usepackage[caption=false,font=normalsize,labelfont=sf,textfont=sf]{subfig}
\usepackage{textcomp}
\usepackage{stfloats}
\usepackage{url}
\usepackage{verbatim}
\usepackage{graphicx}
\usepackage{cite}
\usepackage{hyperref}
\usepackage{amsmath,amsfonts,amssymb,amsthm,bbm}
\usepackage{commath}
\usepackage{mathtools}
\usepackage{xargs}
\usepackage{xcolor}  
\usepackage[table]{colortbl}
\usepackage{multirow}
\usepackage{booktabs}
\usepackage{arydshln}
\usepackage[percent]{overpic}
\usepackage{pgfplots}
\pgfplotsset{width=10cm,compat=1.9}
\usepackage{tikz}
\usepackage{tikzscale}
\usetikzlibrary{shapes,arrows,positioning,spy,calc}
\usepackage{wrapfig}
\usepackage[export]{adjustbox}
\usepackage[colorinlistoftodos,prependcaption,textsize=tiny]{todonotes}
\usepackage{accents}

\newcommandx{\Ted}[2][1=]{\todo[inline,linecolor=blue,backgroundcolor=blue!25,bordercolor=blue,#1]{#2}}
\newcommandx{\Joao}[2][1=]{\todo[inline,linecolor=green,backgroundcolor=red!25,bordercolor=red,#1]{#2}}
\newcommandx{\Sen}[2][1=]{\todo[inline,linecolor=red,backgroundcolor=red!25,bordercolor=red,#1]{#2}}

\begin{document}

\newcommand{\RefSec}{Section }
\newcommand{\RefFig}{Fig. }
\newcommand{\RefFigs}{Figs. }
\newcommand{\RefTab}{Table }
\newcommand{\RefAlg}{Algorithm }

\title{Estimating Fog Parameters from a Sequence of Stereo Images}

\author{Yining Ding, João F. C. Mota, Andrew M. Wallace, and Sen Wang\textsuperscript{*}
\thanks{Y. Ding is with the Edinburgh Centre for Robotics, the School of Mathematical and Computer Sciences, Heriot-Watt University, Edinburgh EH14 4AS, U.K. (e-mail: yd2007@hw.ac.uk).}
\thanks{J. F. C. Mota and A. M. Wallace are with the School of Engineering and Physical Sciences, Heriot-Watt University, Edinburgh EH14 4AS, U.K. (e-mail: \{j.mota, a.m.wallace\}@hw.ac.uk).}
\thanks{S. Wang is with the Sense Robotics Lab, Department of Electrical and Electronic Engineering, Imperial College London, London SW7 2AZ, U.K. (e-mail: sen.wang@imperial.ac.uk).}
\thanks{Manuscript received xx xx, 2024; revised xx xx, 2024.}
\thanks{* Corresponding author}}

\markboth{Journal of \LaTeX\ Class Files,~Vol.~14, No.~8, August~2021}%
{Shell \MakeLowercase{\textit{et al.}}: A Sample Article Using IEEEtran.cls for IEEE Journals}

\IEEEpubid{0000--0000/00\$00.00~\copyright~2021 IEEE}

\maketitle

\begin{abstract}
We propose a method which, given a sequence of stereo foggy images, estimates the parameters of a fog model and updates them dynamically.
In contrast with previous approaches, which estimate the parameters sequentially and thus are prone to error propagation, our algorithm estimates all the parameters simultaneously by solving a novel optimisation problem.
By assuming that fog is only locally homogeneous, our method effectively handles real-world fog, which is often globally inhomogeneous.
The proposed algorithm can be easily used as an add-on module in existing visual Simultaneous Localisation and Mapping (SLAM) or odometry systems in the presence of fog.
In order to assess our method, we also created a new dataset, the Stereo Driving In Real Fog (SDIRF), consisting of high-quality, consecutive stereo frames of real, foggy road scenes under a variety of visibility conditions, totalling over 40 minutes and 34k frames.
As a first-of-its-kind, SDIRF contains the camera's photometric parameters calibrated in a lab environment, which is a prerequisite for correctly applying the atmospheric scattering model to foggy images.
The dataset also includes the counterpart clear data of the same routes recorded in overcast weather, which is useful for companion work in image defogging and depth reconstruction.
We conducted extensive experiments using both synthetic foggy data and real foggy sequences from SDIRF to demonstrate the superiority of the proposed algorithm over prior methods.
Our method not only produces the most accurate estimates on synthetic data, but also adapts better to real fog.
We make our code and SDIRF publicly available\footnote{\url{https://github.com/SenseRoboticsLab/estimating-fog-parameters}} to the community with the aim of advancing the research on visual perception in fog.
\end{abstract}


\section{Introduction}
\IEEEPARstart{F}{og} is formed when small water droplets are suspended in the air.
They interact with light, for example via scattering, causing severe visual degradation, which in turn poses significant challenges to visual perception. Foggy scenarios, despite their low probability of occurrence, are thus important edge cases that cannot be ignored for extremely safety-oriented systems such as autonomous vehicles.
The amount of visual degradation depends on the depth of the corresponding scene point, as explained by the atmospheric scattering model.

\begin{figure}[!t]
    \centering
    \includegraphics[trim={0cm 0.5cm 0cm 0cm},clip,width=\linewidth]{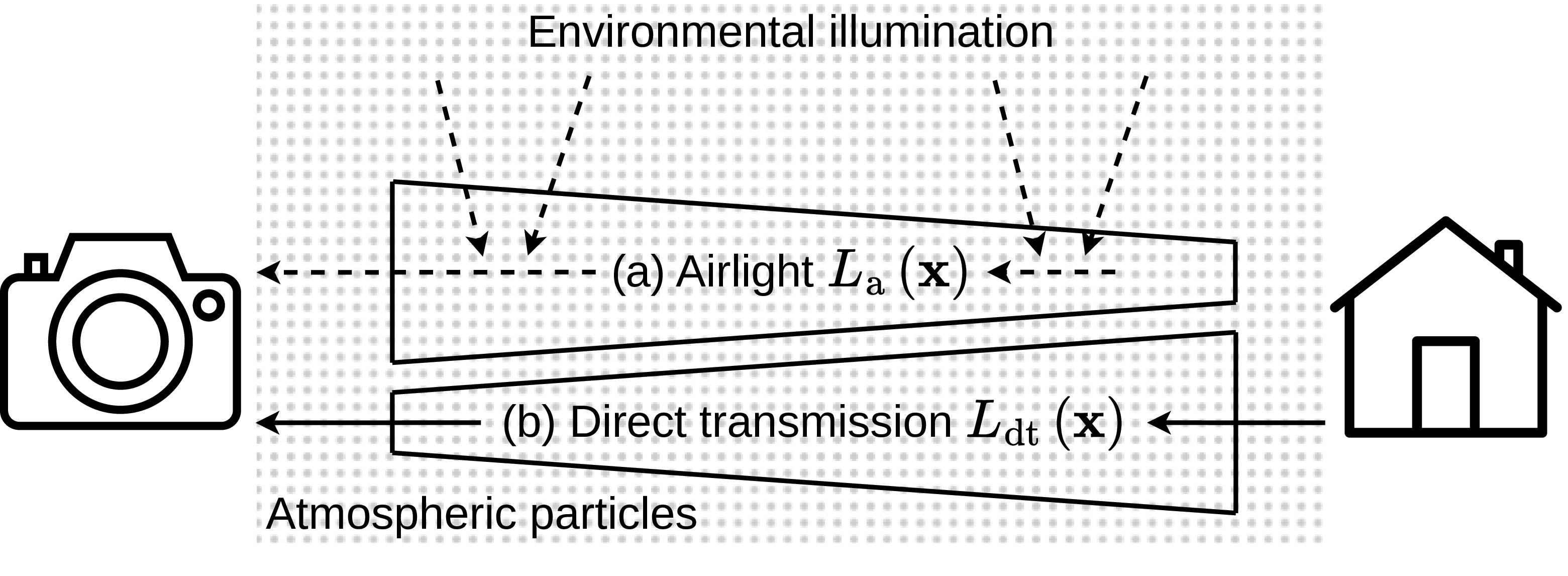}
    \setlength{\abovecaptionskip}{-13pt}
    \caption{The atmospheric scattering model.
    (a) Airlight: The atmospheric particles act as a light source by reflecting environmental illumination towards the camera, causing the radiance of the airlight $ L_{\text{a}} \left( \mathbf{x} \right) $ to increase with distance.
    (b) Direct transmission: The atmospheric particles also scatter away the incident light that traverses from a scene point to the camera, causing the radiance of direct transmission $ L_{\text{dt}} \left( \mathbf{x} \right) $ to attenuate with distance.
    }
    \setlength{\belowcaptionskip}{-13pt}
    \label{fig:atmospheric_scattering}
\end{figure}

\IEEEpubidadjcol
\textbf{Atmospheric scattering model.}
\RefFig \ref{fig:atmospheric_scattering} illustrates the atmospheric scattering model \cite{narasimhan2002vision}, which decomposes the total radiance originating on a scene point and reaching the camera under foggy conditions into direct transmission and airlight.
The quantity of light transmitted via direct transmission (resp.\ airlight) decreases (resp.\ increases) with the distance from the scene point to the camera.
Formally, let $\mathbf{x} \in \mathbb{Z}_{+}^{2}$ denote the pixel coordinates (in the image plane of the camera) associated with the scene point. Then, the total radiance $ L\left( \mathbf{x} \right) \in \mathbb{R}_{+}$ reaching that point can be decomposed as
\begin{equation}    \label{eq:asm_radiance}
    L\left( \mathbf{x} \right)
    = L_{\text{dt}} \left( \mathbf{x} \right) + L_{\text{a}} \left( \mathbf{x} \right)
    = L_{\text{c}}\left( \mathbf{x} \right) t\left( \mathbf{x} \right) + L_{\infty}\left( 1 - t\left( \mathbf{x} \right) \right)  \text{,}
\end{equation}
where $L_{\text{dt}} \left( \mathbf{x} \right) \in \mathbb{R}_{+}$ is the direct transmission,
$L_{\text{a}} \left( \mathbf{x} \right) \in \mathbb{R}_{+}$ is the airlight,
$L_{\text{c}} \left( \mathbf{x} \right) \in \mathbb{R}_{+}$ is the fog-free radiance of the scene point,
$L_{\infty} \in \mathbb{R}_{+}$ is the radiance of the atmospheric light, i.e., the airlight at infinite distance,
and $t \left( \mathbf{x} \right) \in \left( 0,1 \right)$ is the transmission coefficient, which controls the combination between $L_{\text{c}} \left( \mathbf{x} \right)$ and $L_{\infty}$ as a function of the distance $d\left( \mathbf{x} \right) \in \mathbb{R}_{++}$ between the scene point and the camera:
\begin{equation}    \label{eq:trans}
    t\left( \mathbf{x} \right) = \exp{ \left( -\beta d\left( \mathbf{x} \right) \right) }   \text{.}
\end{equation}
The parameter $\beta \in \mathbb{R}_{++}$ is the scattering coefficient and measures the density of fog, being related to visibility $V_{\text{MOR}} \in \mathbb{R}_{++}$ (also known as the meteorological optical range \cite{mor}) as
\begin{equation}    \label{eq:vis}
    V_{\text{MOR}} = -\ln \left(0.05\right) / \beta     \text{,}
\end{equation}
where we assume that $V_{\text{MOR}}$ is measured in meters.

For simplicity, in the rest of this paper we will omit $ \mathbf{x} $ from any variables, e.g., in \eqref{eq:asm_radiance} and \eqref{eq:trans}, whenever their dependence on the pixel coordinates is clear from context.

Note that we limit the scope of this work to daytime fog, mist or haze.
In other inclement weather conditions, such as rain and snow, where the particles present in the atmosphere typically demonstrate significant spatial and temporal inhomogeneity, the above atmospheric model no longer applies \cite{narasimhan2002vision}. 

\textbf{Fog parameters.}
The fog parameters are $L_{\infty}$ and $\beta$, and their knowledge is key both to defog the input images and to construct an accurate depth map of the scene.
Estimating the fog parameters accurately is thus crucial for improving the safety of autonomous vehicles and mobile robots operating in challenging weather.
Specifically, estimating $L_{\infty}$ is an important step in most non-deep learning-based image defogging methods such as \cite{he2010single, berman2016non}.
$\beta$ is also an essential parameter because, as \eqref{eq:trans} suggests, it determines the relation between the transmission coefficient $t$ and the distance $d$ (from which the scene depth can be calculated given the pixel coordinates $\mathbf{x}$ and the camera's intrinsic parameters).
Consequently, an accurate estimate of $\beta$ is a prerequisite for simultaneous defogging and stereo reconstruction methods \cite{caraffa2012stereo, li2015simultaneous, ding2022variational}.

Prior work on single image defogging \cite{narasimhan2002vision, narasimhan2003contrast, tan2008visibility, he2010single, berman2016non} typically assumes that $\beta$ is constant (i.e., a homogeneous medium) over horizontal paths.
In our work, leveraging a sequence of images, we adapt this assumption to local homogeneity.
That is, we assume that the fog is homogeneous only within a local space, corresponding to the local map constructed in the first step of our method (\RefSec \ref{subsec:representation_of_a_local_map}).
Experimental results show that our method effectively handles real-world fog, even when the fog is inhomogeneous over larger areas, thus validating our assumption.

\textbf{Intensity vs radiance in fog parameter estimation.}
Although \eqref{eq:asm_radiance} was derived to study the scattering phenomenon in the field of photometry/radiometry, almost all existing literature on fog parameter estimation or defogging applies it directly to pixel intensity values.
Hence, \eqref{eq:asm_radiance} is rewritten as
\begin{equation}
    I = J t  + A \left( 1-t \right)    \text{,}    \label{eq:asm_intensity}
\end{equation}
where $I \in [0, 255]$ is the observed intensity of the scene point, $J \in [0, 255]$ is the fog-free intensity of the scene point, and $A \in [0, 255]$ is the intensity of the atmospheric light.
These quantities are counterparts of $L$, $L_{\text{c}}$ and $L_{\infty}$ in \eqref{eq:asm_radiance}.
\eqref{eq:asm_radiance} and \eqref{eq:asm_intensity} can be applied to either a grayscale image or each colour channel in an RGB image independently \cite{narasimhan2003contrast}.
However, using \eqref{eq:asm_intensity} rather than \eqref{eq:asm_radiance} implicitly neglects any non-linearities incurred in the mapping from scene radiance to pixel intensity saved in a compressed image format such as JPEG or PNG.
There can exist many sources of non-linearities in an image sensing pipeline, the most prominent one being the gamma correction \cite{szeliski2022computer}.
Nevertheless, to the best of our knowledge, no existing fog parameter estimation method takes gamma correction into account, and no existing foggy dataset for autonomous driving provides the photometric parameters of the camera.
We will demonstrate that estimating $\beta$ based on \eqref{eq:asm_intensity} rather than \eqref{eq:asm_radiance} results in a different optimisation problem (\RefSec \ref{subsubsec:theoretical_analyses}) and introduces a bias (\RefSec \ref{subsubsec:gamma_correction}).

\textbf{Problem statement.}
Given a sequence of stereo images taken by a vehicle driving in fog, our goal is to estimate the parameters $\beta$, $L_{\infty}$ and relevant $L_{\text{c}}$s of the atmospheric scattering model in \eqref{eq:asm_radiance} and \eqref{eq:trans}, and to update them dynamically.

\textbf{Our approach and contributions.}
We seek to find a set of distance-radiance curves, defined by \eqref{eq:asm_radiance} and \eqref{eq:trans} and characterised by the above parameters, that most closely fits the observed data, which is generated from a local 3D feature map built by a visual Simultaneous Localisation and Mapping (SLAM) or odometry system (e.g., \cite{mur2017orb}).
As the ego-vehicle moves in an environment, the local map [\RefFig \ref{fig:method}(a)] is updated, and so are the observations [\RefFig \ref{fig:method}(b)], our regression problem [\RefFig \ref{fig:method}(c)], and our parameter estimation results.
As no existing real foggy dataset meets our needs, we collected our own dataset and are releasing it to the public.

We summarise our contributions as follows.
\begin{itemize}
    \item
    We propose an optimisation-based method which estimates all the fog parameters \emph{simultaneously}.
    Compared to prior approaches, all of which adopt a \emph{sequential} estimation strategy, our method is less sensitive to error propagation.
    The proposed method is purely model-based, with its estimated fog parameters constrained via physical principles.
    Our only assumption is local homogeneity of the fog, a constraint which real fog in general satisfies.

    \item
    We demonstrate through comprehensive experimental results that our method
    a) outperforms competitive methods both quantitatively and qualitatively on simulated data;
    b) achieves the best performance (qualitatively) on real data. Specifically, it distinguishes thin from thick fog better than prior methods and is able to respond adaptively to spatially variant fog. Also, its estimate of atmospheric light is closer to the colour of the horizon.

    \item
    We publish the Stereo Driving In Real Fog (SDIRF) dataset, the first foggy dataset comprising consecutive stereo images of real road scenes under various visibility conditions.
    Our dataset also includes the counterpart clear images of the same routes recorded in overcast weather.
    Additionally, we calibrate the camera's photometric parameters to make SDIRF photometrically ready for the deployment of the atmospheric scattering model.
    
\end{itemize}

This paper is an extension of our previous work \cite{ding2024estimating}, in which only synthetic data was used to evaluate the proposed fog parameter estimation method.
Here, we refine our method by making its initialisation fully automatic, release the new, real SDIRF dataset, and add extensive evaluations on it.

\textbf{Organisation.}
In the next section, we review literature on fog parameter estimation and discuss existing datasets that are publicly available in the field of autonomous driving.
In \RefSec \ref{sec:methodology}, we explain in detail the proposed fog parameter estimation methodology.
In \RefSec \ref{sec:sdirf_dataset}, we introduce our self-collected SDIRF dataset and describe how we carried out the camera's photometric calibration.
In \RefSec \ref{sec:experiments}, we conduct extensive experiments to evaluate our fog parameter estimation method and compare it with its competitors using both synthetic and real data.
We conclude in \RefSec \ref{sec:conclusion}.

\section{Related Work}  \label{sec:related_work}
\subsection{Fog Parameter Estimation}
Almost all existing methods operate at pixel intensity level, i.e., \eqref{eq:asm_intensity}, without knowledge of the photometric parameters, that is, they estimate $A$ rather than $L_{\infty}$ or $\beta$.
Early approaches estimate $A$ from multiple images of the same scene acquired under different conditions, such as visibility \cite{narasimhan2000chromatic} or manually changed polarisation \cite{schechner2001instant}.
Such methods are thus inapplicable in autonomous vehicle or mobile robot scenarios.
In addition, some methods rely on very strong assumptions, such as the presence of a sky region in the image.
In the rest of this section, we focus on existing work that processes a single image, a stereo pair of images or a sequence of images acquired by an onboard camera.

\textbf{Estimation of $\boldsymbol{A}$.}
This is a critical step in non-deep learning-based single image defogging methods.
To this end, \cite{tan2008visibility} obtains $A$ from the pixels that have the highest intensity in the input image, 
whereas \cite{he2010single} relies on the dark channel prior to locate the most haze-opaque region in the image and then computes $A$ from these pixel intensities.
In turn, \cite{chiang2011underwater} estimates $A$ as the brightest pixel value among all local minima, 
and \cite{berman2016non} locates $A$ in RGB space by leveraging the observation that fog transforms the distribution of pixel intensities from tight clusters to stretched lines (dubbed ``haze-lines'').
Given the limited amount of information embedded in a single image, some of these approaches have demonstrated their general effectiveness in estimating $A$ and therefore are adopted by later conventional methods such as \cite{chen2016robust}, and some pioneering deep learning-based methods such as \cite{cai2016dehazenet}.
Even some video defogging methods such as \cite{li2015simultaneous} directly follow \cite{he2010single}'s approach in estimating $A$, due to its robustness and simplicity.
Similarly, \cite{cai2016real} applies firstly \cite{chiang2011underwater}'s method to compute an $A$ value from the current frame. To impose temporal consistency, they then refine their estimate of $A$ by calculating a weighted average of this $A$ value and the $A$ estimate from the previous frame.

\textbf{Estimation of $\boldsymbol{\beta}$.}
Whenever the value of $t$ can be inferred directly from $I$ [cf.\ \eqref{eq:asm_intensity}], most defogging methods (e.g., \cite{he2010single, berman2016non}) bypass the estimation of the parameter $\beta$ in \eqref{eq:trans}.
This topic can be categorised into \emph{perceptual} estimation and \emph{quantitative} estimation.
Methods including \cite{choi2015referenceless, ling2017optimal, guo2022density} achieve referenceless prediction of perceptual fog density from a single image.
Although their predicted perceptual fog density indices may correlate well with human judgements, the authors make no attempt to show how these perceptual indices can be mapped to a \emph{numerical} value of $\beta$.
To the best of our knowledge, the \emph{quantitative} estimation of $\beta$ is hardly addressed in the existing literature.
As \eqref{eq:trans} implies, $\beta$ is the key linkage between the problem of defogging and the problem of scene depth estimation, and consequently an accurate estimate of its value plays a crucial part in various existing simultaneous defogging and stereo reconstruction methods \cite{caraffa2012stereo, li2015simultaneous, ding2022variational}.
In general, estimating $\beta$ entails observing the same object (more precisely, the same $J$) at a range of known distances, which makes this task extremely challenging at best and not always possible when only a single image or even only a stereo pair of images is available.
As a special case, \cite{hautiere2006automatic} estimates $\beta$ from just a single image but requires the image to contain both the sky and the road, the latter assumed homogeneous and flat (so that a known depth can be associated with each image row from the road after calibration).
These are indeed very strong and application-specific constraints, making the method inapplicable to general scenes.
In contrast, \cite{li2015simultaneous} uses a sequence of images and performs structure-from-motion to facilitate observations of the same object at a range of known distances.
After $A$ is estimated following \cite{he2010single}, they use each pair of observations, whose inverse depth difference is large enough, to estimate $\beta$ by inverting the atmospheric scattering model.
Then all the estimates are gathered, from which they build a histogram of $\beta$ and choose the value from the highest bin.

To summarise, estimating the fog parameters from a sequence of images \cite{li2015simultaneous, cai2016real} is more robust compared to using a single image or a stereo pair of images \cite{tan2008visibility, he2010single, chiang2011underwater, berman2016non, hautiere2006automatic}, because more information is available and there are fewer assumptions or constraints to be made.
Nevertheless, existing methods still have a few shortcomings.
\cite{cai2016real} estimates $A$ only and introduces a weighted average scheme to enforce its temporal consistency.
However, as a key factor in controlling such consistency, the weight itself becomes a learnable parameter and requires fine-tuning for overall optimal performance in different scenarios.
As will be shown in \RefSec \ref{sec:experiments}, the strategy for estimating $A$ and $\beta$ proposed in \cite{li2015simultaneous} has severe drawbacks.
Firstly, $A$ is still estimated from a single image (i.e., the current frame), and thus possibly temporally inconsistent.
Secondly, estimating $\beta$ requires a previous estimate of $A$; any error in the latter estimate thus propagates to $\beta$.
See our supplementary material for a theoretical qualitative analysis of the error propagation.

Distinct from the existing methods that estimate the fog parameters sequentially, we propose an optimisation-based method that simultaneously estimates them.
It assumes only local homogeneity of the fog, which is very realistic.

\subsection{Foggy Datasets for Autonomous Driving}
Publicly available datasets have vastly aided the research and development of perception algorithms for autonomous vehicles.
The overwhelming majority of them \cite{geiger2012we, cordts2016cityscapes, maddern20171, barnes2020oxford, sun2020scalability, caesar2020nuscenes, mao2021one}, however, do not contain any foggy scene, the presence of which can pose significant challenges to a driver-assistance system.

\textbf{Real fog.}
Real fog happens rarely.
Only a few existing datasets include real foggy scenes and they are listed below.
DrivingStereo \cite{yang2019drivingstereo} contains four stereo sequences that are labelled ``foggy''. Nevertheless, in all of them the visibility is still relatively good.
BDD100K \cite{yu2020bdd100k} uses a monocular camera, from which the depth of the scene can be recovered only up to a scale using the pinhole camera model.
SeeingThroughFog \cite{bijelic2020seeing} features a number of stereo foggy sequences. However, the consecutive frames of only its left camera are published.
RADIATE \cite{sheeny2021radiate} pays particular attention to radar imaging in adverse weather. It has four stereo foggy sequences, only one of which was recorded while the ego-vehicle was on the move. Unfortunately, in that sequence there is consistently a considerable amount of water residual on the camera casing, which significantly blocks the view.

\begin{figure*}[!t]
    \centering
    \includegraphics[trim={0.05cm 0.05cm 0.30cm 7.00cm},clip,width=.98\linewidth]{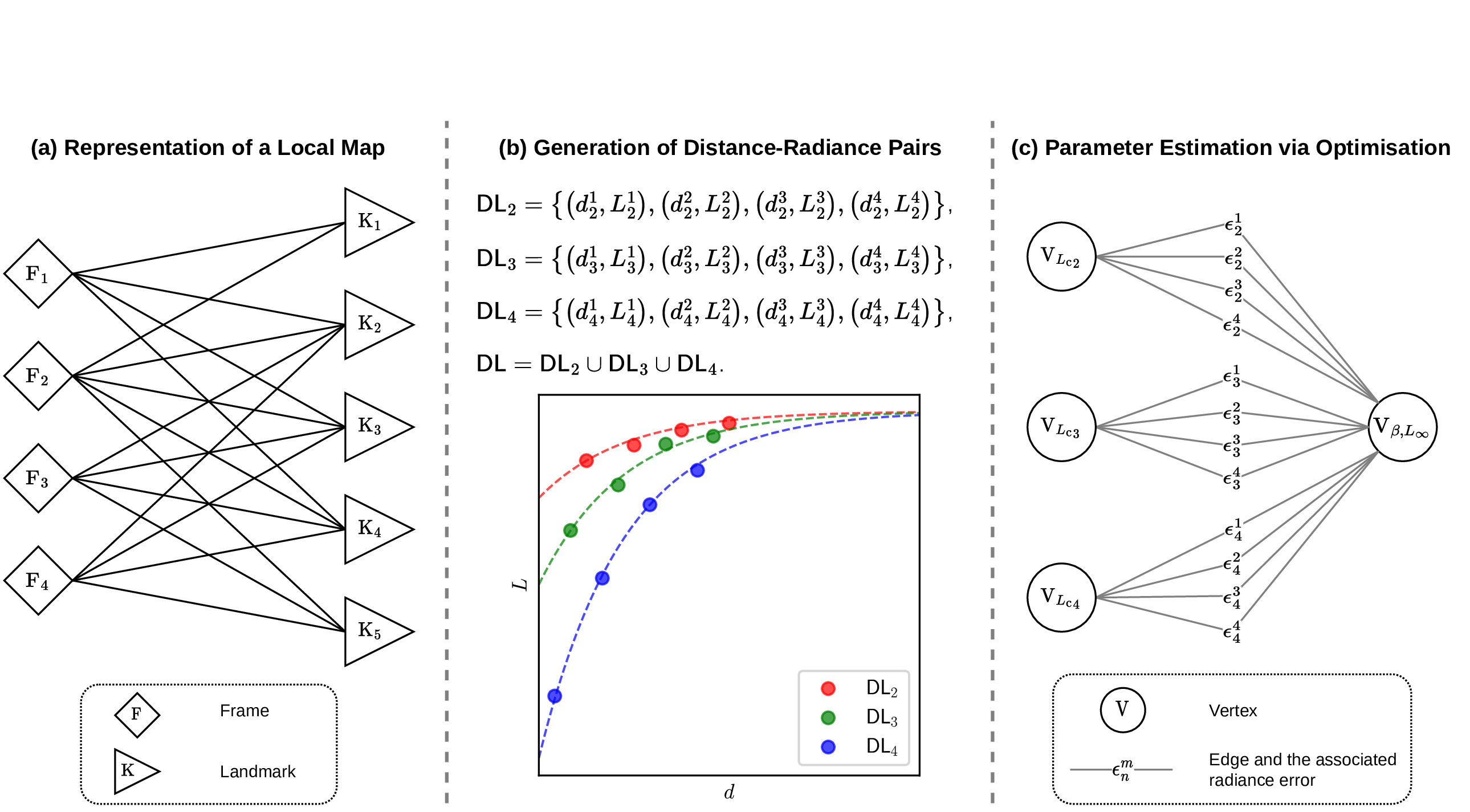} 
    \setlength{\abovecaptionskip}{-10pt}
    \vspace{0.4cm}
    \caption{
    Overview of our three-step method.
    (a) An example of a local map represented as a bipartite graph consisting of four frames, five landmarks and 17 edges which describe their observation relations (\RefSec \ref{subsec:representation_of_a_local_map}).
    (b) The corresponding distance-radiance pairs (\RefSec \ref{subsec:generation_of_dr_pairs}) and their scatter plot. The distance-radiance pairs of the same landmark share the same colour. The dashed curves are generated by \eqref{eq:pred_L_m_n} using ground truth values. There is no distance-radiance pair associated with $\text{K}_1$ or $\text{K}_5$ because they appear in a very small number of frames ($< 4$).
    (c) The corresponding optimisation problem (\RefSec \ref{subsec:parameter_estimation_via_optimisation}) depicted by a hypergraph.
    Note these figures are \emph{illustrative}.
    In reality, the local map typically contains many more landmarks, with each landmark being observed in many more frames.
    The graph is thus much larger in practice.
    }
    \setlength{\belowcaptionskip}{-10pt}
    \label{fig:method}
\end{figure*}

\textbf{Synthesised fog.}
Synthesised fog has been widely used with the aim of enriching foggy data, typically in the following three ways.
a) Real scenes with artificial fog:
\cite{daniel2017low, gruber2019pixel} deploy a fog machine to generate artificial fog in a controlled environment.
Although images are recorded in a wide range of visibility conditions, they only include very few preset scenes.
More importantly, the scenes remain static, which dramatically limits their use.
b) Real scenes with simulated fog:
To this end, dense ground truth depth must be obtained first before adding fog to clear images according to \eqref{eq:asm_intensity}.
Most work adopting this approach considers indoor scenes rather than outdoor ones because dense depth measurements are less difficult to acquire \cite{ren2016single, li2017aod}.
Indoor scenes, however, usually have a very limited depth range and therefore are, in general, not representative of outdoor ones.
In contrast, \cite{li2018benchmarking, sakaridis2018semantic} add simulated fog to outdoor scenes.
The authors rely on either monocular depth estimation \cite{liu2015learning} or stereoscopic inpainting \cite{wang2008stereoscopic} to generate dense pseudo-ground truth depth maps.
However, such a process can introduce undesirable artefacts in the synthesised foggy images due to erroneous depth data.
c) Simulated scenes with simulated fog:
In order to prepare training data (or at least part of it to be used for pre-training) for data-hungry deep neural networks, some researchers \cite{song2020simultaneous, yao2022foggystereo} add simulated fog to simulated scenes for which dense ground truth depth data is available.
However, in no way can such completely simulated data replace real recordings for real-world, extremely safety-oriented applications.

To summarise, no existing dataset focuses specifically on real road scenes recorded by a vehicle driving in fog comprising high-resolution, consecutive left and right images.
To address the above concerns, we present SDIRF.
Our dataset has the following two additional features.
\begin{itemize}
    \item We calibrated the camera's photometric parameters to make SDIRF photometrically ready for the deployment of the atmospheric scattering model.
    
    \item We also collected the counterpart clear images in overcast weather of the same routes.
    Such data can be useful for quantitative evaluation of downstream depth estimation and image defogging tasks.
\end{itemize}

\section{Methodology}   \label{sec:methodology}

In a nutshell, given a sequence of stereo foggy images, our method simultaneously estimates the parameters $\beta$, $L_{\infty}$ and relevant $L_{\text{c}}$s of the atmospheric scattering model in \eqref{eq:asm_radiance} and \eqref{eq:trans}, and dynamically updates them as new images become available.
\RefFig \ref{fig:method} depicts from left to right the three steps of our method: the representation of a local map (\RefSec \ref{subsec:representation_of_a_local_map}), the generation of distance-radiance pairs (\RefSec \ref{subsec:generation_of_dr_pairs}), and the parameter estimation via optimisation (\RefSec \ref{subsec:parameter_estimation_via_optimisation}).

\subsection{Representation of a Local Map}     \label{subsec:representation_of_a_local_map}
We first use the sequence of stereo images to build a local 3D feature map of the scene using a visual SLAM/odometry system such as \cite{mur2017orb}.
We build a local map, rather than a global one, for two reasons:
a) a local map entails a locally homogeneous fog model, as opposed to a globally homogeneous one;
b) the dimensions of the resulting optimisation problem are smaller and thus can be solved more efficiently.

A local map is a collection of observations describing which local frames observe which local landmarks.
It is represented as a bipartite graph $\mathsf{G}$ [\RefFig \ref{fig:method}(a) shows an example]:
\begin{equation}    \label{eq:bipartite_graph}
    \mathsf{G} = \left( \mathsf{F}, \mathsf{K}, \mathsf{E} \right) \text{,}
\end{equation}
where $\mathsf{F}$ denotes a set of left image frames, $\mathsf{K}$ denotes a set of 3D landmarks, and $\mathsf{E}$ denotes a set of edges each connecting a frame in $\mathsf{F}$ to a landmark in $\mathsf{K}$.
More specifically, an edge $\left( m,n \right) \in \mathsf{E}$ exists between the $m$th frame $\text{F}_m \in \mathsf{F}$ and the $n$th landmark $\text{K}_n \in \mathsf{K}$ only if $\text{F}_m$ observes $\text{K}_n$.

Let $ \mathsf{E}_n \subseteq \mathsf{E} $ denote the set of edges incident to $\text{K}_n$.
$\vert \mathsf{E}_n \vert$ is therefore the number of frames that observe $\text{K}_n$, and $ \mathsf{E} $ can be partitioned as
\begin{equation}    \label{eq:E}
    \mathsf{E} = \bigcup_{n=1}^{\vert \mathsf{K} \vert} \mathsf{E}_n \text{.}
\end{equation}

\subsection{Generation of Distance-Radiance Pairs}    \label{subsec:generation_of_dr_pairs}
Using the local map built in the previous step, we now generate observations that will serve as data in the subsequent optimisation step.
More specifically, these observations are distance-radiance pairs [\RefFig \ref{fig:method}(b)].
We use $\mathsf{DL}_n$ to denote the set of distance-radiance pairs associated with landmark $\text{K}_n$:
\begin{equation}   \label{eq:DLn}
    \mathsf{DL}_n = \{ \left( d_n^m, L_n^m \right) \, : \, \left( m, n \right) \in \mathsf{E}_n \} \text{,}
\end{equation}
where $d_n^m$ denotes the Euclidean distance between $\text{K}_n$ and $\text{F}_m$, and $L_n^m$ denotes the radiance of $\text{K}_n$ observed in $\text{F}_m$.
The distance $d_n^m$ can be calculated from the output of a sparse feature-based visual SLAM/odometry system, which typically consists of camera poses and landmark positions.
$I_n^m$, which denotes the pixel intensity of landmark $\text{K}_n$'s corresponding 2D feature point in frame $\text{F}_m$, is also typically available.
Next, we will explain how we derive $L_n^m$ from $I_n^m$.

\textbf{Gamma expansion/compression.}
We use $g \, : \, [0, 255] \to \mathbb{R}$ to denote the mapping from pixel intensity $I$ to radiance $L$.
$g$ is essentially a gamma expansion \cite{szeliski2022computer}, the inverse operation of a digital camera's image signal processor (ISP):
\begin{equation}    \label{eq:intensity_to_radiance}
    g\left( I \right) \coloneq \alpha I^{\gamma} + \zeta \text{,}
\end{equation}
where $\alpha > 0$, $\gamma > 1$ (hence the name ``gamma expansion'') and $\zeta \in \mathbb{R}$ are the photometric parameters that characterise $g$.
Inversely, we use $g^{-1} \, : \, \mathbb{R} \to [0, 255] $ to map radiance to intensity:
\begin{equation}    \label{eq:radiance_to_intensity}
    g^{-1} \left( L \right) \coloneq {\left( \frac{L - \zeta}{\alpha} \right)}^{\frac{1}{\gamma}}   \text{.}
\end{equation}
Calibrating the camera entails discovering the photometric parameters $\alpha$, $\gamma$ and $\zeta$.
See \RefSec \ref{subsec:photometric_calibration} for the details of our calibration procedure.

We use $\mathsf{DL}$ to denote the overall set of distance-radiance pairs, which can be expressed as the union of some (disjoint) sets in \eqref{eq:DLn}:
\begin{equation}    \label{eq:DL}
    \mathsf{DL} = \bigcup_{n \, : \, \vert \mathsf{E}_n \vert \geq \xi_{\text{F}}} \mathsf{DL}_n \text{,}
\end{equation}
where $\xi_{\text{F}} \in \mathbb{Z}_{++}$ is a threshold that ensures that a landmark is considered only if it is observed in at least $\xi_{\text{F}}$ frames.
As a result, the number of disjoint $\mathsf{DL}_n$s that form $\mathsf{DL}$ is typically smaller than $\vert \mathsf{K} \vert$.

Finally, in order to make the estimation of the fog parameters reliable, we require $\mathsf{DL}$ to consist of at least $\xi_{\text{K}} \in \mathbb{Z}_{++}$ disjoint $\mathsf{DL}_n$s:
\begin{equation}    \label{eq:xi_K}
    \vert \{ n \, : \, \vert \mathsf{E}_n \vert \geq \xi_{\text{F}} \} \vert \geq \xi_{\text{K}}   \text{.}
\end{equation}
This condition is a prerequisite for the next step, estimating the fog parameters via optimisation.

\subsection{Parameter Estimation via Optimisation}    \label{subsec:parameter_estimation_via_optimisation}
This step estimates the fog parameters $\beta$ and $L_{\infty}$, together with the clear radiance $L_{\text{c}}$ of the relevant landmarks, by minimising a cost function that uses the observations $\mathsf{DL}$ generated in the previous step.
The problem of interest can be represented as a hypergraph, in which vertices represent variables to optimise, and edges represent observation errors \cite{kummerle2011g}.
An edge connects two vertices that contribute to the underlying observation error.
In such a hypergraph [\RefFig \ref{fig:method}(c) shows an example], there are two types of vertices.
\begin{itemize}
    \item $ \text{V}_{\beta, L_{\infty}} $ encodes the fog parameters $\beta$ and $L_{\infty}$.
    When the fog is locally homogeneous, $\beta$ and $L_{\infty}$ are invariant within a local space and, in that case, there is only one such vertex.

    \item $ \text{V}_{{L_{\text{c}}}_n} $ encodes the clear radiance ${L_{\text{c}}}_n$ of the $n$th landmark $\text{K}_n$.
    The number of occurrences of such vertex is the same as the number of disjoint subsets in $\mathsf{DL}$ \eqref{eq:DL}.
\end{itemize}

According to the atmospheric scattering model in \eqref{eq:asm_radiance} and \eqref{eq:trans}, we can compute the predicted radiance value of the $n$th landmark $\text{K}_n$ observed in the $m$th frame $\text{F}_m$ from the distance $d_n^m$ between $\text{K}_n$ and $\text{F}_m$, the scattering coefficient $\beta$, the atmospheric light radiance $L_{\infty}$, and $\text{K}_n$'s clear radiance ${L_{\text{c}}}_n$:
\begin{equation}    \label{eq:pred_L_m_n}
    \begin{split}
        {}_{\text{pred}}L_n^m & = {L_{\text{c}}}_n \exp{ \left( -\beta d_n^m \right) } + L_{\infty} \left( 1 - \exp{\left( -\beta d_n^m \right)} \right) \\
        & = \left( {L_{\text{c}}}_n - L_{\infty} \right) \exp{ \left( -\beta d_n^m \right) } + L_{\infty}   \text{.}
    \end{split}
\end{equation}
We define an error term $\epsilon_n^m \in \mathbb{R}$ to be the difference between the observed radiance $ L_n^m $ and the corresponding estimated radiance ${}_{\text{pred}}L_n^m$:
\begin{equation}    \label{eq:e_m_n}
    \begin{split}
        \epsilon_n^m & = L_n^m - {}_{\text{pred}}L_n^m \\
        & = L_n^m - \left[ \left( {L_{\text{c}}}_n - L_{\infty} \right) \exp{ \left( -\beta d_n^m \right) } + L_{\infty} \right]    \text{.}
    \end{split}
\end{equation}
We can see that each $\epsilon_n^m$ depends on $\beta$, $L_{\infty}$ and ${L_{\text{c}}}_n$, and is therefore associated with an edge between $ \text{V}_{\beta, L_{\infty}} $ and $ \text{V}_{{L_{\text{c}}}_n} $ in the hypergraph.
We define
\begin{equation}   \label{eq:Epsilon_n}
    \mathcal{E}_n = \{ \epsilon_n^m \, : \, \left( m, n \right) \in \mathsf{E}_n \}
\end{equation}
as the set of radiance errors associated with $\text{K}_n$.
Also, $\mathcal{E}$ will represent the overall set of radiance errors, which can be expressed as the union of some (disjoint) sets in \eqref{eq:Epsilon_n}:
\begin{equation}    \label{eq:Epsilon}
    \mathcal{E} = \bigcup_{n \, : \, \vert \mathcal{E}_n \vert \geq \xi_{\text{F}}} \mathcal{E}_n \text{.}
\end{equation}
We define each residual term to be a loss function, e.g., Huber loss or square loss, $\ell \, : \, \mathbb{R} \to \mathbb{R}_{+}$ of $\epsilon_n^m$.
The total cost function is a weighted sum of all residual terms.
Our goal is to solve:
\begin{equation}    \label{eq:optimisation_problem}
    \begin{array}[t]{cl}
    \underset{\beta, L_{\infty}, \{ {L_{\text{c}}}_n \, : \, \mathcal{E}_n \subset \mathcal{E} \}}{\text{minimise}}
    &
    \sum_{ n \, : \, \mathcal{E}_n \subset \mathcal{E} } \sum_{ m \, : \, \epsilon_n^m \in \mathcal{E}_n } w_n^m \ell \left( \epsilon_n^m \right) 
    \\[0.3cm]
    \text{subject to} 
    &
    l_{\beta} \leq \beta \leq u_{\beta}
    \\
    & l_{L_{\infty}} \leq L_{\infty} \leq u_{L_{\infty}} 
    \\ 
    & l_{{L_{\text{c}}}_n} \leq {L_{\text{c}}}_n \leq u_{{L_{\text{c}}}_n} \,,
    \end{array}
\end{equation}
where $ \{ {L_{\text{c}}}_n \, : \, \mathcal{E}_n \subset \mathcal{E} \} $ are the radiances of the relevant landmarks, $w_n^m \in \mathbb{R}_{+}$ is the weight associated with $\epsilon_n^m$, and $l$s and $u$s are the lower and upper bounds of the parameters, respectively.
In the following, we first analyse \eqref{eq:optimisation_problem} and then describe our initialisation scheme.
We also explain how we set each $l$ and $u$, our two-stage strategy for solving \eqref{eq:optimisation_problem}, and finally how we set each weight $w_n^m$.

\subsubsection{Analysis of~\eqref{eq:optimisation_problem}}    \label{subsubsec:theoretical_analyses}
We show that
a) \eqref{eq:optimisation_problem} is non-convex;
b) when the gamma correction in~\eqref{eq:intensity_to_radiance} is non-linear, i.e., $\gamma \neq 1$, solving \eqref{eq:optimisation_problem} using $\epsilon_n^m$ computed in the radiance domain or in the intensity domain yields problems that are not equivalent.

\textbf{Non-convexity.}
Consider an arbitrary error term $\epsilon_n^m$ and, without loss of generality, the square loss.
Then, each unweighted term in the objective of \eqref{eq:optimisation_problem} can be written as
\begin{equation}
    f\left( \beta, L_\infty, {L_{\text{c}}}_n \right) 
    \!=\! 
    \left( L_n^m - \left[ \left( {L_{\text{c}}}_n - L_{\infty} \right) \exp{ \left( -\beta d_n^m \right) } + L_{\infty} \right] \right)^2\!.
    \label{eq:f_loss}
\end{equation}
A function is convex if and only if it is convex when restricted to any line intersecting its domain \cite[\S3.1.1]{boyd2004convex}. Consider then
\eqref{eq:f_loss} restricted to the line $\beta = L_\infty = \nu, {L_{\text{c}}}_n = 0$, for $\nu \geq 0$:
\begin{equation*}
    \phi\left(\nu\right) := f\left( \nu,\nu,0 \right)
    = \left( b + \nu\exp\left(-a\nu\right) - \nu \right)^2    \text{,}
    \label{eq:f_loss_}
\end{equation*}
where $a, b \geq 0$ are constants.
The 2nd-order derivative of $\phi$ with respect to $\nu$ is
\begin{multline*}
 \phi''\left(\nu\right)
     = 2 \left[\exp\left(-a\nu\right) \left(1-a\nu\right) - 1\right]^2 
    + 
    2 \big[b \\ + \nu\exp\left(-a\nu\right) - \nu\big] 
    a \exp\left(-a\nu\right) \left(\nu - 2\right)\,.
\end{multline*}
Setting, for example, $a=b=1$, one obtains $\phi''\left(0.2\right) \approx -2.6 < 0$, where the function is concave, and $\phi''\left(1\right) \approx 1.73 > 0$, where the function is convex.
Thus, $f$ is neither concave nor convex.
In other words, it is non-convex.
See our supplementary material for a visual example of non-convexity.

\textbf{Impact of gamma correction.}
As above, consider an arbitrary error term $\epsilon_n^m$ and a square loss. When we use radiance, the corresponding unweighted term in \eqref{eq:optimisation_problem} is given by \eqref{eq:f_loss}.
Suppose now that we apply no gamma correction.
In this case, we directly use intensity and apply \eqref{eq:asm_intensity}, making the corresponding term in \eqref{eq:optimisation_problem}
\begin{equation}
f_{\text{int}}(\beta_{\text{int}}, A, J_n) =
    \left( I_n^m - \left[ \left( J_n - A \right) \exp{ \left( -\beta_{\text{int}} d_n^m \right) } + A \right] \right)^2,
    \label{eq:loss_term_intensity}
\end{equation}
where $\beta_{\text{int}}$ denotes the scattering coefficient when we use intensity.
Substituting $I_n^m$ with $g^{-1} \left( L_n^m \right)$ defined in \eqref{eq:radiance_to_intensity}, \eqref{eq:loss_term_intensity} becomes
\begin{equation}
    \bigg[ \left(\frac{L_n^m - \zeta}{\alpha}\right)^{\frac{1}{\gamma}} - \left[ \left( J_n - A \right) \exp{ \left( -\beta_{\text{int}} d_n^m \right) } + A \right] \bigg]^2.
\label{eq:loss_term_intensity_transformed}
\end{equation}
When $\gamma = 1$ (i.e., $g$ is affine),  \eqref{eq:loss_term_intensity_transformed} can be written as
\begin{multline}
    \alpha^{-2} \bigg[ L_n^m - \Big[ \left( \left( \alpha J_n + \zeta \right) - \left( \alpha A + \zeta \right) \right) \exp{ \left( -\beta_{\text{int}} d_n^m \right) } 
    \\
    + \left (\alpha A + \zeta \right) \Big] \bigg]^2.
\label{eq:loss_term_intensity_transformed_affine_gamma_correction}
\end{multline}
Comparing \eqref{eq:f_loss} and \eqref{eq:loss_term_intensity_transformed_affine_gamma_correction}, we observe that the two minimisation problems are related by a positive scaling and the following one-to-one mappings for the variables: $\beta \leftarrow \beta_{\text{int}}$, ${L_{\text{c}}}_n \leftarrow \alpha J_n + \zeta$ and $L_{\infty} \leftarrow \alpha A + \zeta$.
Thus, when $\gamma = 1$, \eqref{eq:f_loss} and \eqref{eq:loss_term_intensity} yield equivalent problems \cite[\S4.1.3]{boyd2004convex}.

In the typical case where $\gamma \neq 1$, the gamma correction alters the structure of the model such that the above equivalence no longer holds.
To see this, let $h\left(\gamma\right) := [(L_n^m - \zeta)/\alpha]^{1/\gamma}$ and $c:= (L_n^m - \zeta)/\alpha$.
A Taylor expansion of $h\left(\gamma\right)$ around $\gamma=1$ yields 
\begin{align}
    &
    h\left(1\right) + h'\left(1\right)\left(\gamma-1\right) + \frac{1}{2}h''\left(1\right)\left(\gamma-1\right)^2 + \cdots 
    \notag
    \\
    =\,\,& 
    c - \left(\ln{c}\right) c \left(\gamma-1\right) + \frac{1}{2} \left(\ln{c}\right) \left( 2 + \ln{c} \right) c \left(\gamma-1\right)^2 + \cdots \text{.}
    \label{eq:taylor}
\end{align}
We can see that if any of the 1st-order and the subsequent higher-order terms in \eqref{eq:taylor} is non-zero, then \eqref{eq:f_loss} and \eqref{eq:loss_term_intensity} no longer yield equivalent problems.
In \RefSec \ref{subsubsec:gamma_correction}, we will show this experimentally and investigate how the estimate of $\beta$ is affected by this non-linearity.

\subsubsection{Initialisation}
We will solve \eqref{eq:optimisation_problem} with an iterative algorithm, e.g., Levenberg–Marquardt.
However, since \eqref{eq:optimisation_problem} is non-convex, it is critical to initialise $\beta$, $L_{\infty}$ and $ \{ {L_{\text{c}}}_n \, : \, \mathcal{E}_n \subset \mathcal{E} \} $ properly.
Our initialisation strategy is as follows.
Assuming the fog to be locally homogeneous, if we have access to previous estimates $\hat{\beta}$, ${\hat{L}}_{\infty}$ and $ \{ {{\hat{L}}_{\text{c}}}_n \, : \, \mathcal{E}_n \subset \mathcal{E} \} $, obtained from the last run of our fog estimation process, we use these values as initialisation.
Otherwise (i.e., if the fog estimation process has never run before, or if ${L_{\text{c}}}_n$ has never been estimated before), we set $\beta=0.014$, which is the geometric mean of its lower and upper bounds (\RefSec \ref{subsubsec:param_bounds}).
For $L_{\infty}$, we use the radiance of all landmarks observed from the maximal distance.
And for ${L_{\text{c}}}_n$, we use the radiance of the corresponding landmark observed from the minimal distance.

\subsubsection{Parameter Bounds}    \label{subsubsec:param_bounds}
We set $l_{\beta} = 0.001$ and $u_{\beta} = 0.2$, which, according to \eqref{eq:vis}, corresponds to a visibility range of $[15, 3000]$ meters, values that are conservative.

Next, we set the bounds for each ${L_{\text{c}}}_n$, and build a candidate set for $l_{L_{\infty}}$ at the same time, as explained below.
For each ${L_{\text{c}}}_n$, we first determine if it is lower or higher than $L_{\infty}$ by computing the slope $k_n$ of the line going through the intensities observed at the maximal and the minimal distances:
\begin{equation}    \label{eq:slope}
    k_n = \left( g^{-1}\left( L_{n}^{d_\text{max}} \right) - g^{-1}\left( L_{n}^{d_\text{min}} \right) \right) / \left( d_\text{max} - d_\text{min} \right) \text{,}
\end{equation}
which will then be compared to a threshold $\eta \in \mathbb{R}_{++}$.
If $k_n > \eta$ (i.e., strongly positive), we set $l_{{L_{\text{c}}}_n} = g \left( 0 \right)$ and $u_{{L_{\text{c}}}_n} = {L_{\text{c}}}_{n}^{d_\text{min}}$, and we add ${L_{\text{c}}}_{n}^{d_\text{max}}$ to the candidate set for $l_{L_{\infty}}$.
If $k_n < -\eta$ (i.e., strongly negative), we set $l_{{L_{\text{c}}}_n} = {L_{\text{c}}}_{n}^{d_\text{min}}$ and $u_{{L_{\text{c}}}_n} = g \left( 255 \right)$.
If neither, we set $l_{{L_{\text{c}}}_n} = g \left( 0 \right)$ and $u_{{L_{\text{c}}}_n} = g \left( 255 \right)$.
See our supplementary material for more details.

Finally, we let $l_{L_{\infty}}$ be the median value of its candidate set, and $u_{L_{\infty}} = g \left( 255 \right)$.
We noticed if we set $u_{L_{\infty}}$ in a similar way to $l_{L_{\infty}}$ (i.e., let $u_{L_{\infty}}$ be the median value of its candidate which consists of ${L_{\text{c}}}_{n}^{d_\text{max}}$ when $k_n < -\eta$), its value is often underestimated.
We think this is caused by the fact that objects that are brighter than $L_{\infty}$ are rare in a foggy scene.

\subsubsection{Two-stage Optimisation}  \label{subsubsec:two_stage_optimisation}
We adopt a two-stage optimisation strategy following \cite{mur2017orb}.
In the first stage, we let $\ell$ be the Huber loss (with parameter $\delta$) in order to mitigate the effect of outlier observations.
In the second stage, we let $\ell$ be the square loss and perform optimisation using inlier observations only.
After the first stage, our system keeps track of the number of times each relevant observation, i.e., each relevant edge in the bipartite graph of \RefFig \ref{fig:method}(a), is classified as an inlier, which is done by evaluating each residual term and comparing it with $\delta$.
That number, $c_n^m \in \mathbb{Z}_{+}$, for landmark $\text{K}_n$ observed in frame $\text{F}_m$, will be used to set the weight $w_n^m$ in \eqref{eq:optimisation_problem}, as explained below.

\subsubsection{Residual Weights}
The partial derivatives of \eqref{eq:e_m_n} are
\begin{subequations}
\label{eq:partialderivsloss}
\begin{align}
    \frac{\partial \epsilon_n^m}{\partial \beta} &= d_n^m \left( {L_{\text{c}}}_n - L_{\infty} \right) \exp{\left( -\beta d_n^m \right)} \text{,}  \label{eq:par_der_e_beta} \\
    \frac{\partial \epsilon_n^m}{\partial L_{\infty}} &= \exp{\left( -\beta d_n^m \right)} - 1  \text{,}   \label{eq:par_der_e_A} \\
    \frac{\partial \epsilon_n^m}{\partial {L_{\text{c}}}_n} &= - \exp{\left( -\beta d_n^m \right)}  \text{.}   \label{eq:par_der_e_J}
\end{align}
\end{subequations}
We argue that the larger the radiance difference between a landmark's $L_{\text{c}}$ and $L_{\infty}$, the more suitable that landmark is for estimating $\beta$.
As can be seen from \eqref{eq:par_der_e_beta}, the partial derivative of $\epsilon_n^m$ with respect to $\beta$ is proportional to $\left( {L_{\text{c}}}_n - L_{\infty} \right)$.
This suggests that when ${L_{\text{c}}}_n$ is close to $L_{\infty}$ this term diminishes, causing difficulties in finding the optimal $\beta$.
Intuitively, when ${L_{\text{c}}}_n$ is close to $L_{\infty}$, the range of the predicted radiance ${}_{\text{pred}}L_n^m$ flattens out according to \eqref{eq:pred_L_m_n} and therefore $\epsilon_n^m$ contains very little information on the inference of $\beta$.

In light of this, we heuristically set the weight in our first optimisation stage to be the product of the following two terms: the absolute difference between the previous estimates ${{\hat{L}}_{\text{c}}}_n$ and ${\hat{L}}_{\infty}$, and the current inlier count of the corresponding observation plus one:
\begin{equation}
    w_n^m = \lvert {{\hat{L}}_{\text{c}}}_n - {\hat{L}}_{\infty} \rvert \cdot \left( c_n^m + 1 \right) \text{.}
\end{equation}
It can be seen that the first term is landmark-dependent, while the second term is observation-dependent.

Our weighting scheme in the first stage is apparently related to iteratively reweighted minimisation strategies proposed in the sparse regression literature, for example \cite{candes2008enhancing, daubechies2010iteratively}, which comes with theoretical guarantees for convergence even for some non-convex problems.
Such a strategy, however, involves solving a sequence of optimisation problems, making it computationally expensive.
We therefore opt to solve just one instance of \eqref{eq:optimisation_problem} at a given frame.
Specifically, each weight $w_n^m$, which is computed from estimates from a previous frame, remains constant while solving~\eqref{eq:optimisation_problem} for the current frame.

In our second optimisation stage where only inlier observations are used, we use uniform weighting.

Results of our ablation study (\RefSec \ref{subsubsec:additional_experiments}) show that our weighting scheme performs better than naively uniformly weighting all residual terms in both optimisation stages.

\section{SDIRF Dataset} \label{sec:sdirf_dataset}
In this section, we introduce the data collection and the photometric calibration of SDIRF.
More details can be found in our supplementary material.

\subsection{On-road Data Collection}
We collected the on-road data during September 2023 in Rosyth, Queensferry and Penicuik near Edinburgh, Scotland.
Foggy data was collected first, and the counterpart clear data in overcast weather was collected two weeks later by traversing the same routes.

We used an off-the-shelf stereo camera ZED 2i which features an electronic synchronised rolling shutter and a built-in inertial measurement unit (IMU).
It was placed behind the windshield of a car and connected to a laptop installed with the ZED software development kit (SDK) and the ZED Robot Operating System (ROS) wrapper.
Considering the way the windshield inclines and that the mounting holes are located at the bottom of the camera but not at its top, it was mounted upside down for convenience and safety reasons (see our supplementary material for a picture of the setup).
We accordingly set the ``camera\_flip” parameter to “true” to allow the SDK to account for this upside-down setup to generate the correct stereo images.
The left frames of these images, whose poses are estimated by a SLAM algorithm in our method and to which the 3D positions of all landmarks are referenced, were actually acquired by the right camera because of this upside-down setup.
Therefore, our photometric calibration was later performed on the right camera (\RefSec \ref{subsec:photometric_calibration}).

We disabled the camera's auto-white balance and auto-exposure functionalities.
Instead, we used a fixed white balance, and manually adjusted the exposure time, in conjunction with the gain, according to the lighting condition of the scene.
We took note of the combination of exposure time and gain used to collect each data sequence.
Later we performed a photometric calibration for each combination of these two parameters (\RefSec \ref{subsec:photometric_calibration}).

During the collection, the data, together with the timestamps, was logged to ROS bags, which were later parsed to generate the following files.
\begin{itemize}
    \item Rectified\footnote{The stereo rectification was performed by the camera's SDK using its factory calibrated stereo parameters.} left and right images at a frame rate of $15$ Hz saved as PNG files, each at a resolution of $1920 \times 580$\footnote{The original image size was $1920 \times 1080$. The images were later cropped by 290 pixels at the top (to remove the mostly-sky region) and by 210 pixels at the bottom (to remove the car's bonnet and interior reflections caused by the windshield). The final image size after cropping has an aspect ratio of $1920/580 \approx 3.31$, which is very close to the aspect ratio ($1241/376 \approx 3.30$) of the images in the KITTI odometry benchmark \cite{geiger2012we}. See our supplementary material for an example image that illustrates the effect of our cropping.}.

    \item IMU data at a rate of $400$ Hz saved as CSV files.

    \item Magnetometer data at a rate of $50$ Hz saved as CSV files.
\end{itemize}

\begin{figure*}[!t]
    \centering
    \setlength\tabcolsep{1pt} 
    \renewcommand{\arraystretch}{0.5}
    \begin{tabular}{cc:c:c:c}

        &
        \tiny{(a)} &
        \tiny{(b)} &
        \tiny{(c)} &
        \tiny{(d)} \\

        \rotatebox[origin=lc]{90}{\tiny{\quad \; Thin fog}} &
        \includegraphics[width=0.244\textwidth]{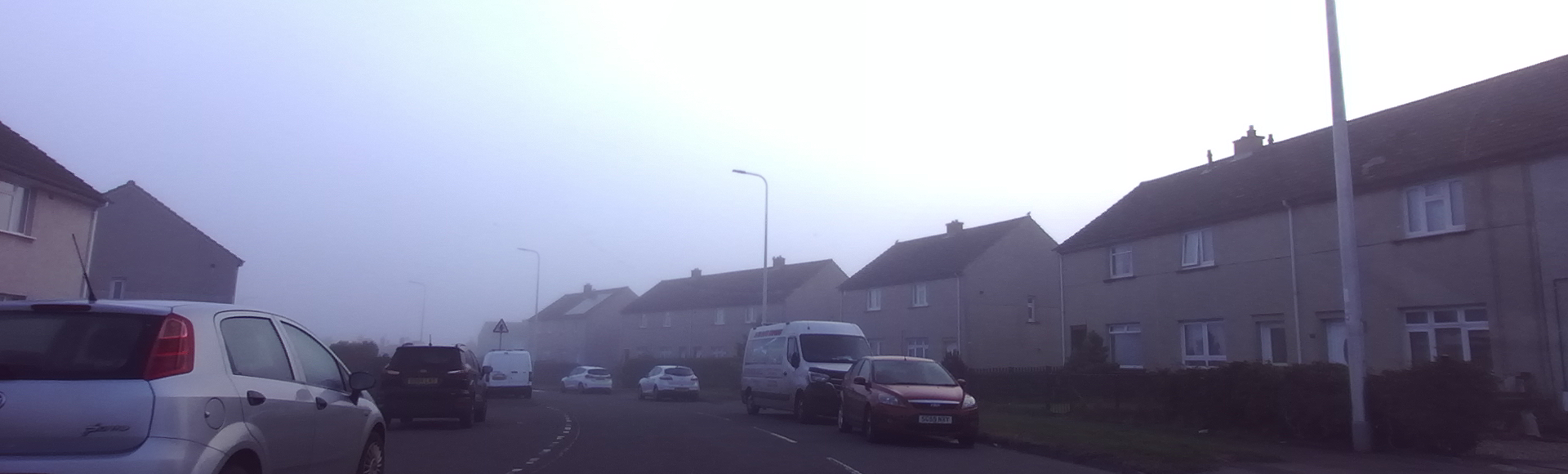} &
        \includegraphics[width=0.244\textwidth]{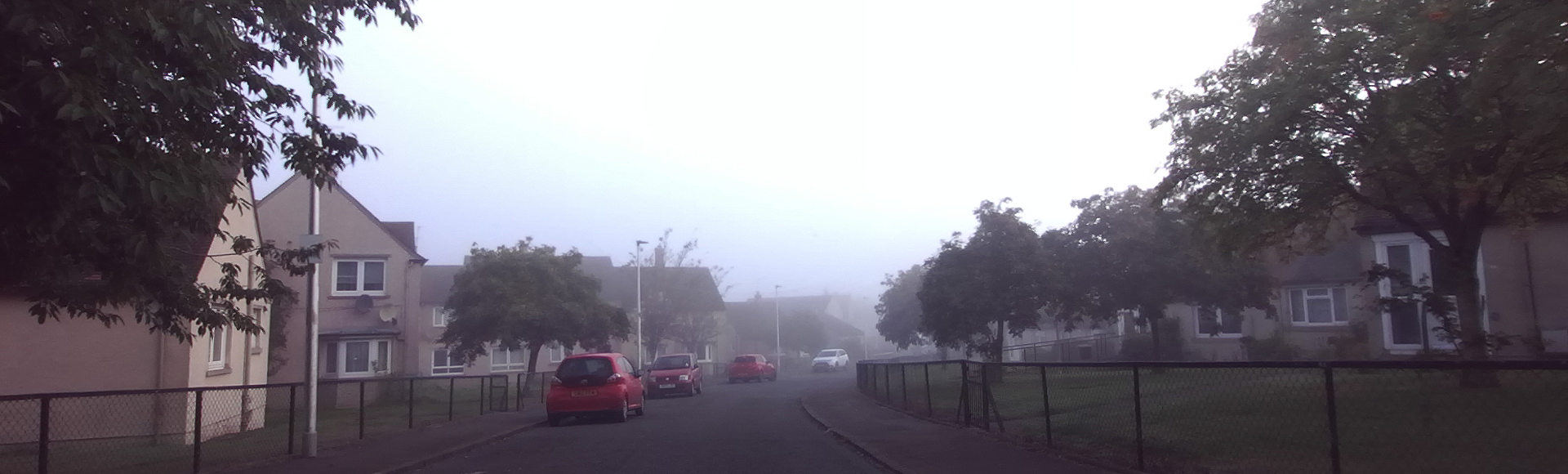} &
        \includegraphics[width=0.244\textwidth]{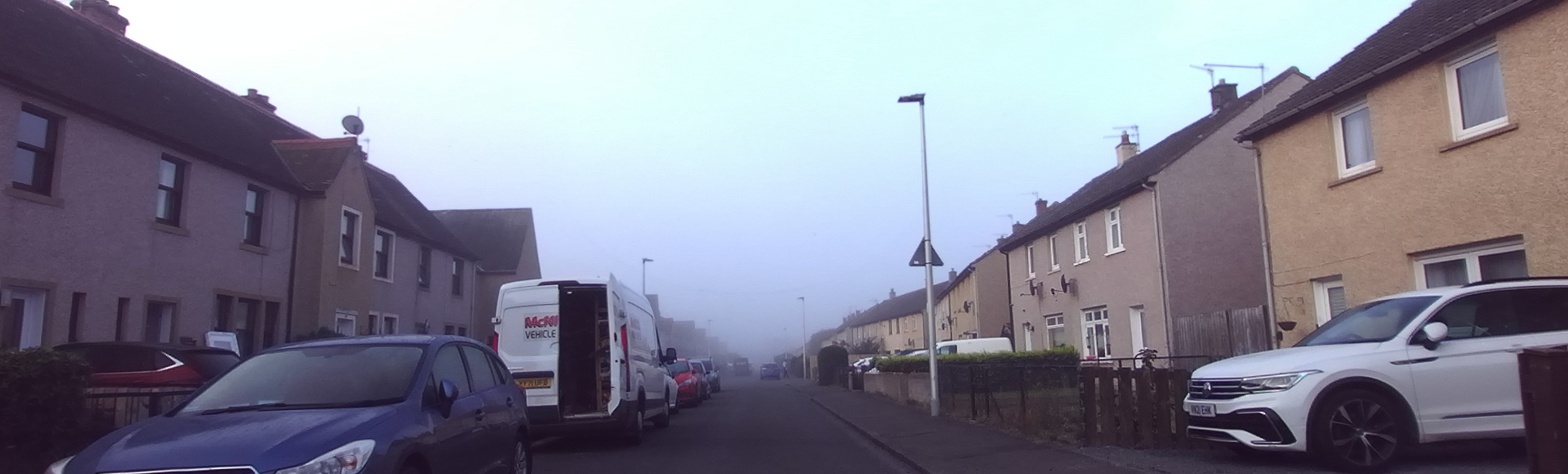} &
        \includegraphics[width=0.244\textwidth]{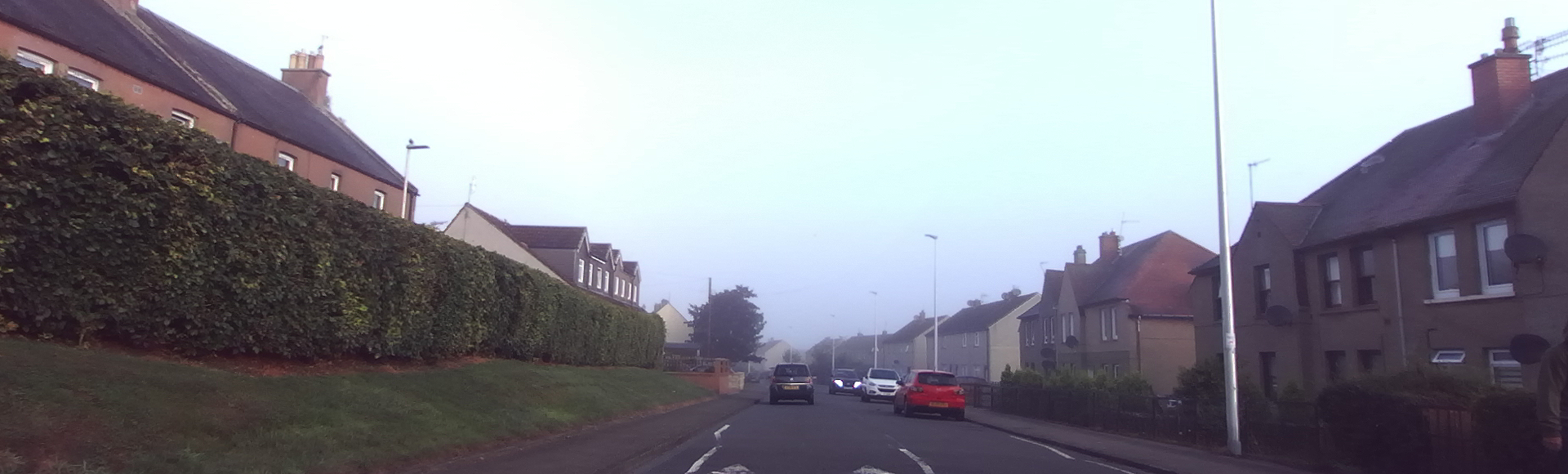} \\

        \rotatebox[origin=lc]{90}{\tiny{\qquad \: Clear}} &
        \includegraphics[width=0.244\textwidth]{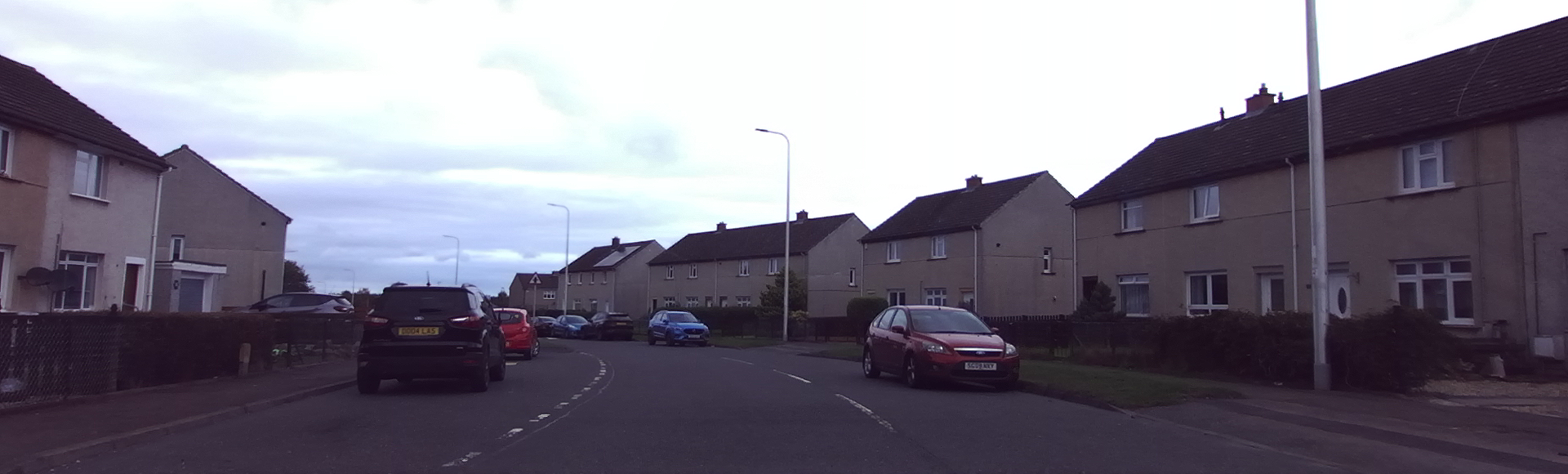} &
        \includegraphics[width=0.244\textwidth]{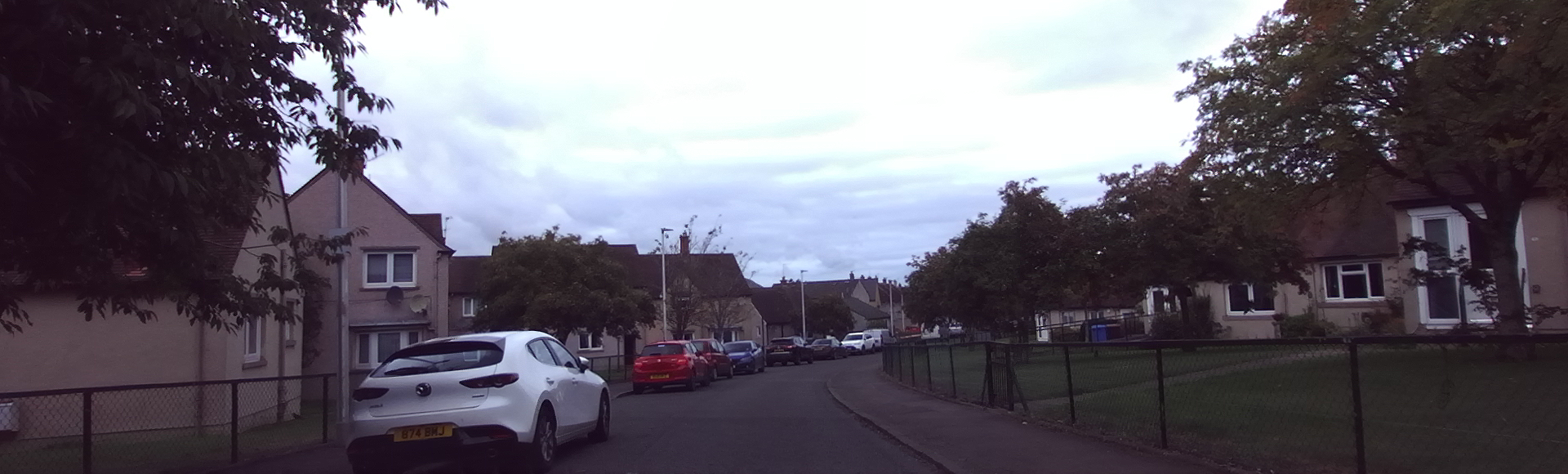} &
        \includegraphics[width=0.244\textwidth]{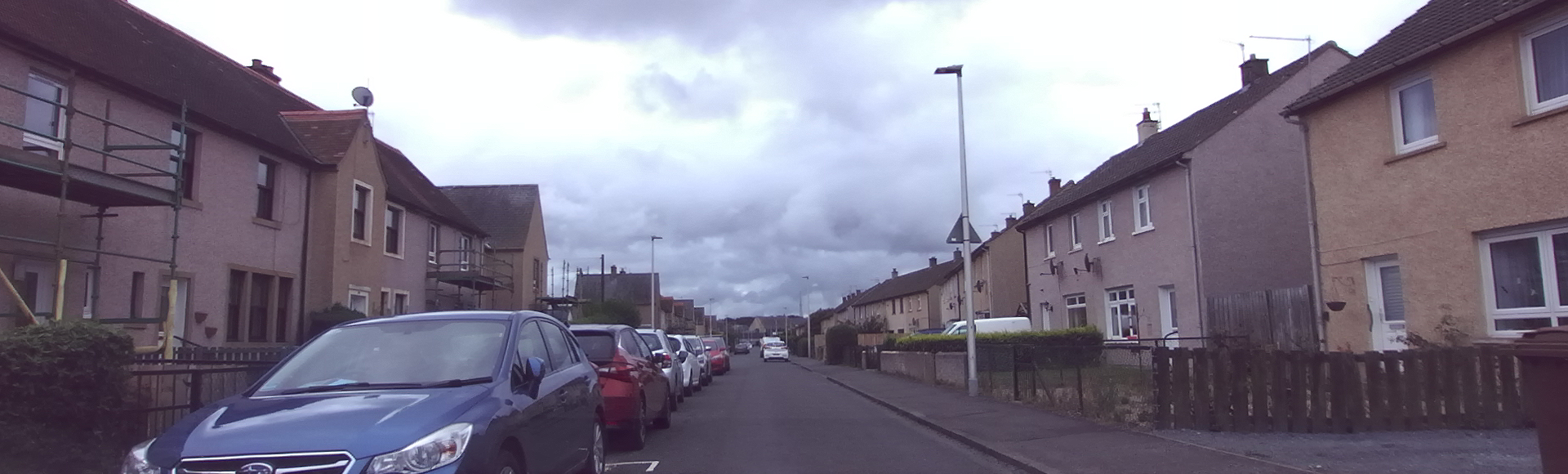} &
        \includegraphics[width=0.244\textwidth]{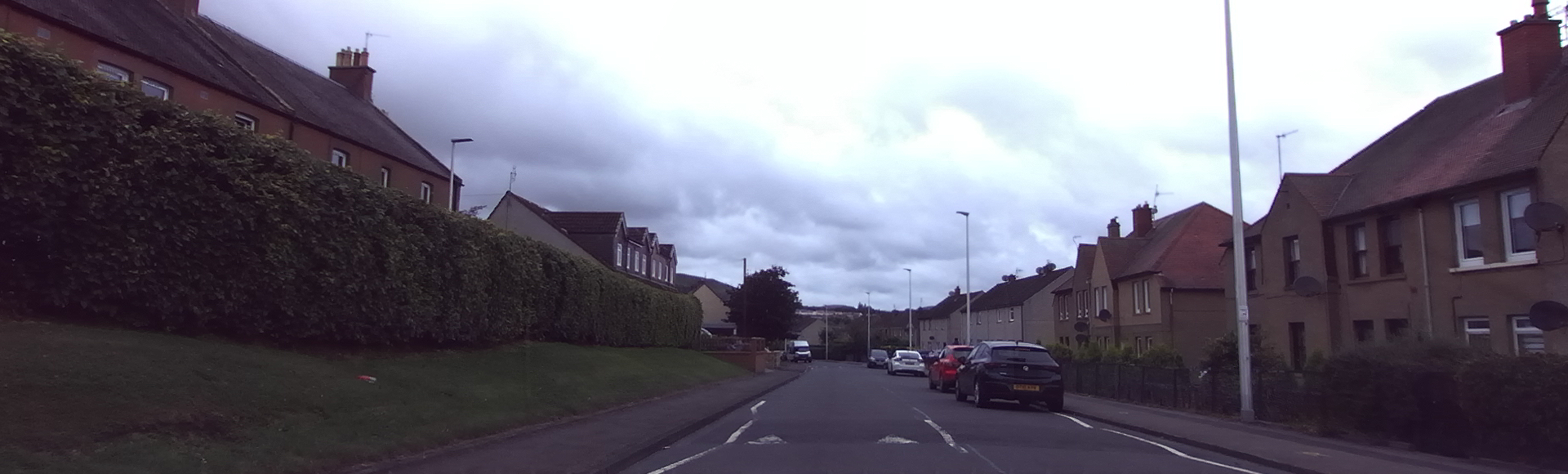} \\

        \hdashline
        \addlinespace[0.4ex]
        
        \rotatebox[origin=lc]{90}{\tiny{\quad \: Thick fog}} &
        \includegraphics[width=0.244\textwidth]{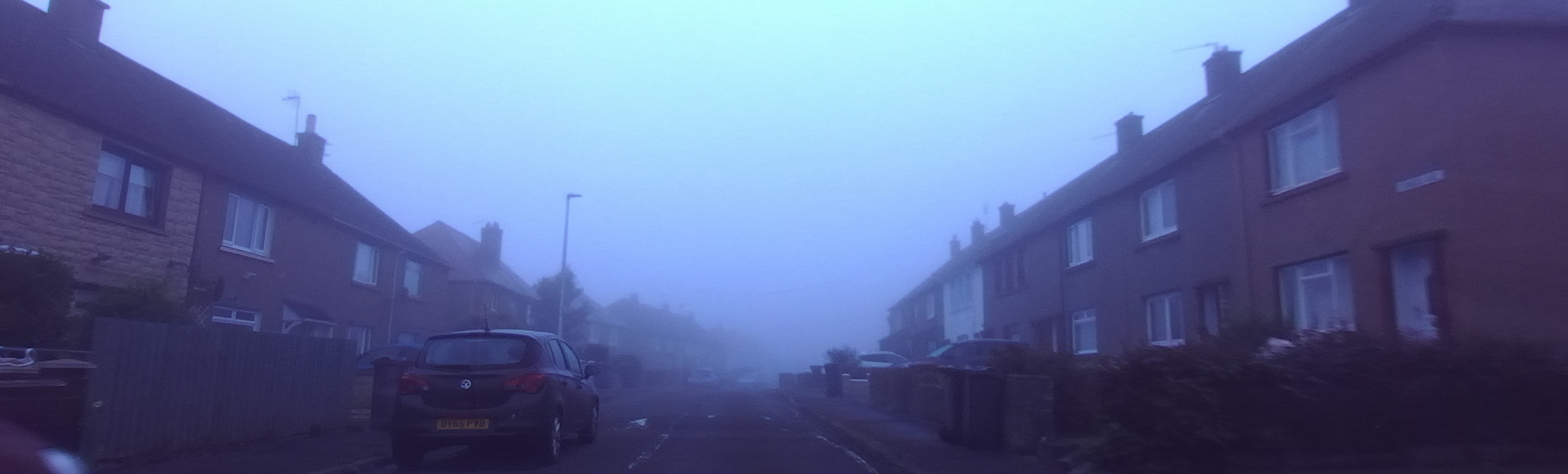} &
        \includegraphics[width=0.244\textwidth]{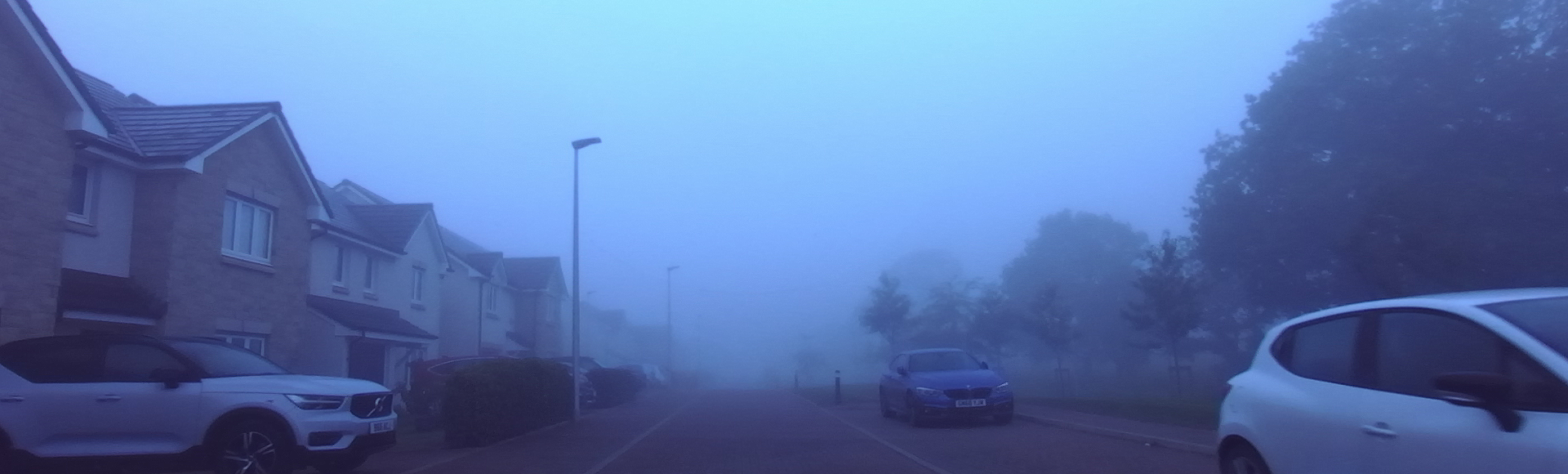} &
        \includegraphics[width=0.244\textwidth]{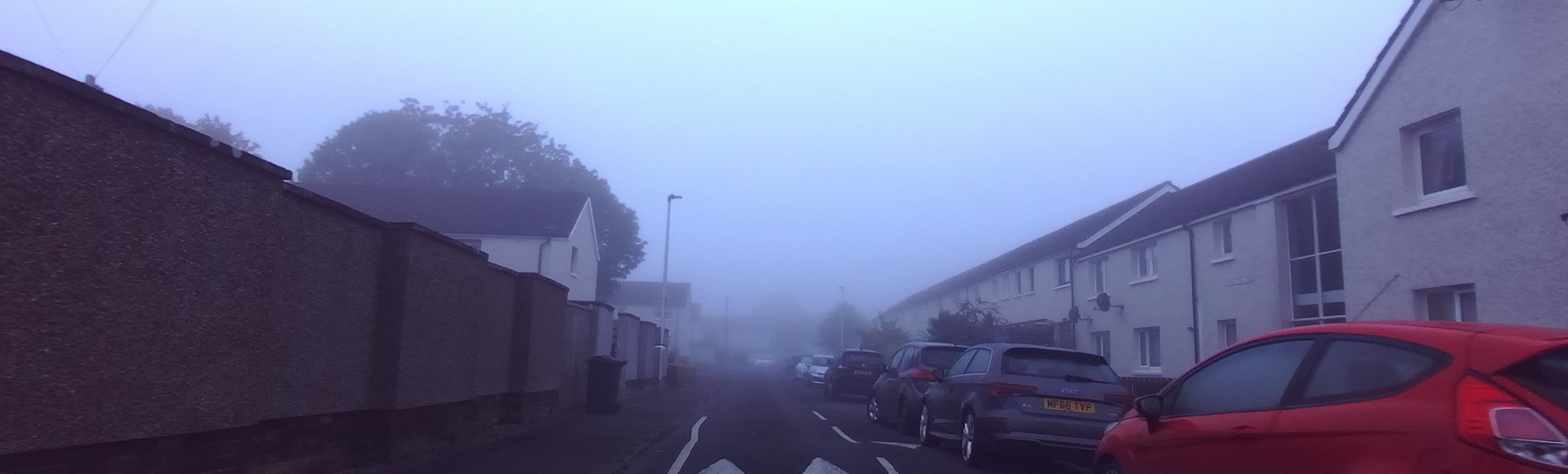} &
        \includegraphics[width=0.244\textwidth]{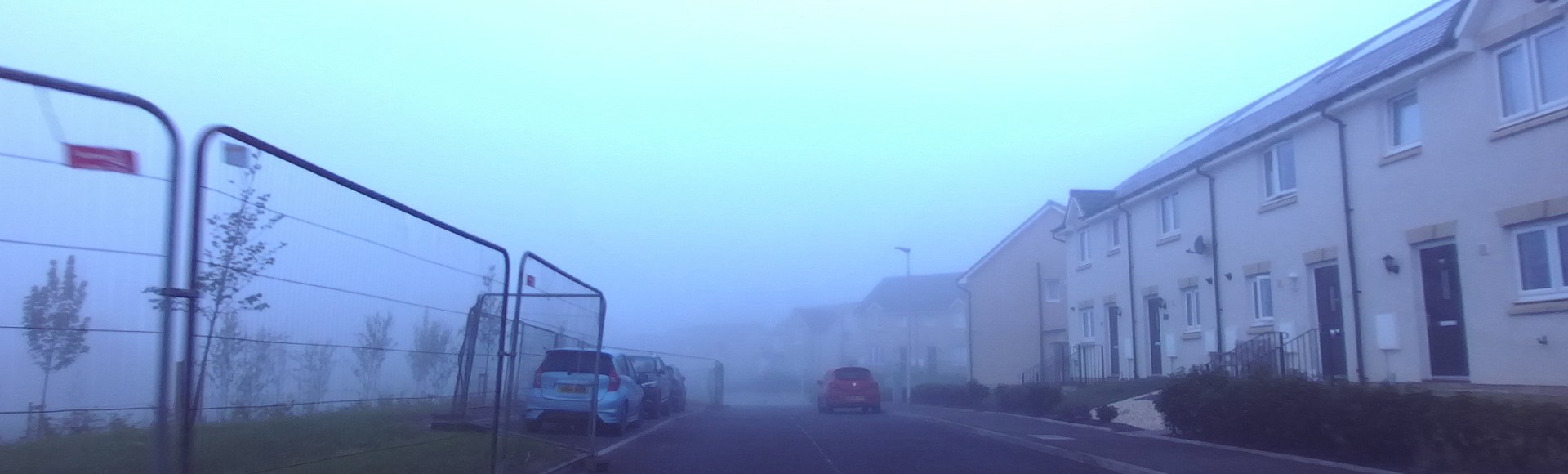} \\

        \rotatebox[origin=lc]{90}{\tiny{\qquad \: Clear}} &
        \includegraphics[width=0.244\textwidth]{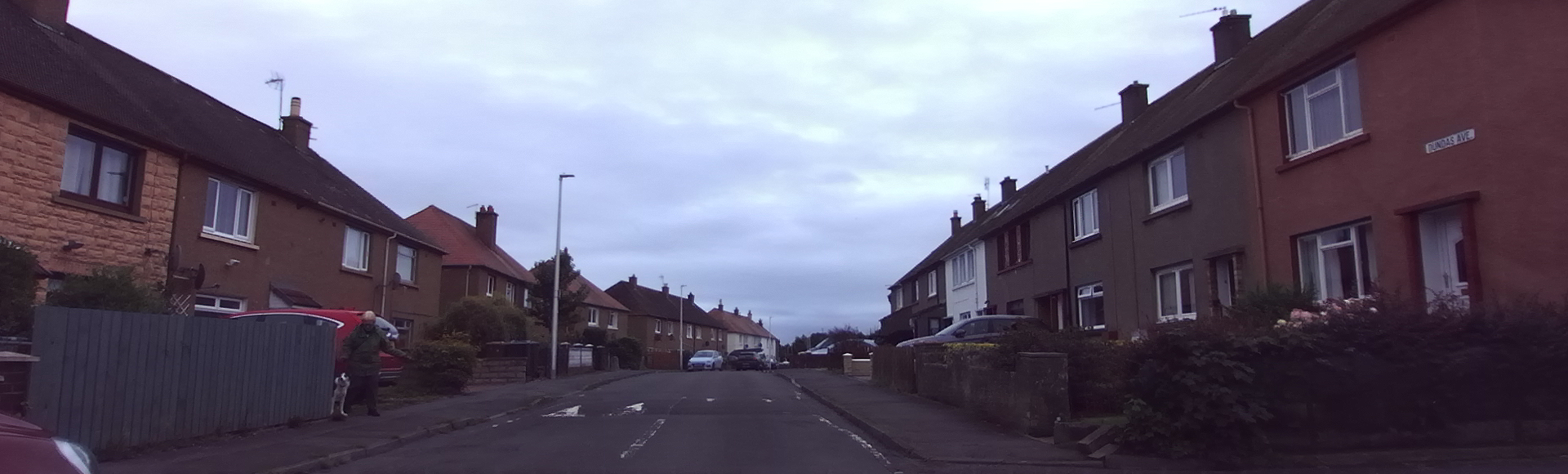} &
        \includegraphics[width=0.244\textwidth]{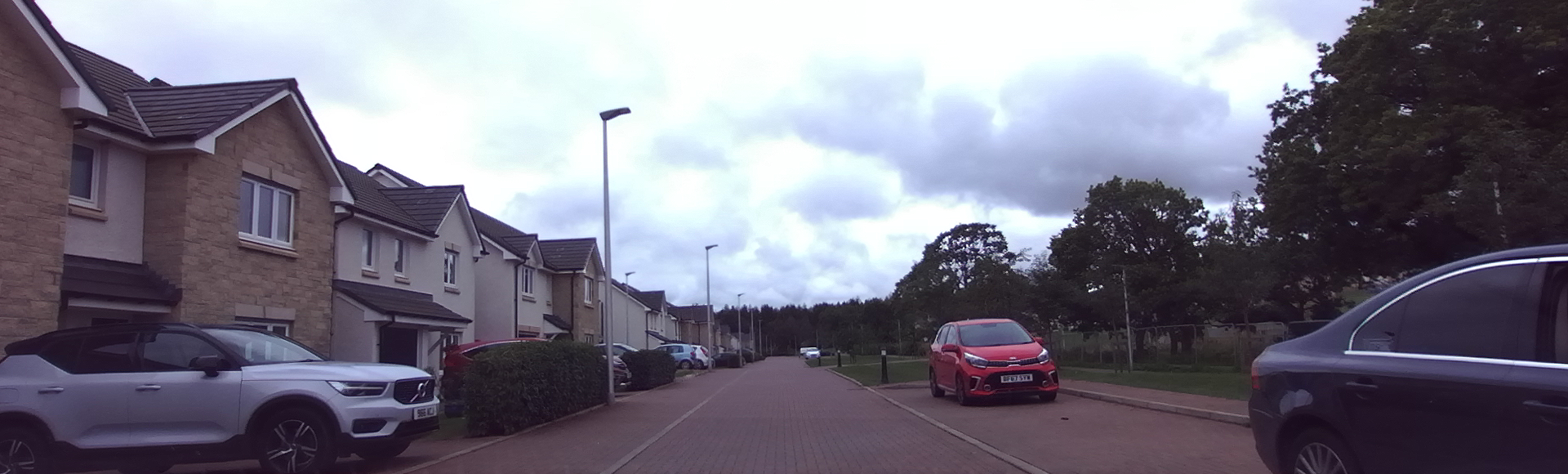} &
        \includegraphics[width=0.244\textwidth]{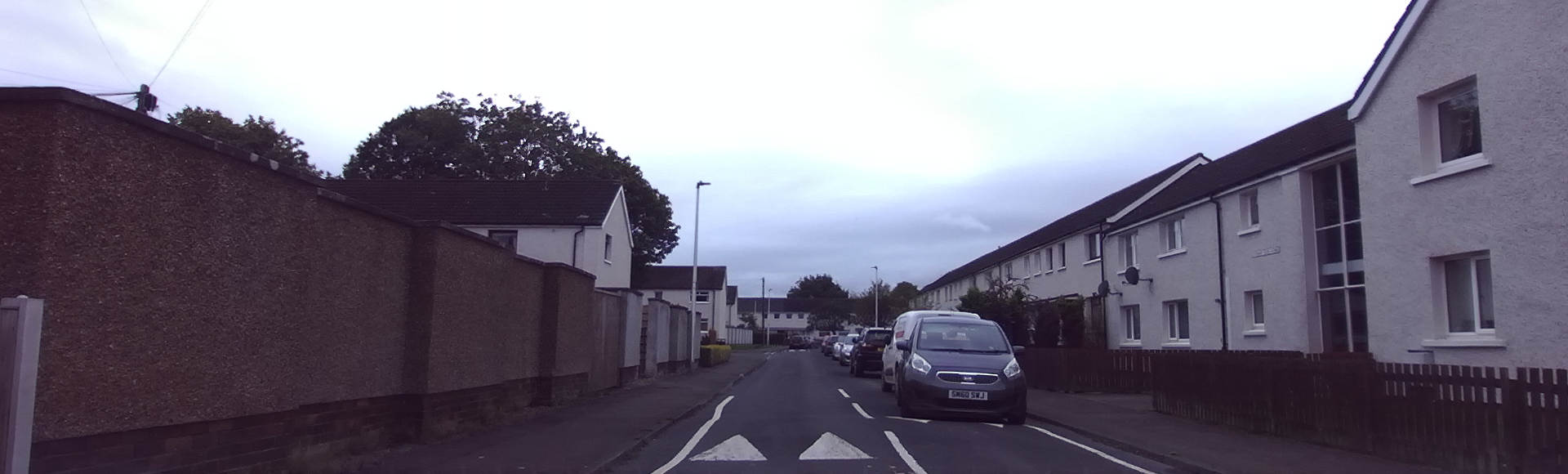} &
        \includegraphics[width=0.244\textwidth]{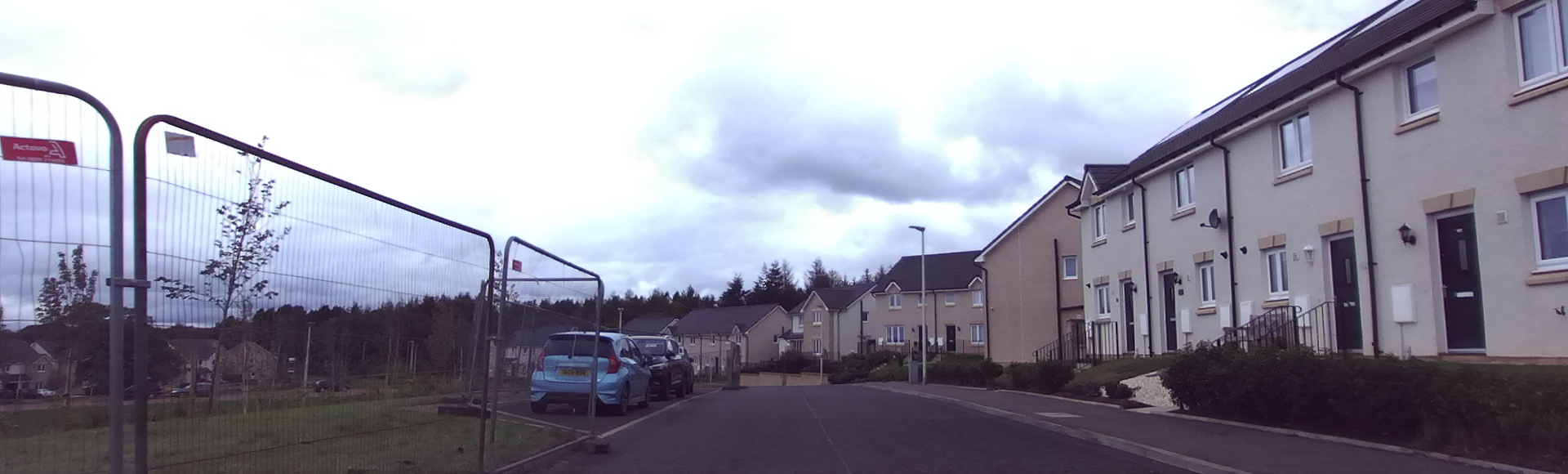} \\
        
    \end{tabular}

    \setlength{\abovecaptionskip}{5pt}
    \caption{Sample images of SDIRF.
    We show eight foggy-clear image pairs, which are grouped into four columns.
    Each column contains a thin fog situation (first row) and a thick fog situation (third row).
    The second and fourth rows show the corresponding clear images taken in overcast weather.
    }
    \setlength{\belowcaptionskip}{-10pt}
    \label{fig:sample_sdirf_images}
\end{figure*}

In total, 52 data episodes were collected.
Apart from just one episode that contains only a foggy sequence, the remaining 51 episodes comprise a foggy sequence and a counterpart clear sequence of the same route.
The total duration of the foggy and the clear videos are 2578 seconds and 2443 seconds, respectively.
Further, by visually examining the images, we subjectively classify the 52 foggy sequences into thin fog (20 sequences totalling 1101 seconds) and thick fog (32 sequences totalling 1477 seconds).
See \RefFig \ref{fig:sample_sdirf_images} for sample images.

\subsection{Photometric Calibration}    \label{subsec:photometric_calibration}

As \eqref{eq:asm_radiance} shows, the atmospheric scattering model operates in the radiance domain.
However, the ZED 2i camera features an onboard ISP and can only save the digitally post-processed intensity data, but not the raw radiance data.
Therefore, for the on-road data that we recorded, we have to infer the radiance values from the intensity values.
This process requires the photometric parameters of the camera to be known, and we achieve this by performing a photometric calibration in a controlled laboratory where the radiance values can be measured by an optical power meter.
Our calibration process focuses on recovering the parameters of gamma correction, which introduces the largest amount of non-linearity in the mapping between radiance and intensity \cite{szeliski2022computer}.
We will explain our calibration setup, the experiments we conducted, and how we infer the photometric parameters from the experimental data.

\subsubsection{Setup}
\begin{figure}[!t]
    \centering
    \setlength\tabcolsep{1pt} 
    \renewcommand{\arraystretch}{0.5}
    {\scriptsize
    \begin{tabular}{ccc}
        \includegraphics[width=0.33\linewidth]{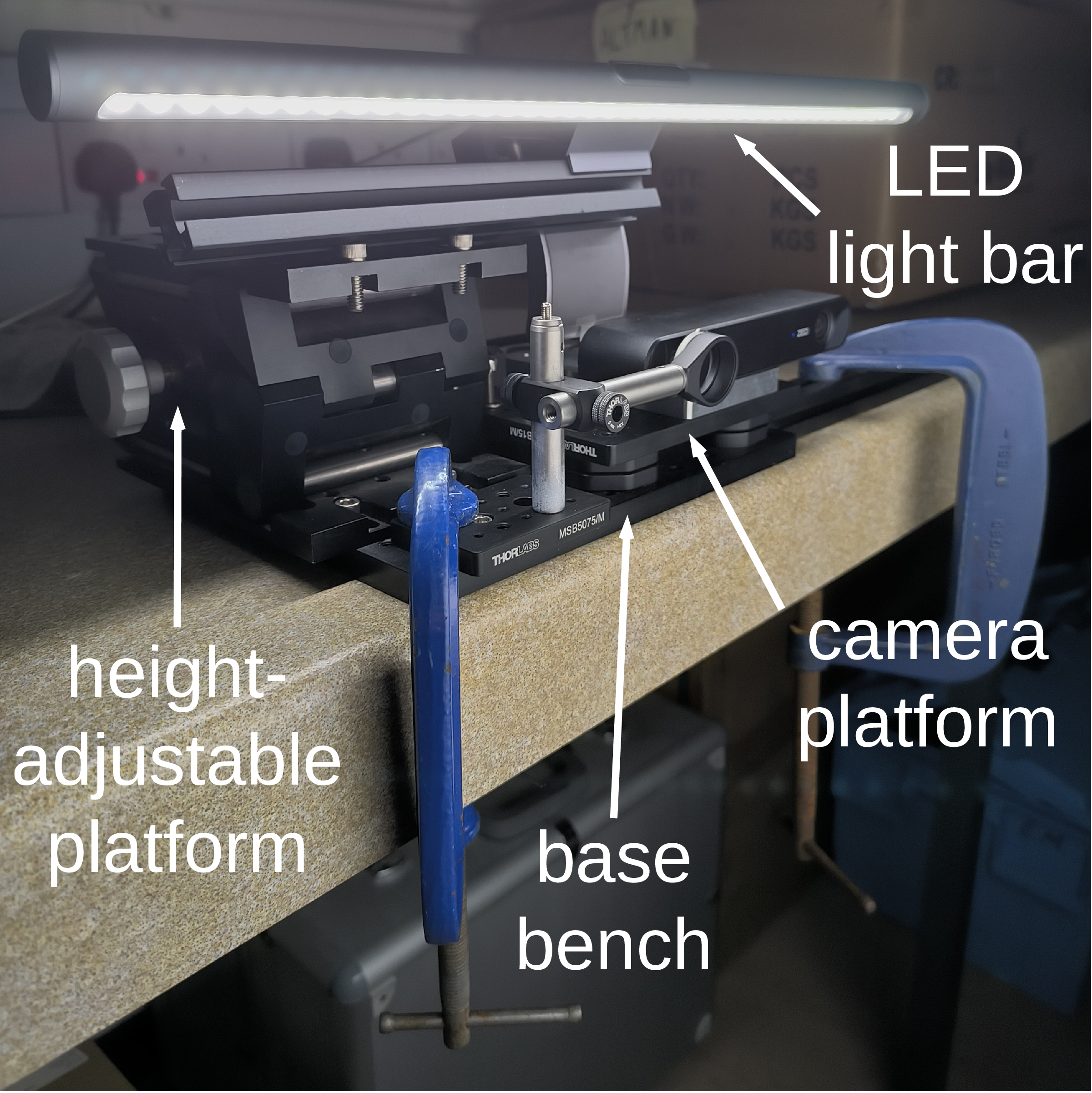} &
        \includegraphics[width=0.33\linewidth]{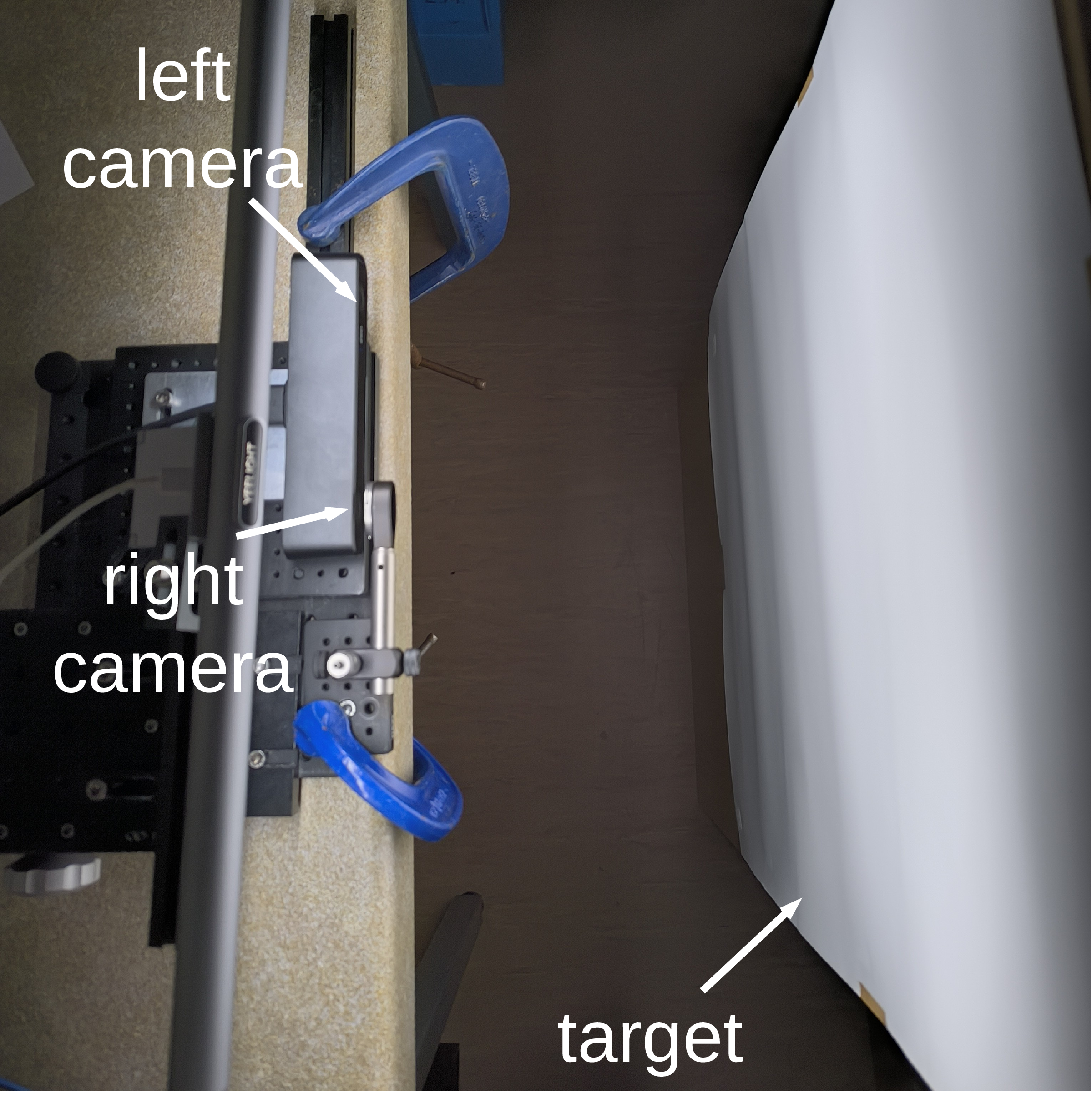} &
        \includegraphics[width=0.33\linewidth]{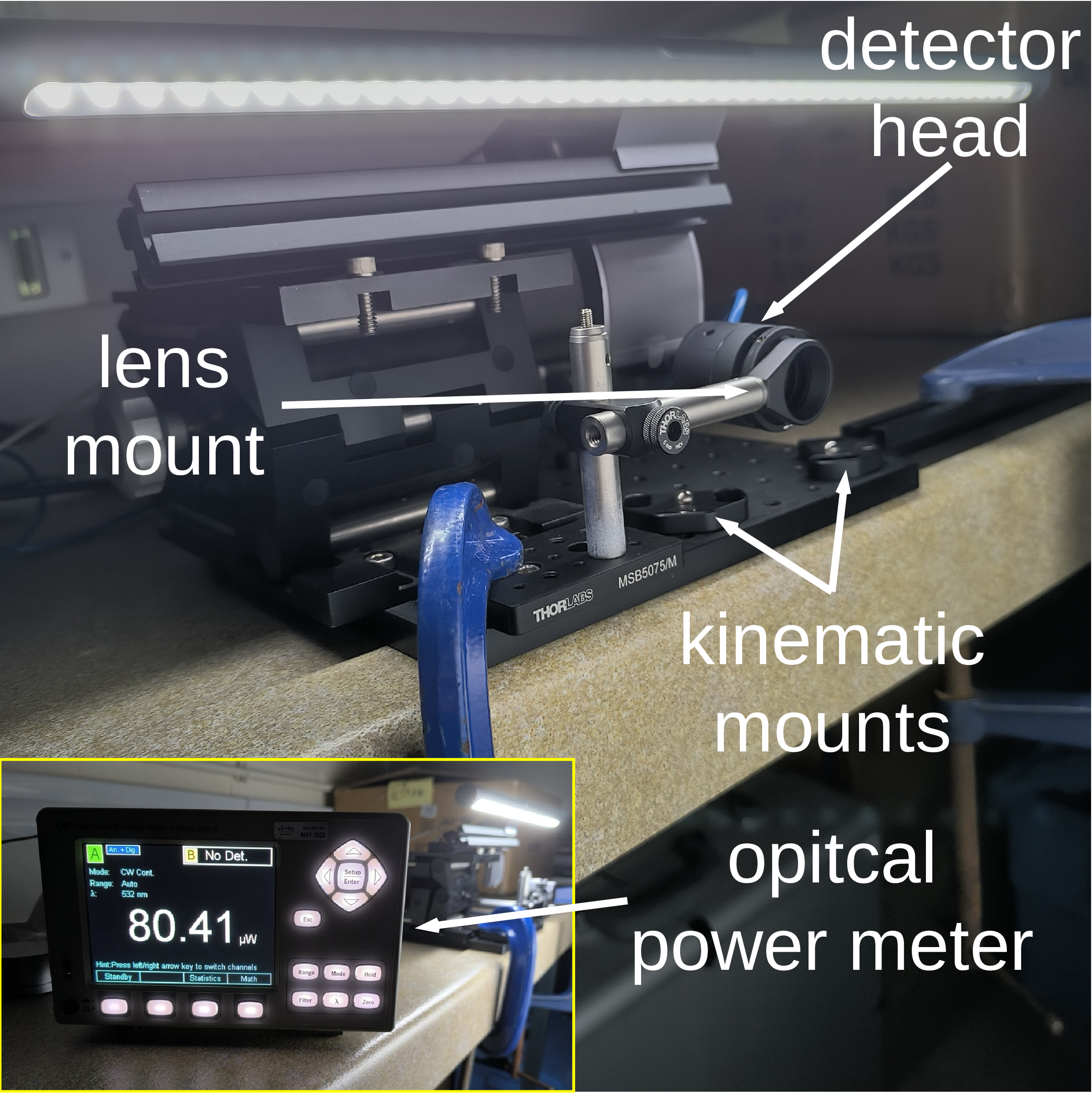} \\

        (a) & (b) & (c) \\
    \end{tabular}
    }
    
    \setlength{\abovecaptionskip}{0pt}
    \caption{Our photometric calibration setup.}
    \setlength{\belowcaptionskip}{-10pt}
    \label{fig:calibration_setup}
\end{figure}

\begin{figure*}[!t]
    \centering
    \begin{minipage}{0.63\textwidth}
        \centering
        \includegraphics[height=2.02in]{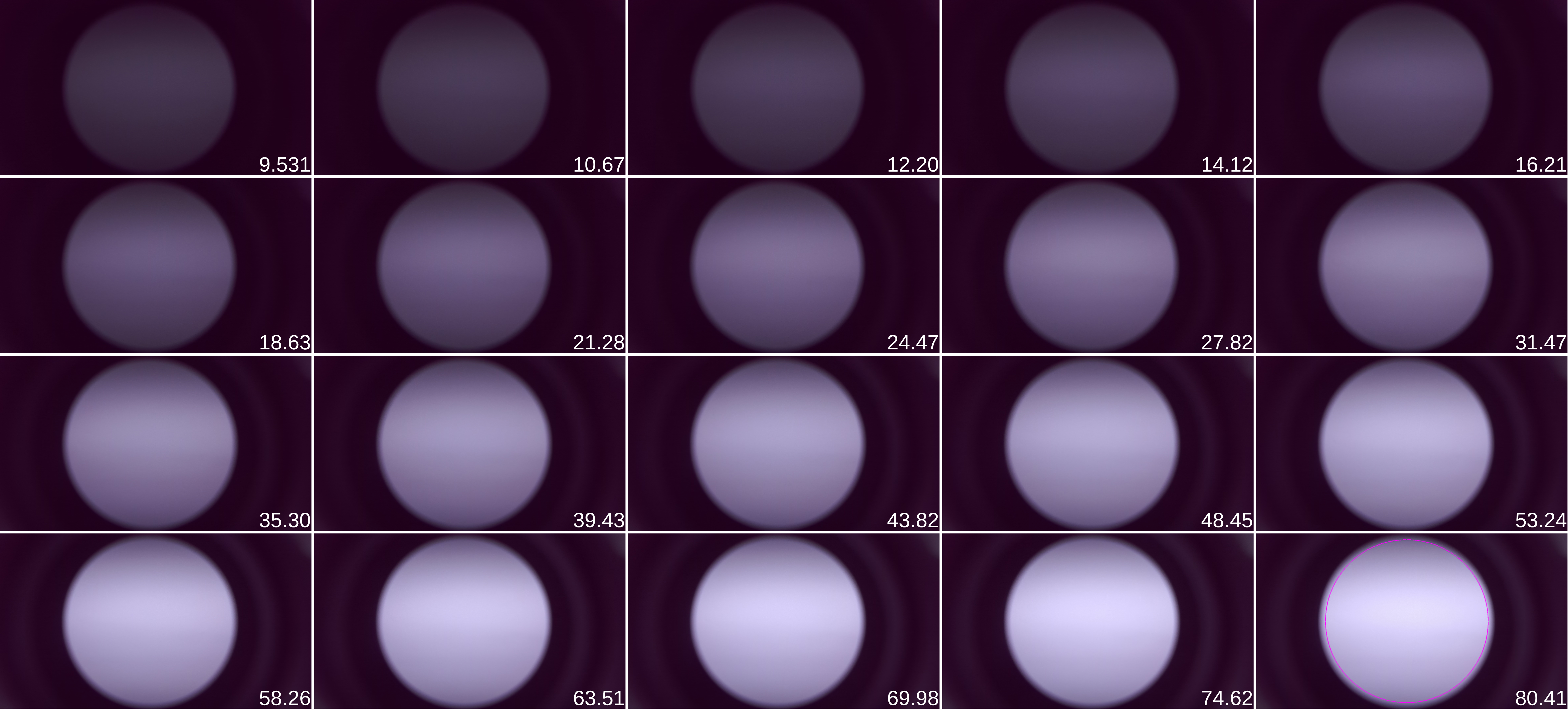}
        \caption{The right images taken at the 20 different levels of brightness, overlaid with the corresponding measured optical power (in $\mu$W).
        }
        \label{fig:calibration_photos}
    \end{minipage}%
    \hfill
    \begin{minipage}{0.35\textwidth}
        \centering
        \includegraphics[trim={.4cm .38cm .37cm .37cm},clip,height=2.02in]{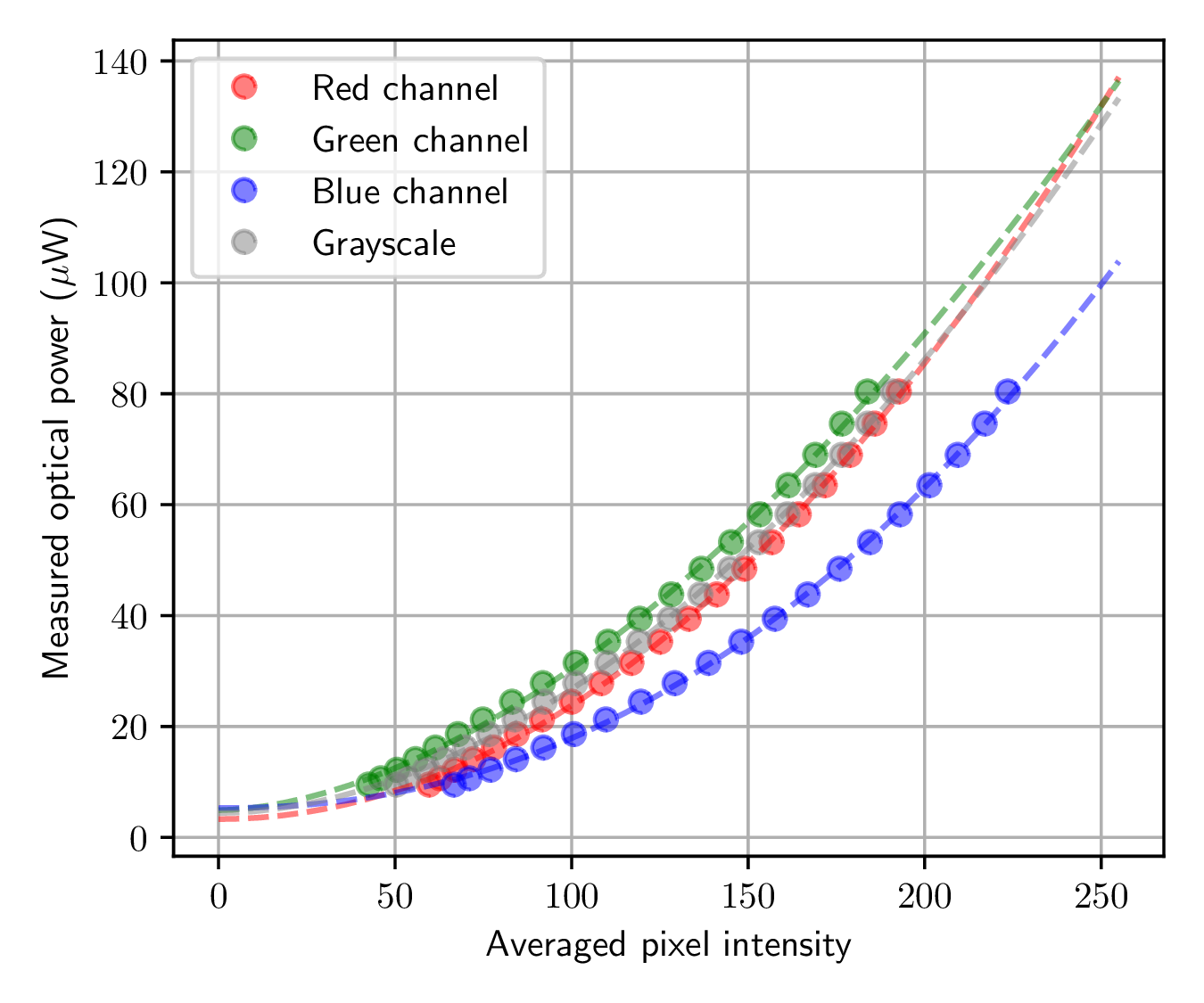}
        \caption{The data points obtained from the image series in \RefFig \ref{fig:calibration_photos} and the fitted curves.
        }
        \label{fig:calibration_fit}
    \end{minipage}
\end{figure*}

Our calibration setup is shown in \RefFig \ref{fig:calibration_setup}.
The only light source was an LED light bar with 20 different, adjustable levels of brightness.
It emitted diffused light against a big white sheet (in matte finish), which was used as the target.
The light bar was affixed to a height-adjustable platform mounted onto the base bench using screws.
The camera was mounted the right way up onto a platform which, in turn, was attached to the base bench via three kinematic mounts.
Using kinematic mounts ensured that the camera's position relevant to the base bench remained the same every time it was removed then reattached.
When attached, its image plane was roughly aligned to the sheet and the principal axis of the right camera intersected the centre of the sheet.
To measure the optical power perceived by the camera, we used an optical power meter paired with an optical power detector.
A lens mount (with no lens mounted) was affixed to the base bench and placed just in front of the right camera's position for attaching the detector head.

\subsubsection{Experiments}
We calibrated the photometric parameters for each combination of exposure time and gain that had been used when collecting the on-road data.
The rest of the camera settings were configured identically to the ones used in our data collection.
To avoid image saturation, the target was placed further away from the test bench when we used longer exposure time.
During calibration, we alternated between the following two modes.
\begin{itemize}
    \item
    Photo mode [\RefFigs \ref{fig:calibration_setup}(a) and \ref{fig:calibration_setup}(b)].
    In this mode, the detector head is removed from the lens mount, the camera's platform is attached to the base bench, and we let the camera take a stereo pair of photos of the target.
    In particular, the right camera images the target through the hole of the lens mount.
    
    \item
    Power measurement mode [\RefFig \ref{fig:calibration_setup}(c)].
    In this mode, the camera's platform is removed from the base bench, the detector head is attached to the lens mount, and we record the measured optical power reading from the power meter screen.
\end{itemize}

Fixing the exposure time and gain to a given combination, \RefFig \ref{fig:calibration_photos} shows the images of the right frame taken at the 20 different levels of brightness.

\begin{figure*}[!t]
    \centering
    \setlength\tabcolsep{3pt} 
    \renewcommand{\arraystretch}{0.5}
    {\scriptsize
    \begin{tabular}{ccc}
    \centering        
        \includegraphics[height=2.59cm]{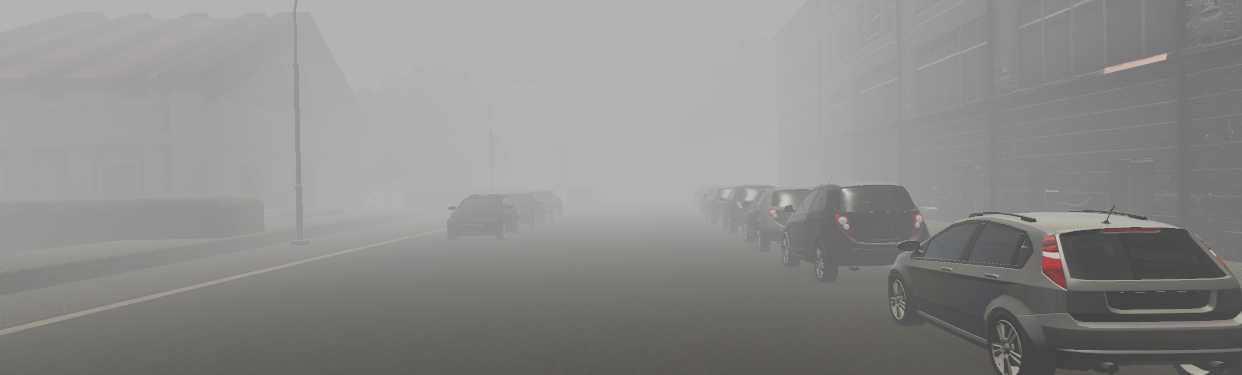} &
        \includegraphics[height=2.59cm]{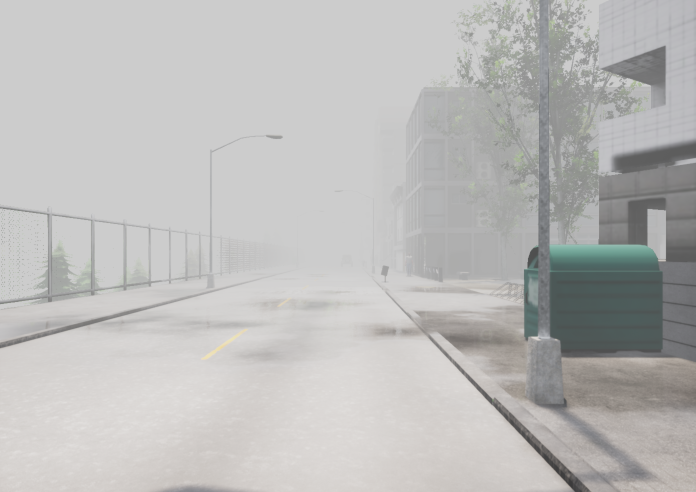} &
        \includegraphics[height=2.59cm]{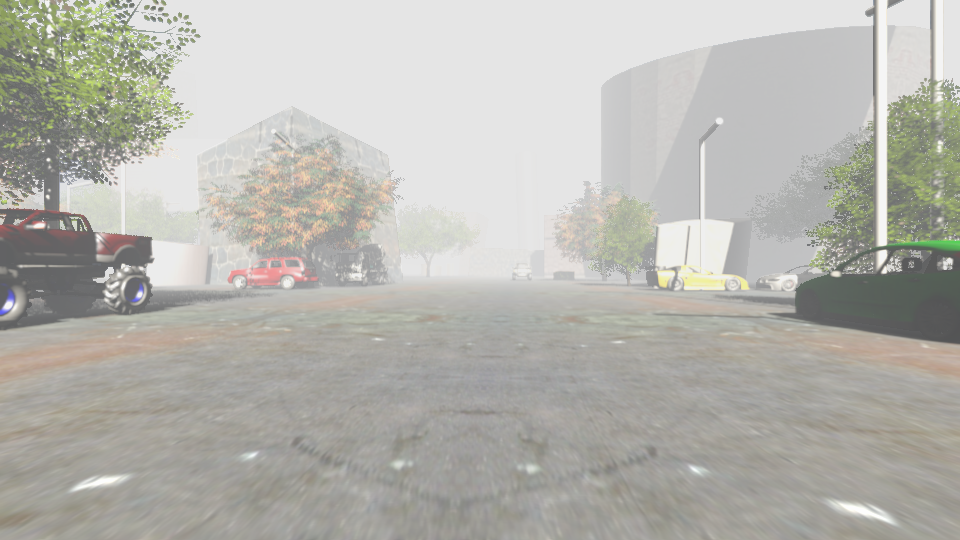} \\

        (a) & (b) & (c) \\
    \end{tabular}
    }

    \setlength{\abovecaptionskip}{3pt}
    \caption{Sample synthetic foggy images for evaluation.
    (a) VKITTI2 ($V_{\text{MOR}} = 40$ m).
    (b) KITTI-CARLA ($V_{\text{MOR}} = 60$ m).
    (c) DRIVING ($V_{\text{MOR}} = 80$ m).
    }
    \setlength{\belowcaptionskip}{-10pt}
    \label{fig:synthetic_foggy_images}
\end{figure*}

\subsubsection{Photometric Parameter Characterisation}
For each colour channel of each image, we need to associate an intensity value with each optical power measured.
We compute the intensity by averaging the pixel values of a circular area with a radius of 500 pixels (see the magenta outline in the very last image of \RefFig \ref{fig:calibration_photos}) within the target region in an image.
Given the intensity values and the corresponding optical power readings, we use least squares fitting to estimate $\alpha$, $\gamma$ and $\zeta$ in \eqref{eq:intensity_to_radiance}.

\RefFig \ref{fig:calibration_fit} plots the data points obtained from the image series in \RefFig \ref{fig:calibration_photos} and the fitted curves.
We can see that all curves bend upwards (i.e., $\gamma>1$), which is in line with the expectation of a gamma expansion.

\section{Experiments}   \label{sec:experiments}
We now describe our experiments.
After introducing the data used for evaluation and competitive methods, we describe our implementation details.
Then, we present thorough results on both synthetic and real data.
Finally, we report two additional experiments that further showcase the superiority of our method over others.  

\subsection{Data for Evaluation}    \label{sec:data_for_evaluation}
We use both synthetic and real data for evaluation.

To generate \emph{synthetic} foggy images we use the following three datasets: the Virtual KITTI 2 dataset \cite{cabon2020virtual} (VKITTI2), the KITTI-CARLA dataset \cite{deschaud2021kitticarla} and the Driving dataset (DRIVING) \cite{mayer2016large}.
They all contain sequences of left and right clear intensity images as well as the corresponding left and right ground truth depth maps.
For each clear image, we first compute a distance map from its ground truth depth map, and then synthesise a corresponding foggy image by applying \eqref{eq:asm_intensity} rather than \eqref{eq:asm_radiance} (because the camera's photometric parameters are not available) to each channel.
For all three colour channels, we fix $A$ at $255 \times 0.7 = 178.5$, $255 \times 0.8 = 204.0$ and $255 \times 0.9 = 229.5$ for VKITTI2, KITTI-CARLA and DRIVING, respectively.
These values of $A$ fall within the typical range $\left[0.7, 1\right]$ which previous work, for example \cite{li2019benchmarking, ren2020single}, extensively adopted to synthesise foggy images.
For each dataset, six different visibility levels at $V_{\text{MOR}} = \{ 30, 40, 50, 60, 70, 80 \}$ meters are tested.
The corresponding ground truth $\beta$ values are calculated according to \eqref{eq:vis}.
See \RefFig \ref{fig:synthetic_foggy_images} for sample synthetic foggy images at various visibility levels, and see our supplementary material for a summary of the synthetic datasets we use for evaluation.

For evaluation on \emph{real} foggy data we use our self-collected SDIRF dataset introduced in the previous section.

\begin{figure*}[!t]
    \centering
    \includegraphics[trim={0.1cm .4cm 0.1cm .7cm},clip,width=\linewidth]{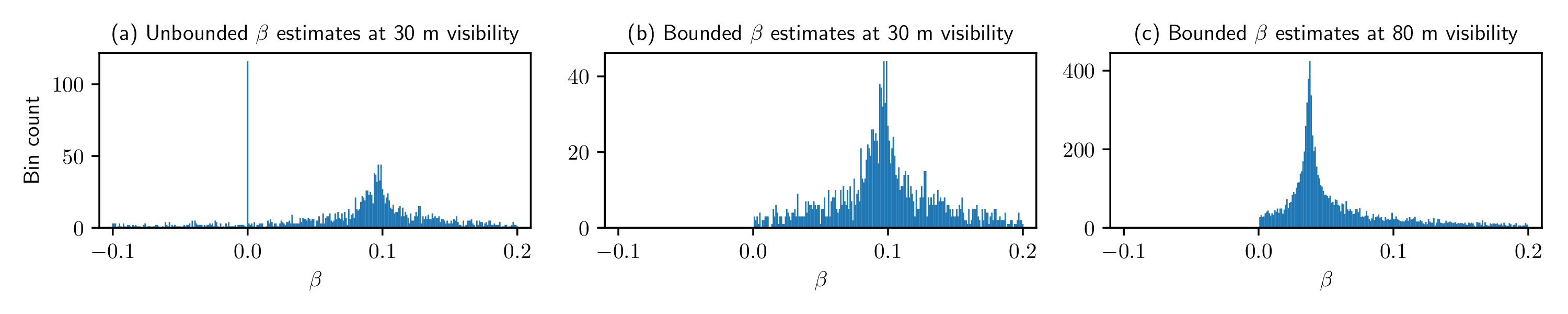} 
    \setlength{\abovecaptionskip}{-10pt}
    \caption{
    $\beta$ histogram examples generated by:
    (a) Li's method \cite{li2015simultaneous};
    (b) and (c) our modified version of it.
    Note that the vertical axes have different scales.
    (a) Unbounded $\beta$ estimates at 30 m visibility. The highest bin, which occurs at zero, leads to a wrong estimate of $\beta$.
    (b) Bounded (within the range $[0.001, 0.2]$) $\beta$ estimates at 30 m visibility. The highest bin occurs at 0.097, which is much closer to the ground truth $\beta$ value of 0.1.
    (c) Bounded (within the range $[0.001, 0.2]$) $\beta$ estimates at 80 m visibility.
    Comparing (c) to (b), we observe that the total number of $\beta$ estimates that are used to build the histogram is typically much larger at a higher visibility level.
    }
    \setlength{\belowcaptionskip}{-10pt}
    \label{fig:beta_hist}
\end{figure*}

\subsection{Competitive Methods}    \label{subsec:competitive_methods}
To the best of our knowledge, there is very limited existing work on estimating both $A$ and $\beta$.
Firstly, we report the results of Berman \textit{et al.} \cite{berman2016non} which estimates $A$ only.
We further compare our method with the fog estimation strategy proposed by Li \textit{et al.} \cite{li2015simultaneous} (estimating both $A$ and $\beta$) as well as \emph{our modified version} of that strategy, which resulted in a major improvement over the original one. 
We made two main modifications:
Firstly, as proposed by \cite{tang2014investigating}, to estimate $A$, we use the median, instead of the maximum \cite{he2010single}, of the 0.1\% pixels with the largest dark channel values.
Secondly, when building the histogram of values of $\beta$, we discard values of $\beta$ outside the range $[0.001, 0.2]$.
This is motivated by the observation, typically at lower visibility, that a proportion of $\beta$ values are negative and that there is a large cluster of $\beta$ centred at the value of zero.
In many situations, the zero bin has the highest counts, leading to a wrong estimate of $\beta$.
\RefFigs \ref{fig:beta_hist}(a) and \ref{fig:beta_hist}(b) show examples of unbounded and bounded $\beta$ histograms, respectively, at 30 m visibility.
As we will show later, this modified version greatly improves the original one's performance and therefore we deem it to be a much stronger baseline method.
Note that the above range of $\beta$ that we use in the modified version of Li's method is consistent with the bounds of $\beta$ that we set in our method (\RefSec \ref{subsubsec:param_bounds}) when solving \eqref{eq:optimisation_problem}.
This ensures a fair comparison between them.

\subsection{Implementation Details}
We empirically set $\xi_{\text{F}} = 4$ in \eqref{eq:DL}, $\xi_{\text{K}} = 15$ in \eqref{eq:xi_K}, $\eta = 2$ when determining the bounds for each relevant ${L_{\text{c}}}_n$ (\RefSec \ref{subsubsec:param_bounds}), and $\delta = 5$ (in the intensity domain) when defining the Huber loss in the first stage of our optimisation (\RefSec \ref{subsubsec:two_stage_optimisation}).
We fix these parameters throughout all our experiments.

To make a fair comparison, for all methods we use the stereo ORB-SLAM2 \cite{mur2017orb} to facilitate multiple observations of the same landmark from a range of known distances.
These observations are established from ORB-SLAM2's local key frames and local map points after its local bundle adjustment.
The fog parameters are updated after ORB-SLAM2's local mapping thread only after the ego-vehicle has moved at least five meters from the origin (for the very first estimation) or from the position of the last update (for subsequent estimations).
We use the Ceres Solver \cite{Agarwal_Ceres_Solver_2022} and choose the Levenberg–Marquardt algorithm to solve \eqref{eq:optimisation_problem}.
See our supplementary material for more implementation details including pseudo-code.

Due to the way we generate synthetic foggy images, we have assumed that both $g$ and $g^{-1}$ are identity mappings for all colour channels.
For real foggy data from SDIRF, in contrast, $g$ and $g^{-1}$ are characterised by the parameters $\alpha$, $\gamma$ and $\zeta$ found during our photometric calibration and therefore are channel-specific.
Unless otherwise specified, the fog parameter results we report in the rest of this section are from the \emph{grayscale} foggy images.

\subsection{Evaluation on Synthetic Data}
We conduct extensive experiments and present both quantitative and qualitative results.

\subsubsection{Quantitative Results}    \label{subsubsec:synthetic_quantitative_results}
We compute the root-mean-square error (RMSE), the mean-absolute error (MAE) and the standard deviation (SD), in both absolute and relative\footnote{A relative metric is calculated as the ratio of the corresponding absolute metric to the ground truth value, and is shown as percentage in \RefTab \ref{tab:results} and \RefFig \ref{fig:RMSE_vs_vis}.} scales, of the $\beta$ and $A$\footnote{$A$ and $L_{\infty}$ are essentially the same for synthetic data since $g$ and $g^{-1}$ have been defined as identity mappings.} estimates.
\RefTab \ref{tab:results} shows the quantitative results of the average $\beta$ and $A$ error metrics on synthetic datasets\footnote{Result of Town04 in KITTI-CARLA at 30 m visibility is excluded as the ORB-SLAM2 loses tracking and provides no valid observation.}.

We observe that in most cases our method performs the best, in terms of both estimation accuracy (i.e., smallest errors) and precision (i.e., the lowest standard deviation).
The \emph{only} exception is VKITTI2's $A$ error metrics, but our $\beta$ metrics are still the best in this case.
A closer look at VKITTI2's results shows that our bad estimates of $A$ stem from countryside scenes with sparse features (Scene02), or when the ego-vehicle is surrounded by other vehicles moving at similar speeds (Scene18).
In either case the ORB-SLAM2's performance has been significantly degraded and therefore produces unreliable distance and/or intensity information.
We will investigate how to address this limitation in our future work.

\begin{table*}[!t]
    \caption{Average $\beta$ and $A$ Error Metrics on Synthetic Datasets.
    Relative Metrics are Shown as Percentage.    
    For all the Metrics, the Lower the Better ($\downarrow$).
    The Table also Contains the Results on our Ablation Study using KITTI-CARLA.
    }
    \setlength{\belowcaptionskip}{-10pt}
    \centering
    \setlength\tabcolsep{4.5pt} 
    {\footnotesize
        \begin{tabular}[m{'width'}]{| l | l | r | r | r | r | r | r | r | r | r | r | r | r |}
            \hline
            \multirow{3}{*}{Dataset} & \multirow{3}{*}{Method} & \multicolumn{6}{c|}{\cellcolor{cyan!20}$\beta$} & \multicolumn{6}{c|}{\cellcolor{blue!25}$A$} \\
            \cline{3-14}
             & & \multicolumn{2}{c|}{RMSE ($\downarrow$)} & \multicolumn{2}{c|}{MAE ($\downarrow$)} & \multicolumn{2}{c|}{SD ($\downarrow$)} & \multicolumn{2}{c|}{RMSE ($\downarrow$)} & \multicolumn{2}{c|}{MAE ($\downarrow$)} & \multicolumn{2}{c|}{SD ($\downarrow$)} \\
            \cline{3-14}
             & & \multicolumn{1}{c|}{Abs.} & \multicolumn{1}{c|}{Rel.} & \multicolumn{1}{c|}{Abs.} & \multicolumn{1}{c|}{Rel.} & \multicolumn{1}{c|}{Abs.} & \multicolumn{1}{c|}{Rel.} & \multicolumn{1}{c|}{Abs.} & \multicolumn{1}{c|}{Rel.} & \multicolumn{1}{c|}{Abs.} & \multicolumn{1}{c|}{Rel.} & \multicolumn{1}{c|}{Abs.} & \multicolumn{1}{c|}{Rel.} \\
            \hline
            \hline

            \multirow{4}{*}{VKITTI2 \cite{cabon2020virtual}} & Berman's \cite{berman2016non} & \multicolumn{2}{c|}{N/A} & \multicolumn{2}{c|}{N/A} & \multicolumn{2}{c|}{N/A} & 4.7547 & 2.66 & 3.7004 & 2.07 & 2.9130 & 1.63 \\
            \cline{2-14}
             & Li's \cite{li2015simultaneous} & 0.0430 & 66.39 & 0.0347 & 52.17 & 0.0225 & 37.17 & 3.8760 & 2.17 & 1.8920 & 1.06 & 3.3408 & 1.87 \\
            \cline{2-14}
             & Li's modified & 0.0113 & 16.74 & 0.0080 & 11.47 & 0.0091 & 14.07 & \textbf{1.5921} & \textbf{0.89} & \textbf{0.8374} & \textbf{0.47} & \textbf{1.2550} & \textbf{0.70} \\
            \cline{2-14}
             & \textbf{Ours} & \textbf{0.0078} & \textbf{10.91} & \textbf{0.0061} & \textbf{8.57} & \textbf{0.0061} & \textbf{8.60} & 2.5495 & 1.43 & 1.9276 & 1.08 & 1.8575 & 1.04 \\
            \hline
            \hline

            \multirow{10}{*}{KITTI-CARLA \cite{deschaud2021kitticarla}} & Berman's \cite{berman2016non} & \multicolumn{2}{c|}{N/A} & \multicolumn{2}{c|}{N/A} & \multicolumn{2}{c|}{N/A} & 12.6372 & 6.19 & 12.3382 & 6.05 & 2.6533 & 1.30 \\
            \cline{2-14}
             & Li's \cite{li2015simultaneous} & 0.0499 & 84.01 & 0.0452 & 75.84 & 0.0199 & 34.66 & 15.9104 & 7.80 & 15.1337 & 7.42 & 4.3683 & 2.14 \\
            \cline{2-14}
             & Li's modified & 0.0166 & 28.41 & 0.0143 & 24.35 & 0.0105 & 18.29 & 10.3676 & 5.08 & 9.6282 & 4.72 & 3.3599 & 1.65 \\
            \cline{2-14}
             & \textbf{Ours} & \textbf{0.0116} & \textbf{20.66} & \textbf{0.0101} & \textbf{18.11} & \textbf{0.0058} & \textbf{10.33} & \textbf{2.5711} & \textbf{1.26} & \textbf{1.8632} & \textbf{0.91} & \textbf{2.1219} & \textbf{1.04} \\
            \cline{2-14}
            \multicolumn{4}{|c}{\phantom{x}} \\
            \cline{2-14}

             & Li's \cite{li2015simultaneous} (GT $A$) & 0.0357 & 59.57 & 0.0250 & 41.48 & 0.0297 & 50.50 & \multicolumn{2}{c|}{-} & \multicolumn{2}{c|}{-} & \multicolumn{2}{c|}{-} \\
            \cline{2-14}
             & Li's modified (GT $A$) & 0.0109 & 19.24 & 0.0081 & 14.47 & 0.0087 & 15.37 & \multicolumn{2}{c|}{-} & \multicolumn{2}{c|}{-} & \multicolumn{2}{c|}{-} \\
            \cline{2-14}
             & Ours (GT $A$) & 0.0099 & 17.76 & 0.0087 & 15.57 & 0.0055 & 9.74 & \multicolumn{2}{c|}{-} & \multicolumn{2}{c|}{-} & \multicolumn{2}{c|}{-} \\
            \cline{2-14}
            \multicolumn{4}{|c}{\phantom{x}} \\
            \cline{2-14}

            & Ours (One-stage) & 0.0122 & 21.27 & 0.0108 & 18.85 & 0.0063 & 10.81 & 3.4662 & 1.70 & 2.5064 & 1.23 & 2.8686 & 1.41 \\
            \cline{2-14}
            & Ours (Uniform weight) & 0.0122 & 21.67 & 0.0107 & 19.05 & 0.0061 & 10.66 & 2.9282 & 1.44 & 2.2646 & 1.11 & 2.2818 & 1.12 \\
            \hline
            \hline

            \multirow{4}{*}{DRIVING \cite{mayer2016large}} & Berman's \cite{berman2016non} & \multicolumn{2}{c|}{N/A} & \multicolumn{2}{c|}{N/A} & \multicolumn{2}{c|}{N/A} & 17.6183 & 7.68 & 11.6009 & 5.05 & 15.5607 & 6.78 \\
            \cline{2-14}
             & Li's \cite{li2015simultaneous} & 0.0465 & 75.56 & 0.0387 & 62.39 & 0.0260 & 43.19 & 14.0963 & 6.14 & 11.2103 & 4.88 & 8.5101 & 3.71 \\
            \cline{2-14}
             & Li's modified & 0.0168 & 26.85 & 0.0117 & 18.99 & 0.0129 & 20.56 & 12.9722 & 5.65 & 9.5988 & 4.18 & 8.7345 & 3.81 \\
            \cline{2-14}
             & \textbf{Ours} & \textbf{0.0051} & \textbf{8.98} & \textbf{0.0037} & \textbf{6.44} & \textbf{0.0033} & \textbf{6.66} & \textbf{1.9021} & \textbf{0.83} & \textbf{1.3379} & \textbf{0.58} & \textbf{1.6629} & \textbf{0.72} \\
            \hline

        \end{tabular}
    }
    \label{tab:results}
\end{table*}

\begin{figure*}
    \centering

    \includegraphics[trim={.38cm 1.2cm .42cm 1.0cm},clip,width=\linewidth]{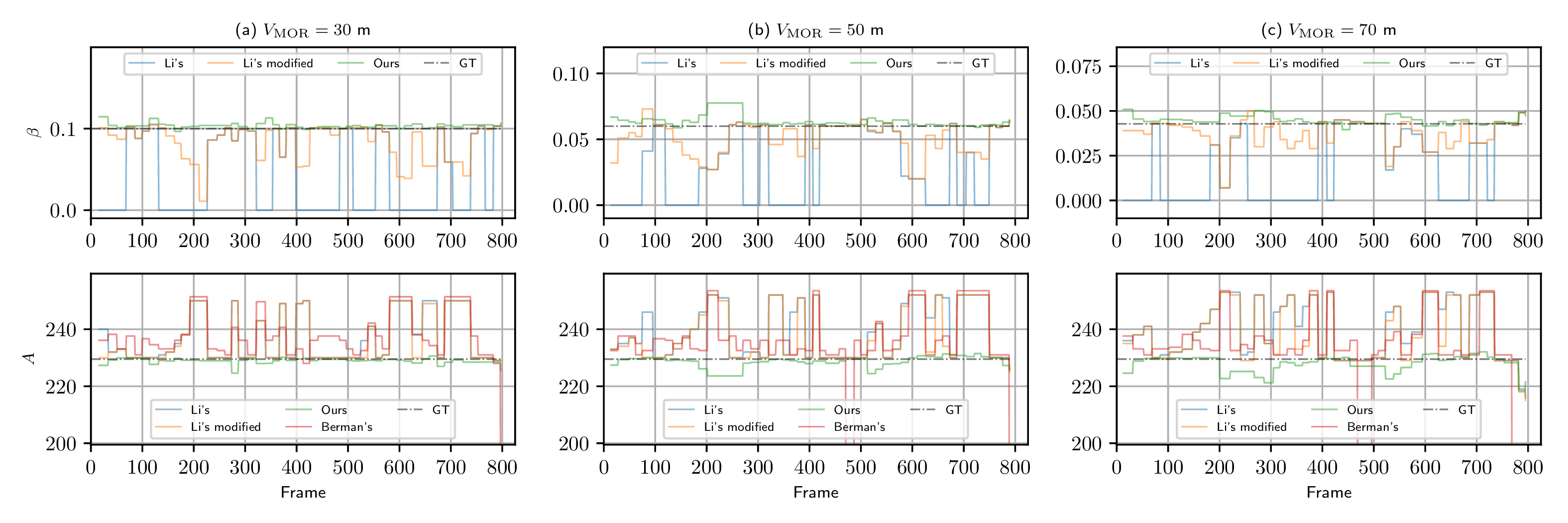}
    \setlength{\abovecaptionskip}{-10pt}
    \caption{Evaluating $\beta$ and $A$ estimates vs frame on scene ``backwards'' in DRIVING at various visibility levels.
    (a) 30 m.
    (b) 50 m.
    (c) 70 m.
    Ground truth values are indicated by black dotted lines.
    We observe that our modified version of Li's method improves the original one's performance slightly in the estimation of $A$ and significantly in the estimation of $\beta$.
    Nevertheless, both of them, as well as Berman's method, result in many large errors.
    The performance of our method surpasses the rest by a large margin.
    We also highlight that how an error in the estimate of $A$ propagates to the estimate of $\beta$ in the results of Li's method and Li's modified method [an underestimate of $\beta$, for example around frame 200 in (a), occurs with an overestimate of $A$] is in line with the theoretical analysis of error propagation that we provide in our supplementary material.}
    \label{fig:driving}
\end{figure*}

\subsubsection{Qualitative Results}
\RefFig \ref{fig:driving} illustrates how the estimates of $\beta$ and $A$ vary with frame at various visibility levels on scene ``backwards'' in DRIVING.

We observe that the values of $\beta$ and $A$ estimated by our method are closer to the ground truth values and more stable compared to other methods.

\subsubsection{Error Metrics given Partial Ground Truth}
We test the estimation performance of $\beta$ on KITTI-CARLA given the ground truth value of $A$.
The quantitative results are shown in the middle block of rows in the middle subtable of \RefTab \ref{tab:results}.

We observe:
a) As expected, all methods perform better when ground truth $A$ is given;
b) For Li's and Li's modified methods, there is a significant improvement in $\beta$'s error metrics when the ground truth $A$ is given. This is not surprising due to their sequential estimation strategy, since an error-free $A$ will indeed benefit the subsequent estimation of $\beta$. This observation adds to the evidence that, in their method, any error in the estimate of $A$ can propagate to the estimate of $\beta$;
c) For our method, such improvement is much less significant. This may suggest that our method, when simultaneously optimising $\beta$ and $A$ with minimal prior knowledge, is able to find a $\beta$ value that is not far from the optimal solution.

\subsubsection{Ablation Study}  \label{subsubsec:additional_experiments}
We conduct an ablation study on KITTI-CARLA to better understand how our optimisation setup affects the performance of the fog parameter estimation.
The following additional settings are experimented with:
a) One-stage: Only the first stage of our optimisation is preserved;
b) Uniform weight: We set $w_n^m = 1$ for all observations in both optimisation stages.
The quantitative results are shown in the bottom block of rows in the middle subtable of \RefTab \ref{tab:results}.

We observe:
a) If one-stage optimisation is performed or a uniform weight is used, the estimation results are inferior to those produced by our full method;
b) These two settings still outperform all competitive methods, despite trailing behind our full method.

\begin{figure}[!t]
    \centering
    \setlength\tabcolsep{1pt} 
    \renewcommand{\arraystretch}{0.5}
    \begin{tabular}{cc}
        \centering
        \includegraphics[trim={.3cm .4cm .2cm .95cm},clip,width=.5\linewidth]{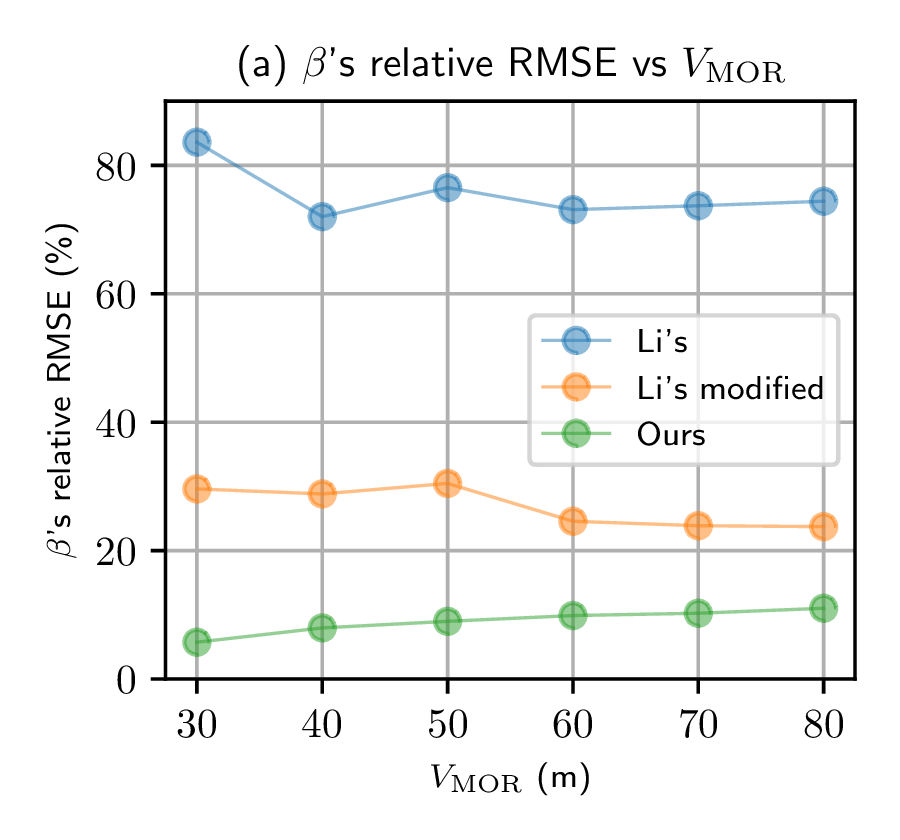} &
        \includegraphics[trim={.3cm .4cm .2cm .95cm},clip,width=.5\linewidth]{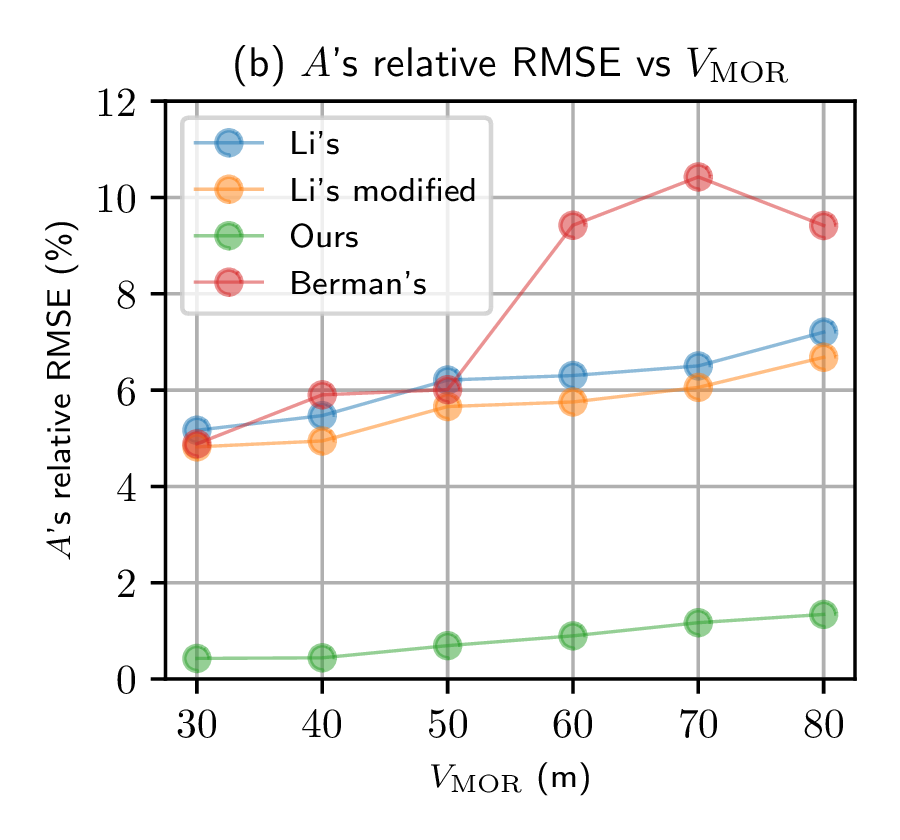} \\
    \end{tabular}
    \setlength{\abovecaptionskip}{0pt}
    \caption{Evaluating the relative RMSE vs $V_{\text{MOR}}$ on DRIVING.
    (a) $\beta$.
    (b) $A$.
    }
    \setlength{\belowcaptionskip}{-10pt}
    \label{fig:RMSE_vs_vis}
\end{figure}

\begin{figure*}[!t]
    \centering
    \setlength\tabcolsep{1pt} 
    \renewcommand{\arraystretch}{0.5}
    \begin{tabular}{ccccc}
        \centering

        &
        \tiny{(a)} &
        \tiny{(b)} &
        \tiny{(c)} &
        \tiny{(d)} \\

        \rotatebox[origin=lc]{90}{\qquad \; \; \tiny{Li's \cite{li2015simultaneous}}} &
        \includegraphics[trim={0.15cm 0.2cm 0.1cm 0.1cm},width=0.244\textwidth]{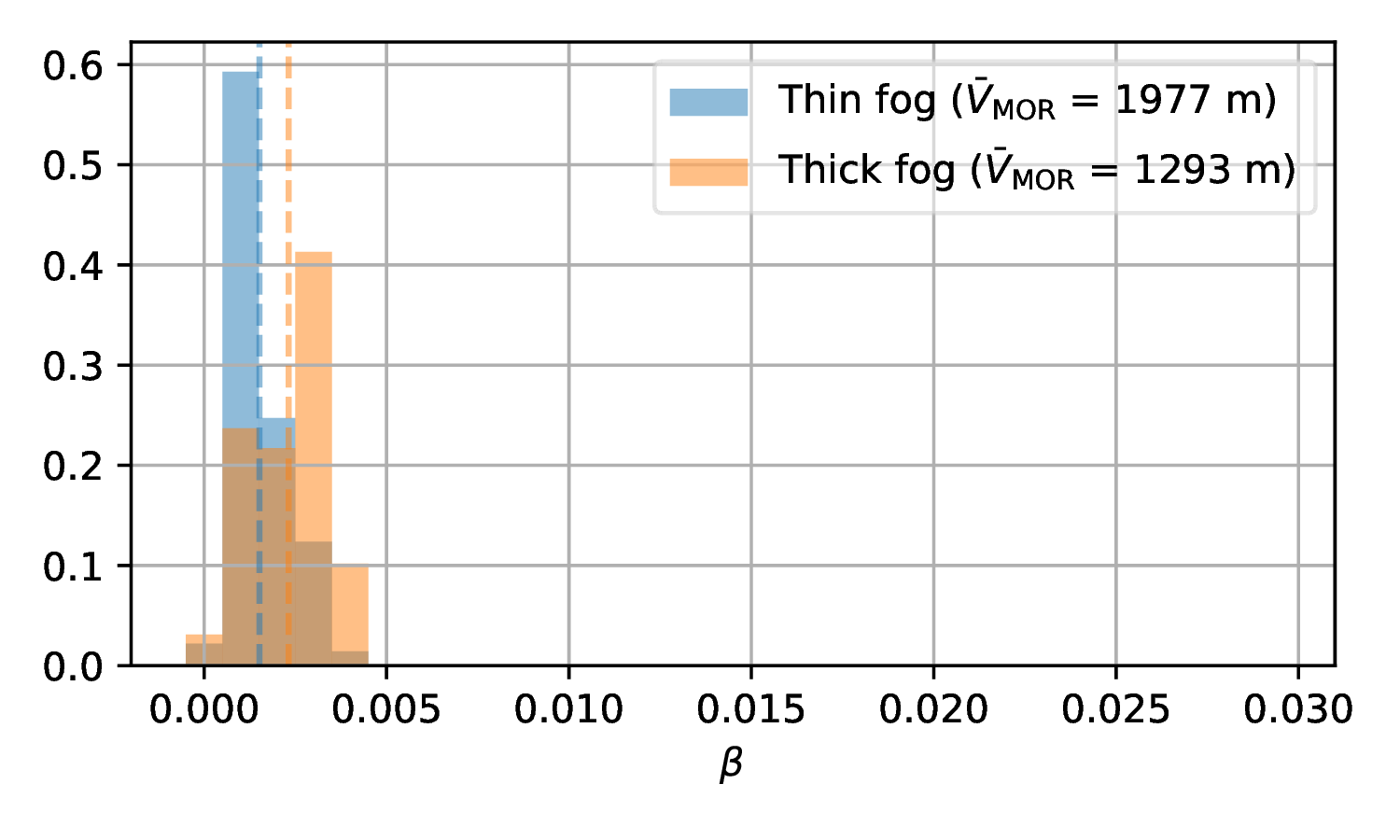} &
        \includegraphics[trim={0.15cm 0.2cm 0.1cm 0.1cm},width=0.244\textwidth]{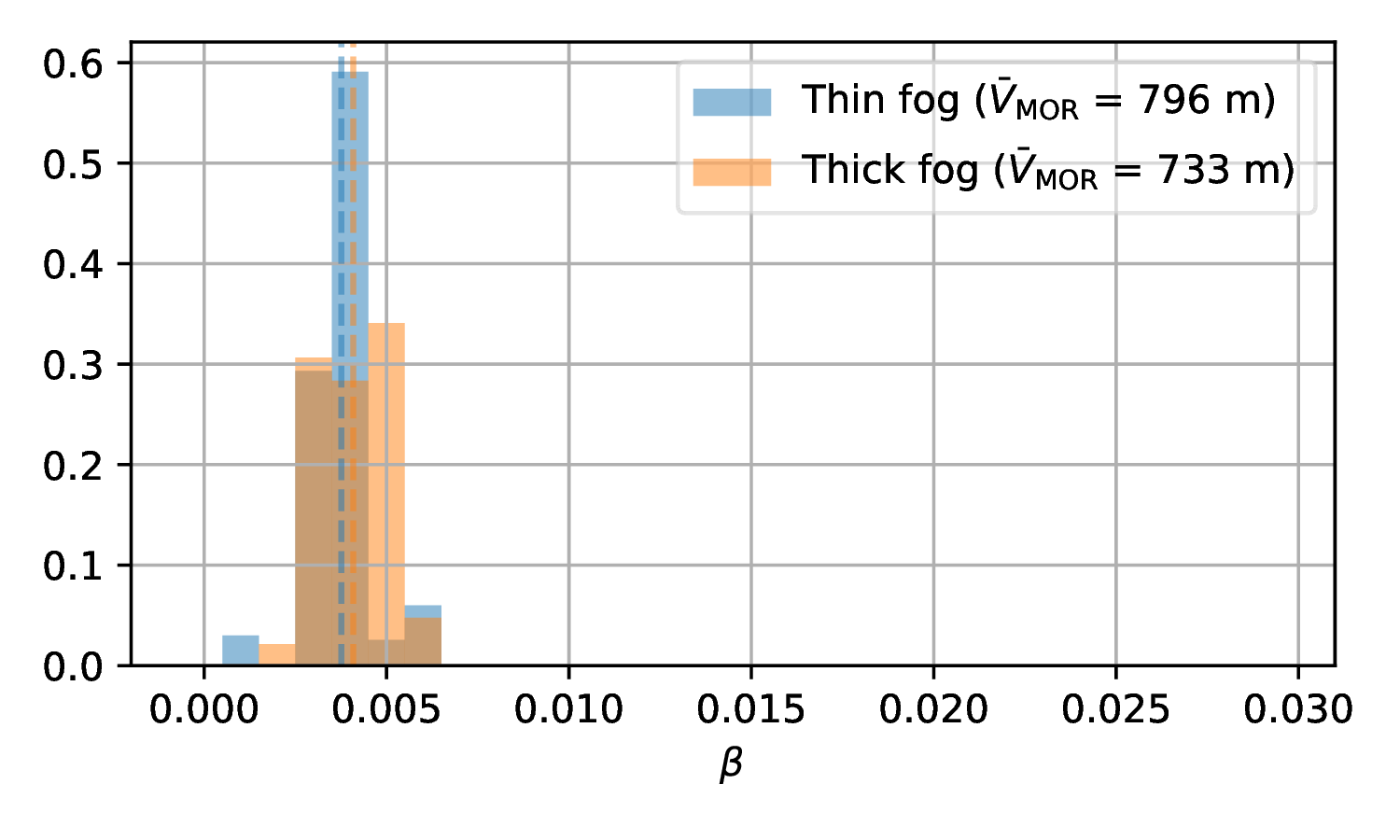} &
        \includegraphics[trim={0.15cm 0.2cm 0.1cm 0.1cm},width=0.244\textwidth]{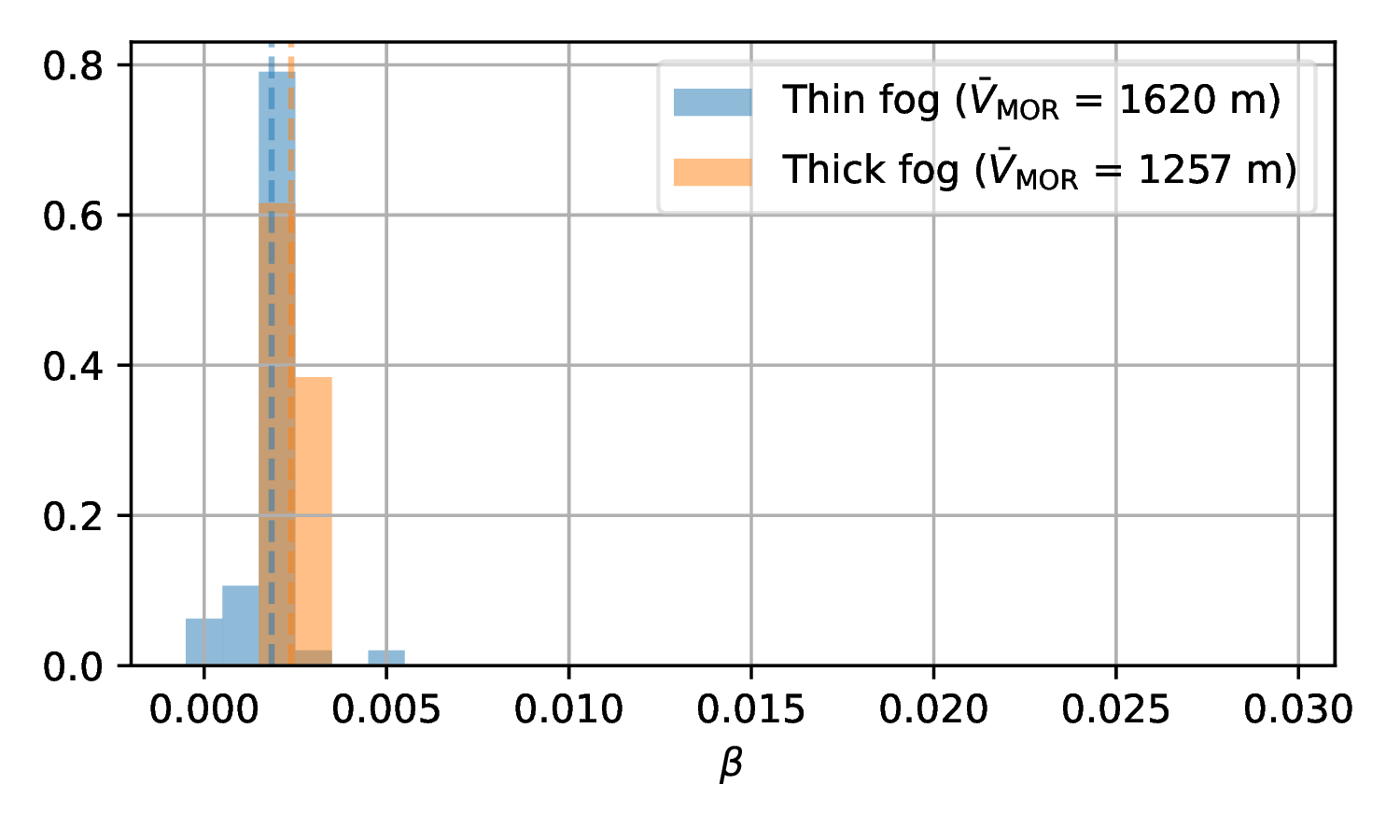} &
        \includegraphics[trim={0.15cm 0.2cm 0.1cm 0.1cm},width=0.244\textwidth]{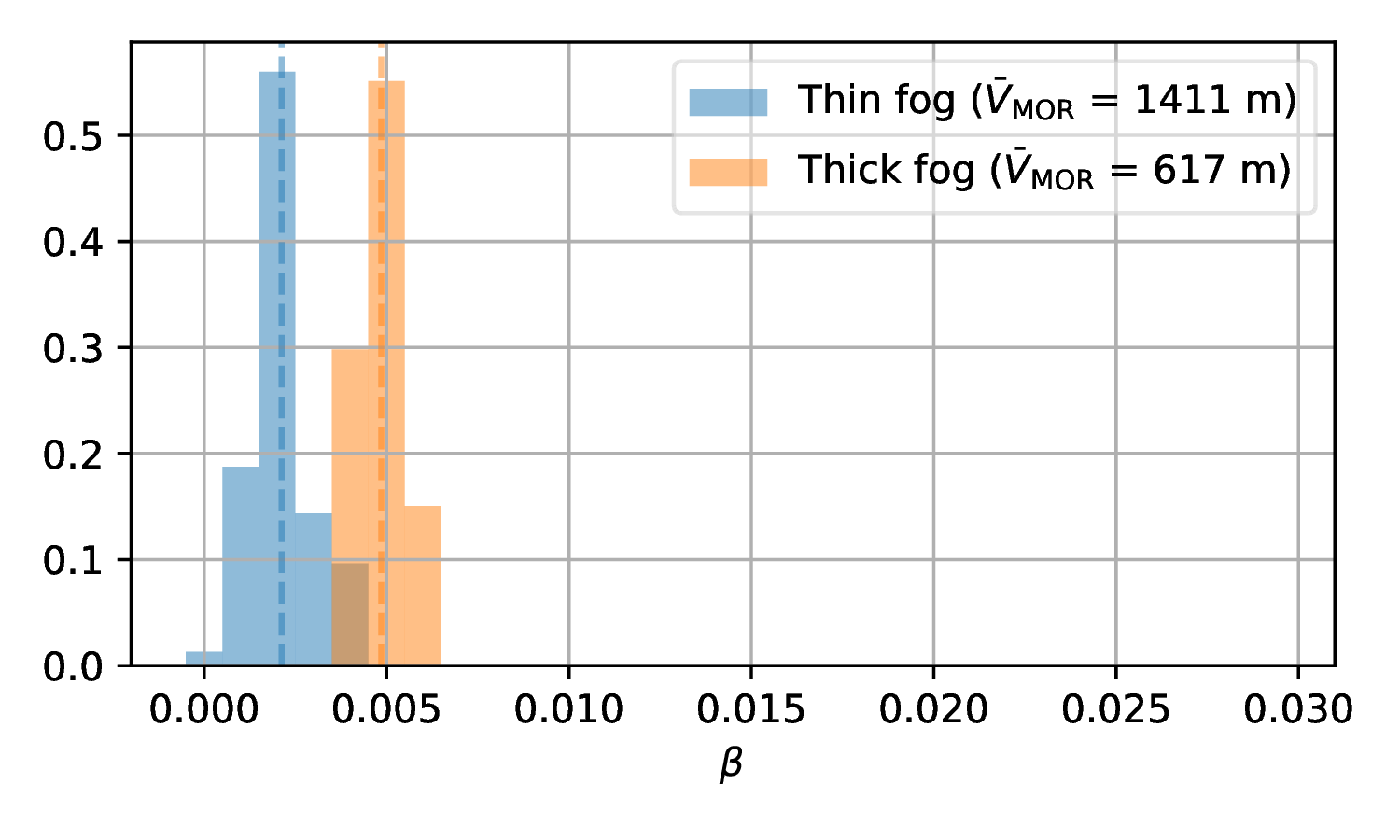} \\

        \rotatebox[origin=lc]{90}{\qquad \, \tiny{Li's modified}} &
        \includegraphics[trim={0.15cm 0.2cm 0.1cm 0.1cm},width=0.244\textwidth]{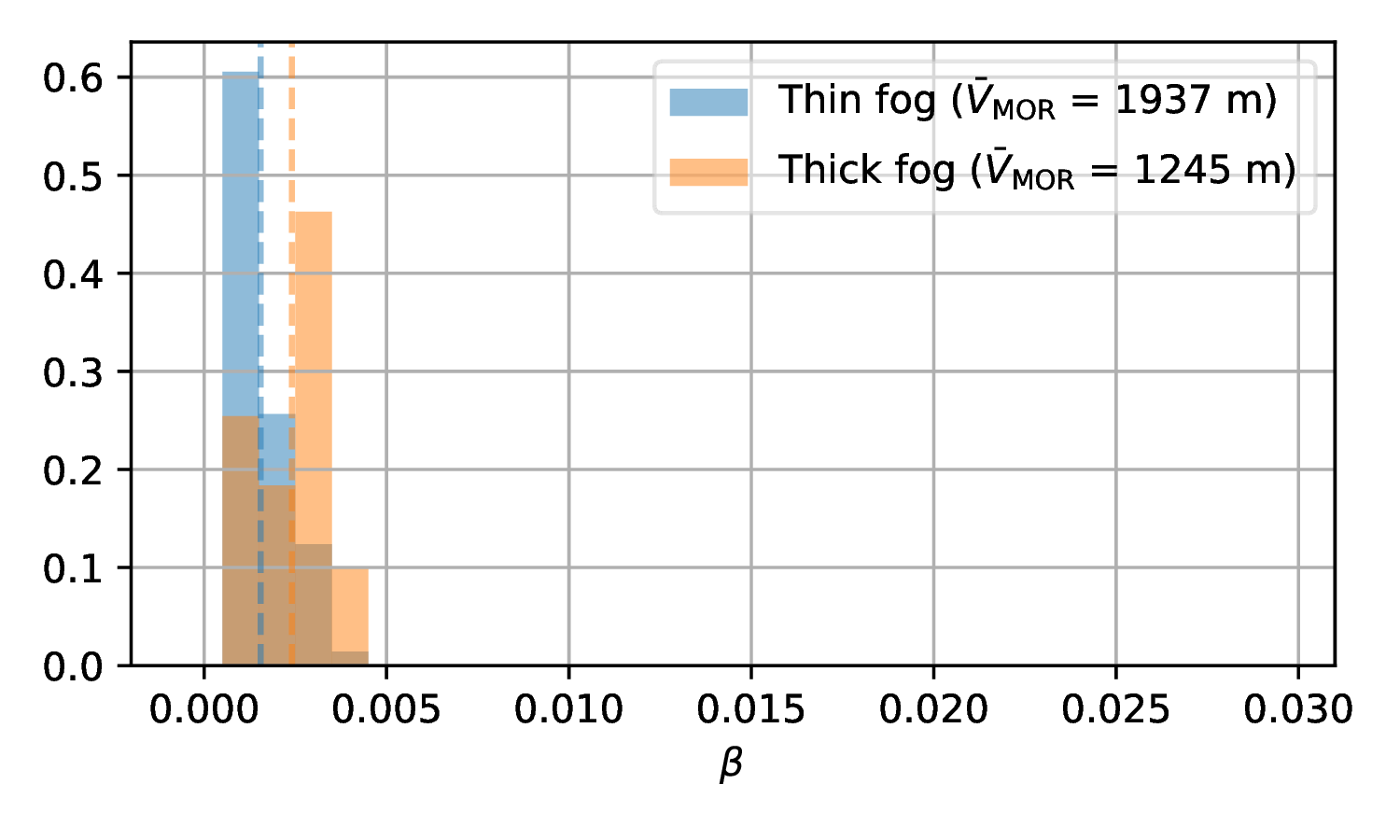} &
        \includegraphics[trim={0.15cm 0.2cm 0.1cm 0.1cm},width=0.244\textwidth]{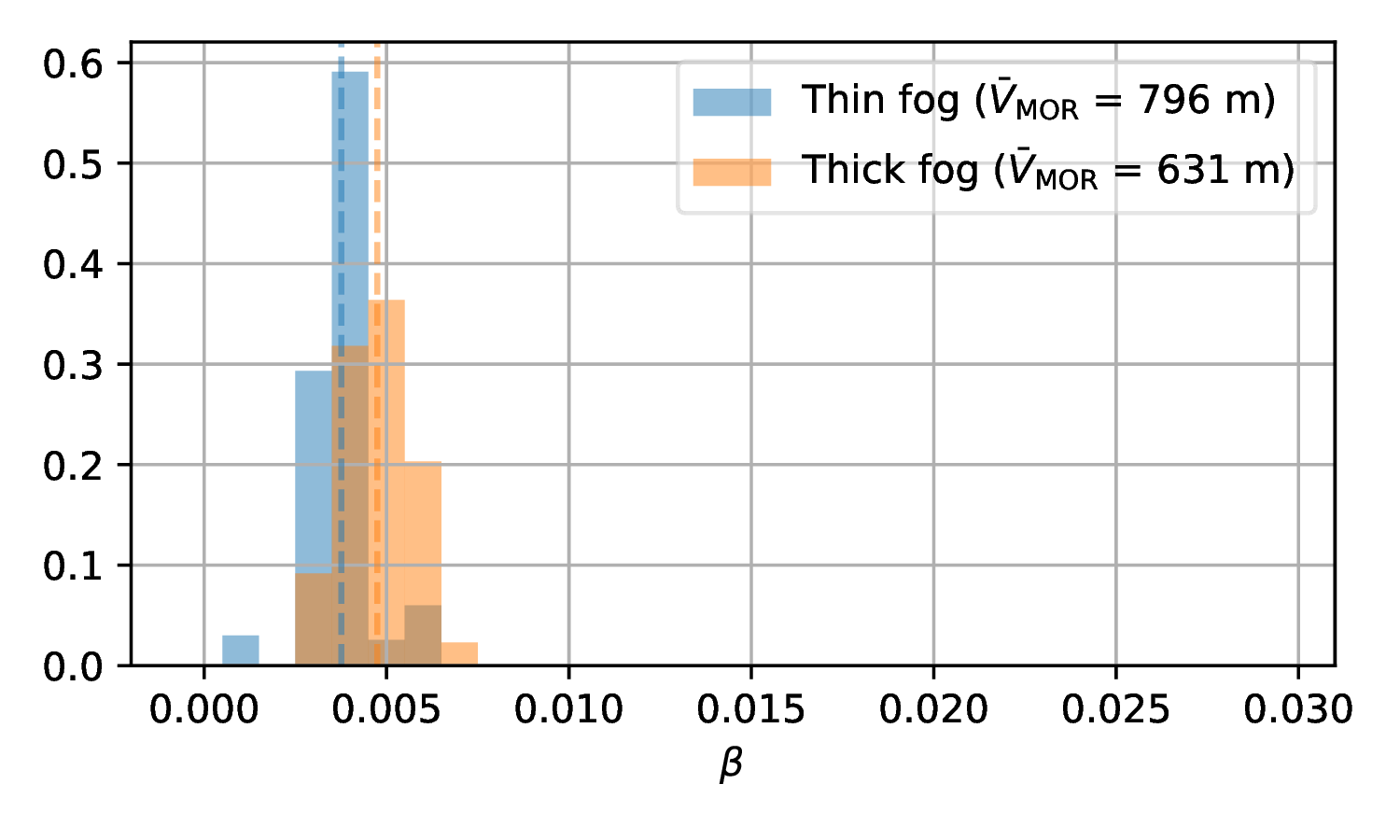} &
        \includegraphics[trim={0.15cm 0.2cm 0.1cm 0.1cm},width=0.244\textwidth]{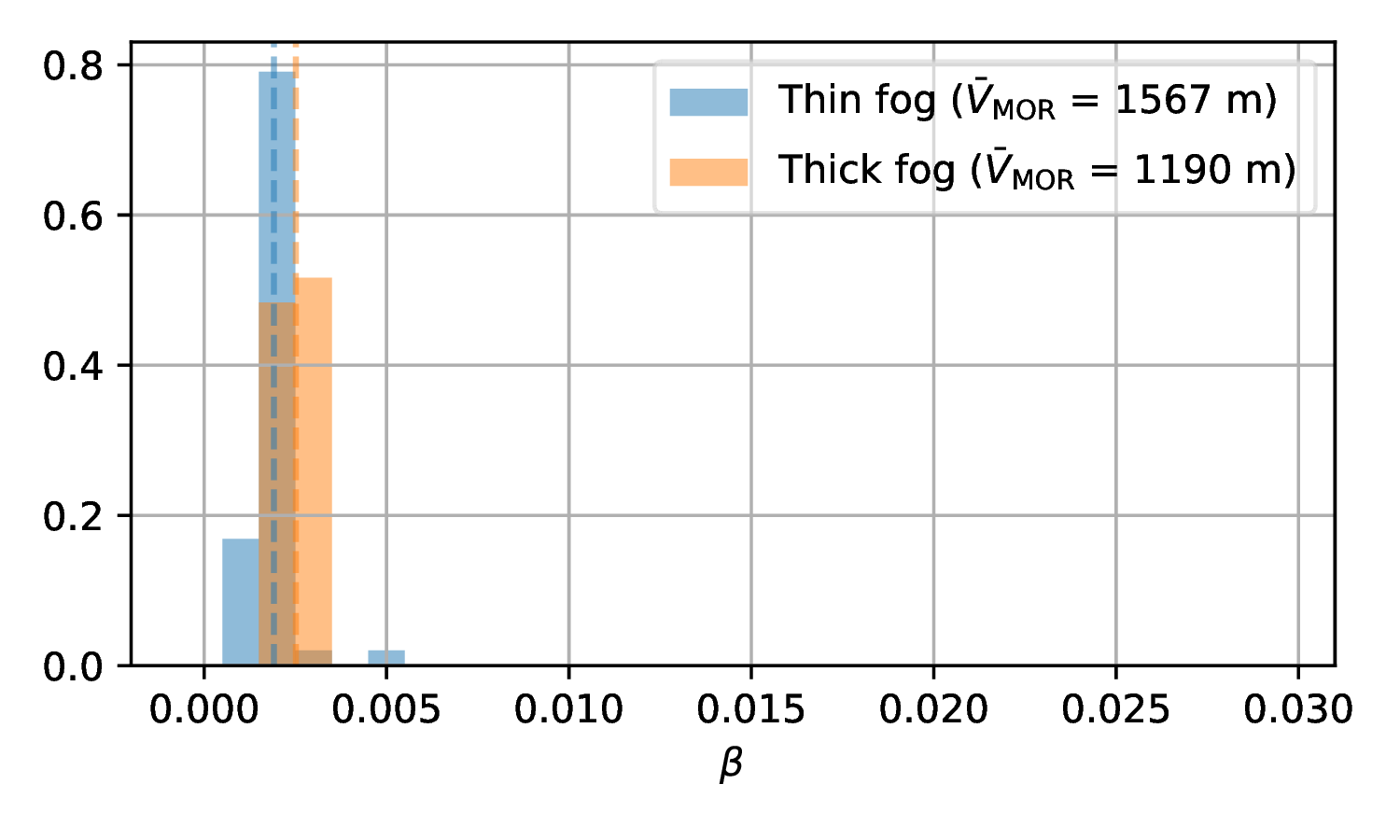} &
        \includegraphics[trim={0.15cm 0.2cm 0.1cm 0.1cm},width=0.244\textwidth]{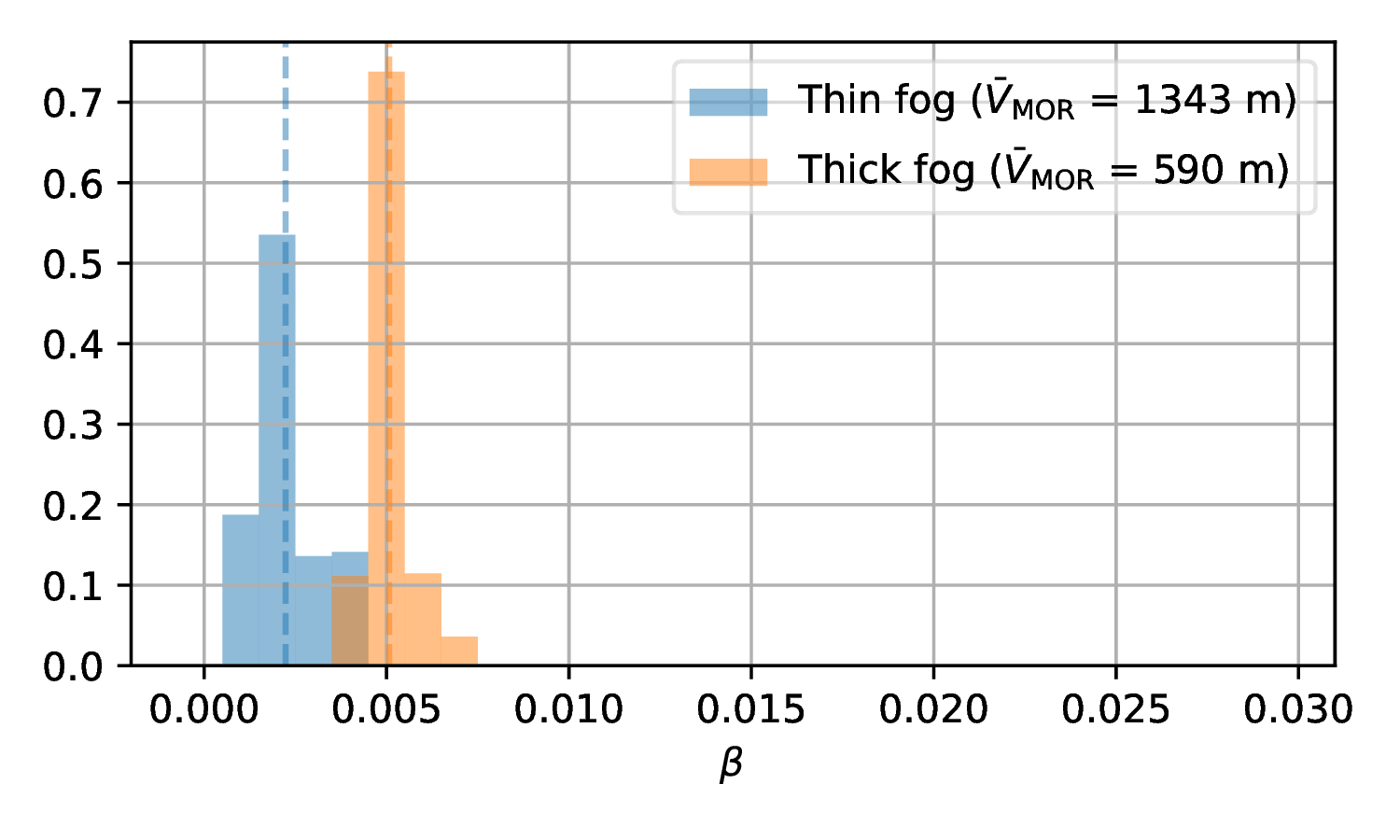} \\

        \rotatebox[origin=lc]{90}{\qquad \; \, \tiny{Ours}} &
        \includegraphics[trim={0.15cm 0.6cm 0.1cm 0.1cm},width=0.244\textwidth]{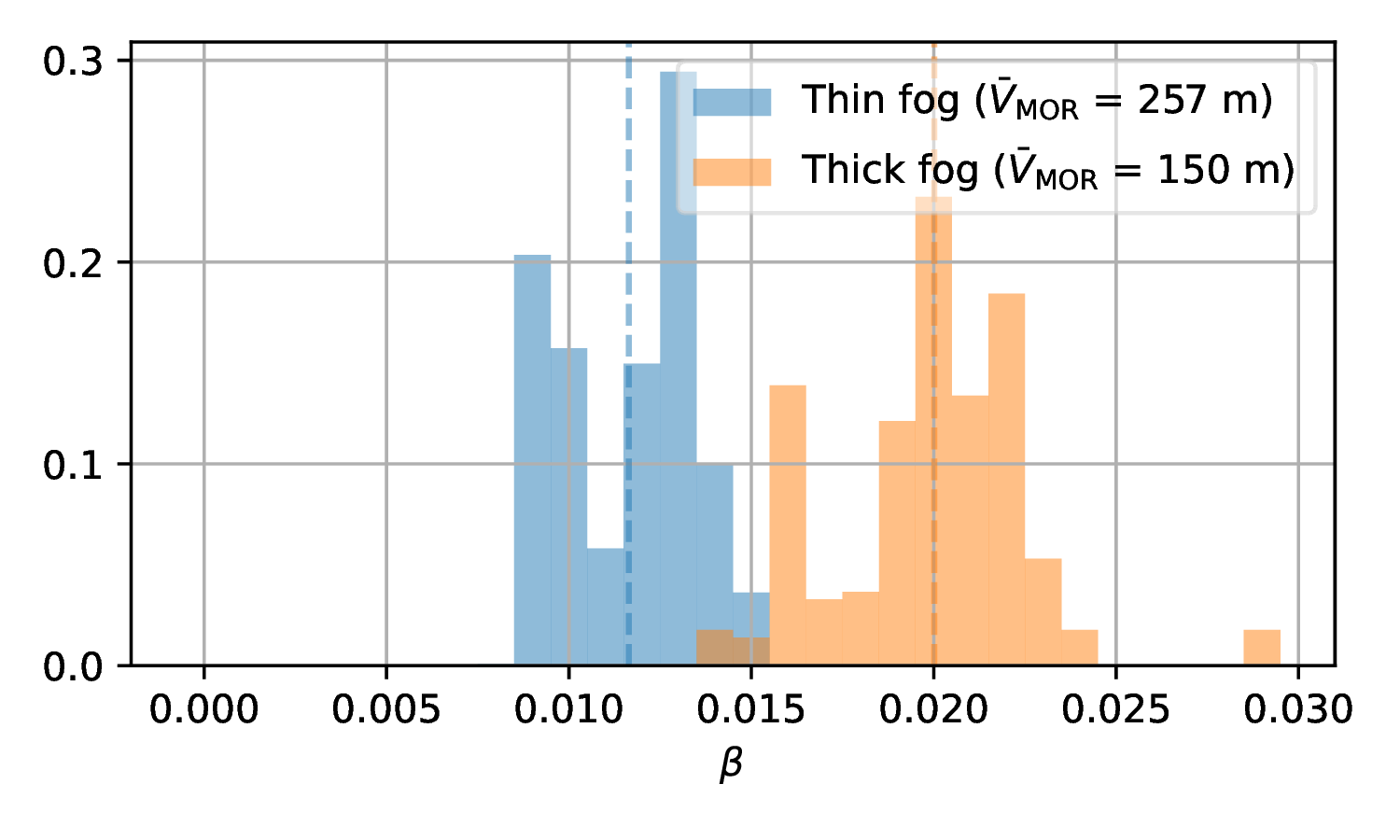} &
        \includegraphics[trim={0.15cm 0.6cm 0.1cm 0.1cm},width=0.244\textwidth]{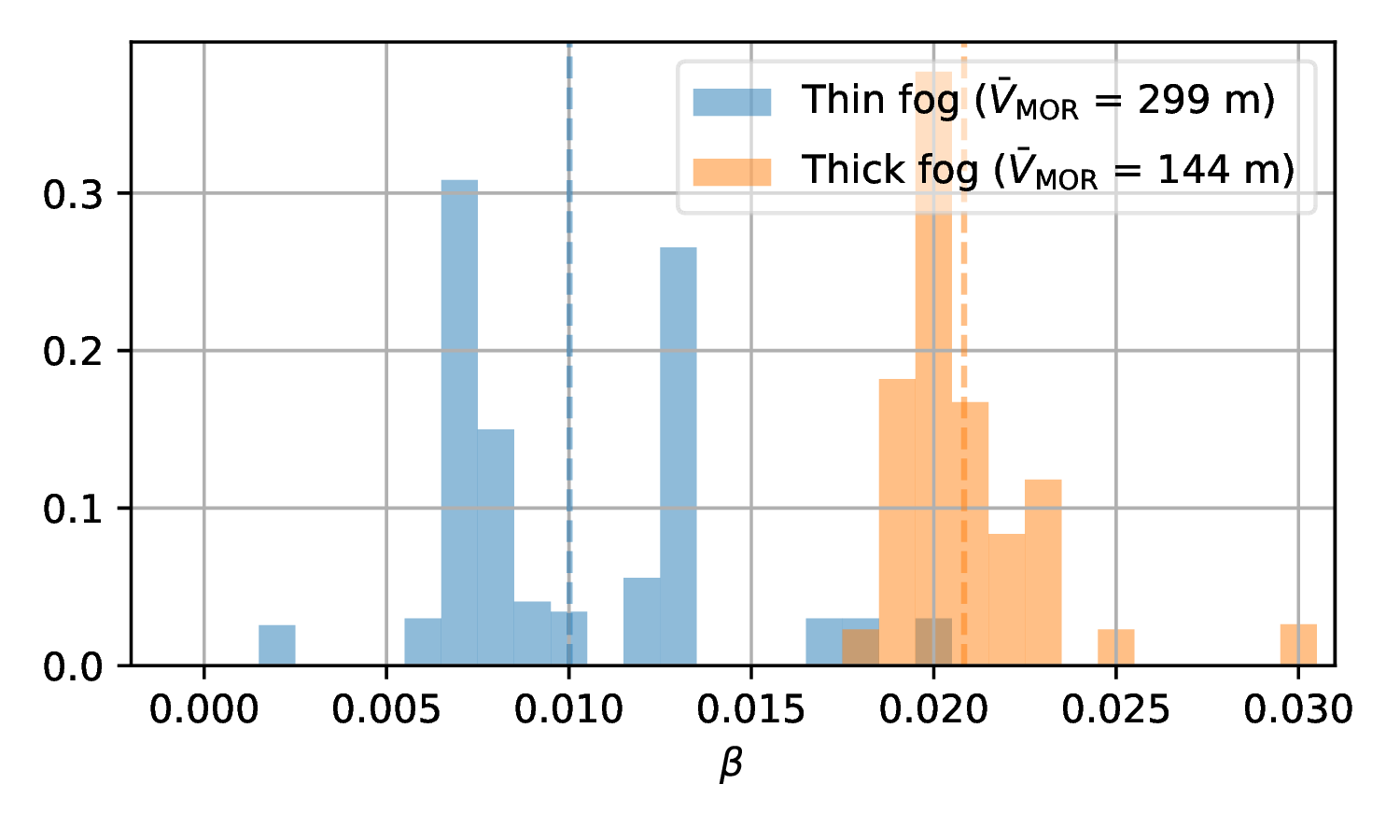} &
        \includegraphics[trim={0.15cm 0.6cm 0.1cm 0.1cm},width=0.244\textwidth]{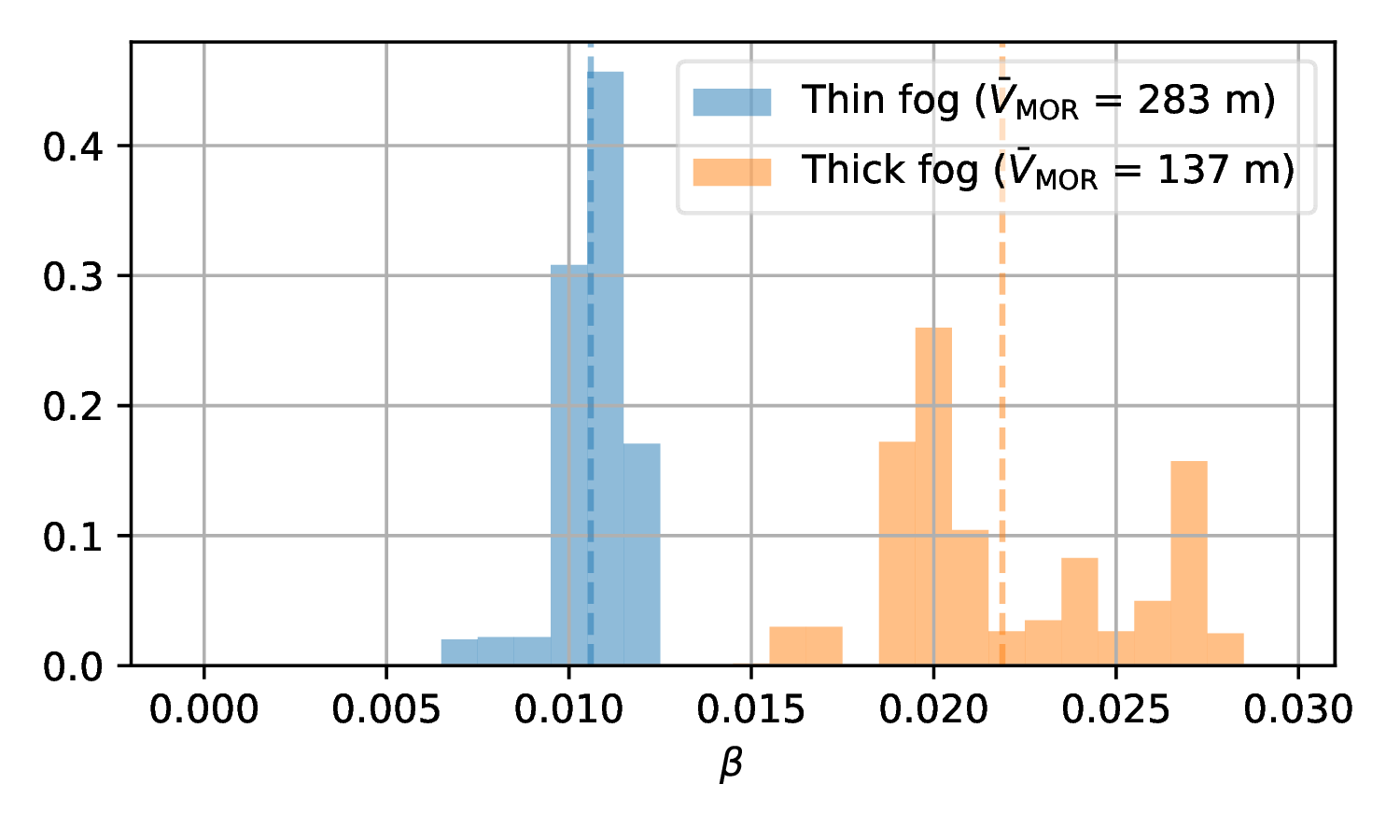} &
        \includegraphics[trim={0.15cm 0.6cm 0.1cm 0.1cm},width=0.244\textwidth]{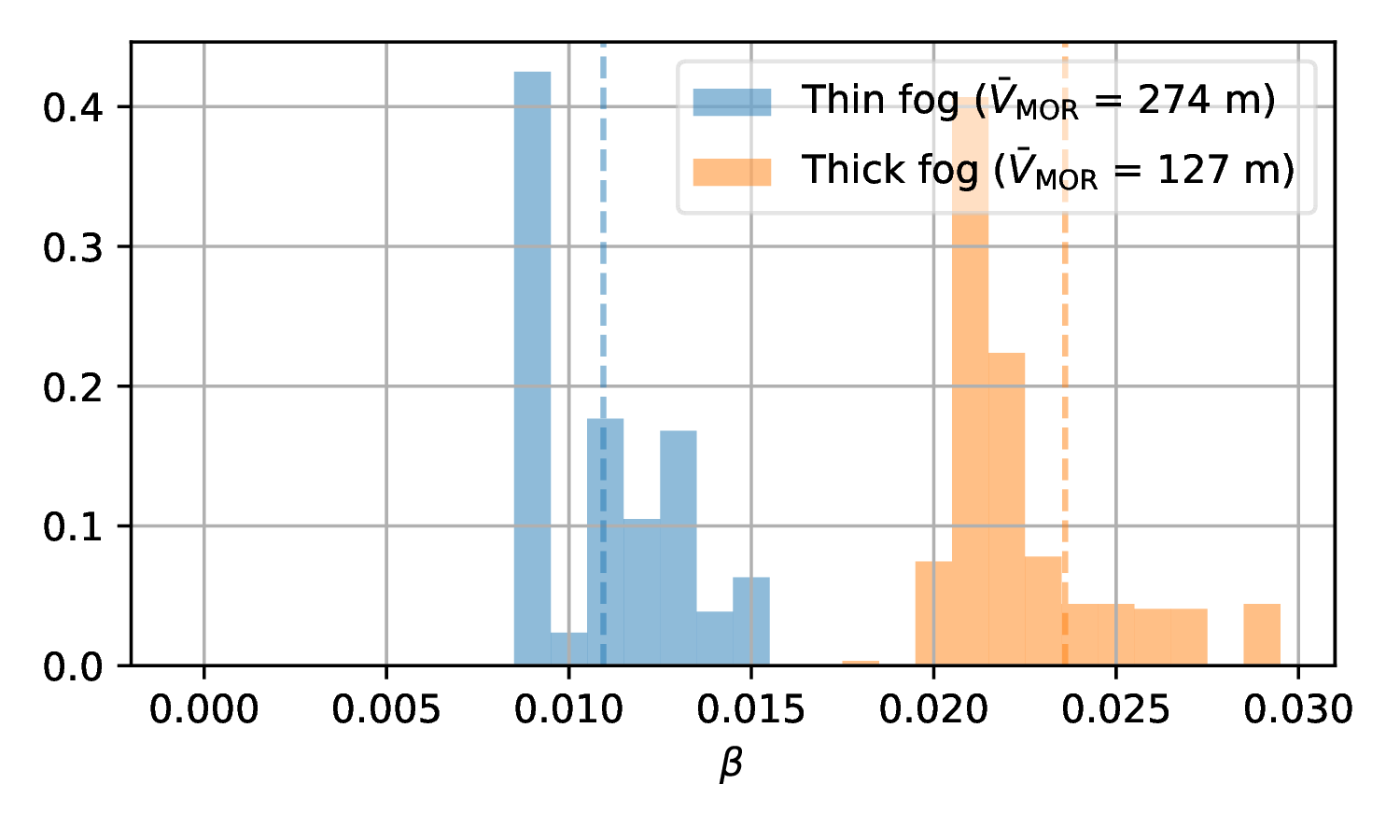} \\
        
    \end{tabular}

    \setlength{\abovecaptionskip}{5pt}
    \caption{Evaluating the inter-sequence consistency of $\beta$ on SDIRF.
    We plot the normalised histograms of the estimated $\beta$ of the whole thin/thick foggy sequences whose first left frames are shown in the corresponding columns of \RefFig \ref{fig:sample_sdirf_images}.
    Each row shows the results of a method.
    Note that all horizontal axes have the same scale.
    Our method is always the best at distinguishing between thin and thick fog by having the least overlap between the two distributions.
    Furthermore, in legend we show the mean visibility, ${\bar{V}}_{\text{MOR}}$, calculated from the mean $\beta$ (indicated by the dashed vertical line) according to \eqref{eq:vis}.
    We observe that the mean visibility values from our method are much more reasonable than the rest when compared with the foggy images in \RefFig \ref{fig:sample_sdirf_images}.
    }
    \setlength{\belowcaptionskip}{-10pt}
    \label{fig:thin_vs_thick_beta_hist}
\end{figure*}

\begin{figure*}[!t]
    \centering
    \setlength\tabcolsep{1pt} 
    \renewcommand{\arraystretch}{0.5}

    \begin{minipage}{0.5\textwidth}
        \centering
        \includegraphics[trim={0.55cm 0.4cm 0.3cm 1.0cm},width=.95\linewidth]{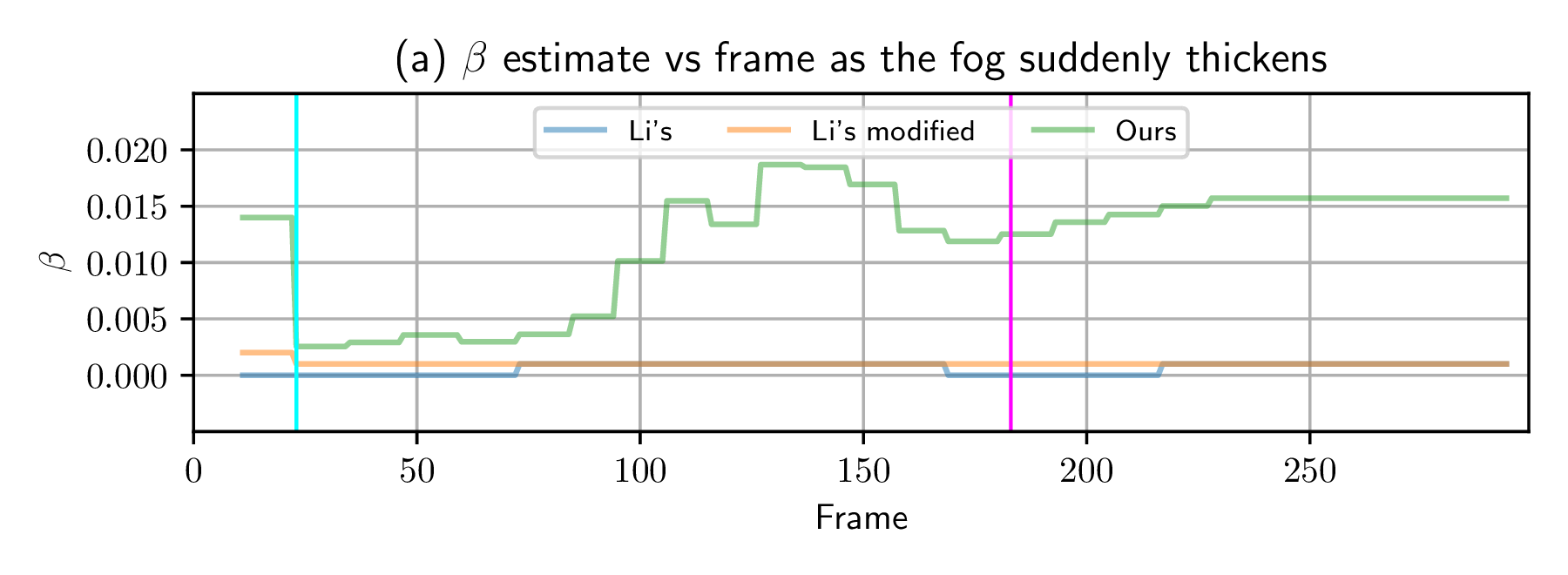}\par
        \vspace*{1mm}
        \begin{tabular}{cc}
            \centering

            \begin{tikzpicture}[node distance = 0cm, anchor=north west, inner sep = 0pt, spy using outlines={rectangle, yellow, magnification=2, every spy on node/.append style={ultra thin}, width=0.05\linewidth, height=0.05\linewidth}]
                \node [anchor=north west,inner sep=-1] (image) at (0,0) {\includegraphics[cfbox=cyan 10pt 0pt, width=.485\linewidth]{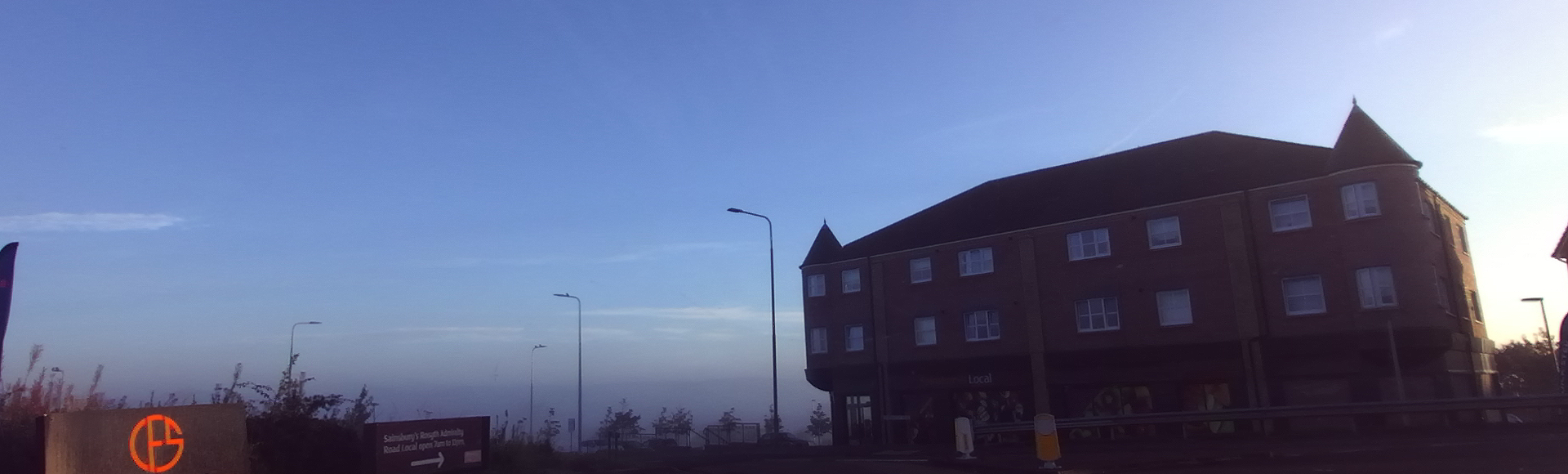}};
                \begin{scope}[x={(image.south east)},y={(image.north west)}]
                    \coordinate (ref) at (.0, .0);
                    \spy on ($(ref) + (1.43, -0.78)$) in node[below left = 0em and 0em of image.north east, line width=0.1mm];
                \end{scope}
            \end{tikzpicture} &
            \begin{tikzpicture}[node distance = 0cm, anchor=north west, inner sep = 0pt, spy using outlines={rectangle, yellow, magnification=2, every spy on node/.append style={ultra thin}, width=0.05\linewidth, height=0.05\linewidth}]
                \node [anchor=north west,inner sep=-1] (image) at (0,0) {\includegraphics[cfbox=magenta 10pt 0pt, width=.485\linewidth]{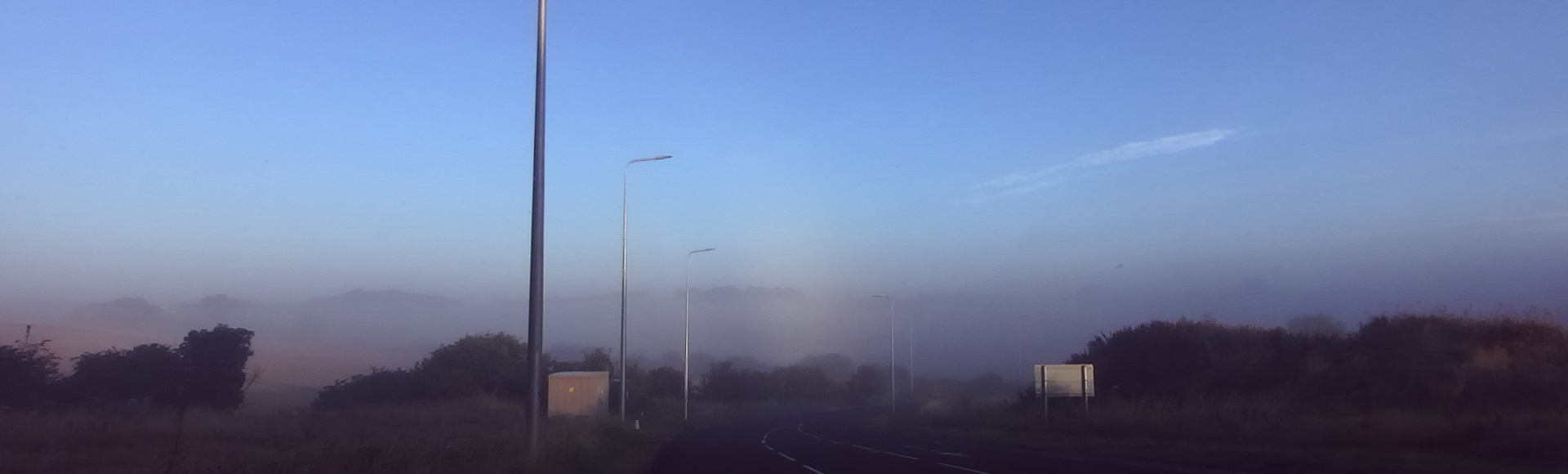}};
                \begin{scope}[x={(image.south east)},y={(image.north west)}]
                    \coordinate (ref) at (.0, .0);
                    \spy on ($(ref) + (2.3, -0.79)$) in node[below left = 0em and 0em of image.north east, line width=0.1mm];
                \end{scope}
            \end{tikzpicture} \\
        \end{tabular}
    \end{minipage}%
    \begin{minipage}{0.5\textwidth}
        \centering
        \includegraphics[trim={0.55cm 0.4cm 0.3cm 1.0cm},width=.95\linewidth]{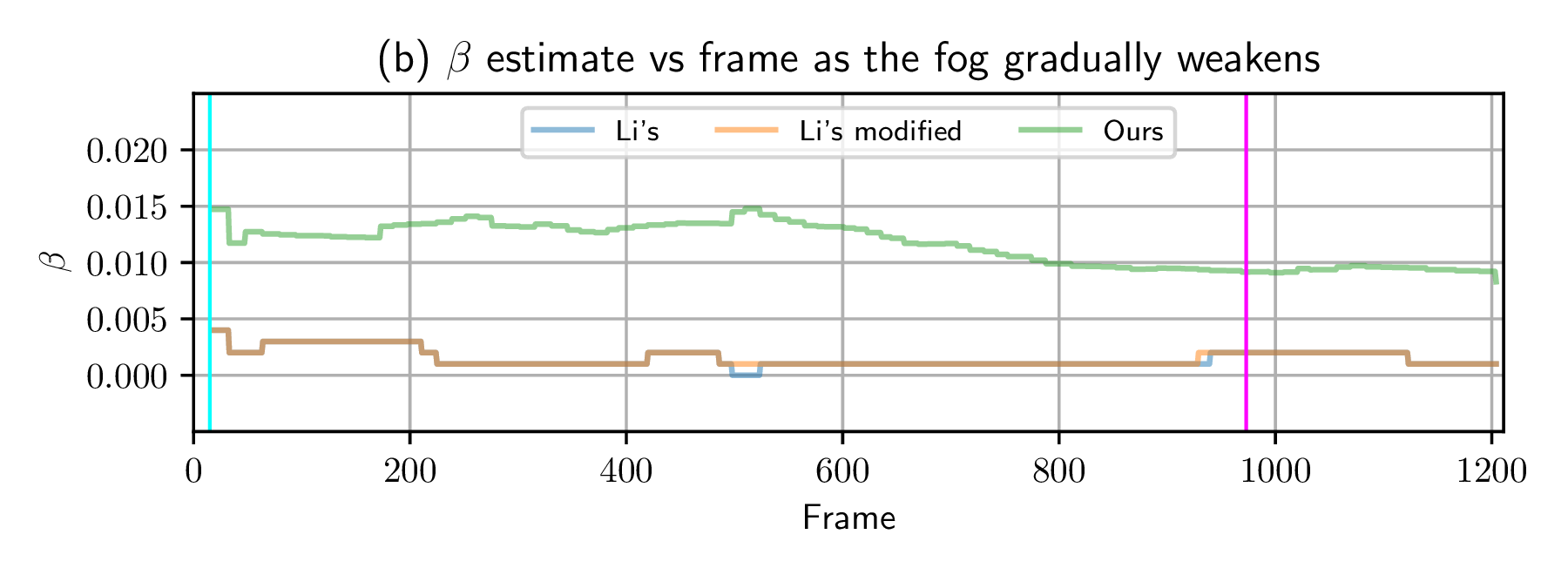}\par
        \vspace*{1mm}
        \begin{tabular}{cc}
            \centering

            \begin{tikzpicture}[node distance = 0cm, anchor=north west, inner sep = 0pt, spy using outlines={rectangle, yellow, magnification=2, every spy on node/.append style={ultra thin}, width=0.05\linewidth, height=0.05\linewidth}]
                \node [anchor=north west,inner sep=-1] (image) at (0,0) {\includegraphics[cfbox=cyan 10pt 0pt, width=.485\linewidth]{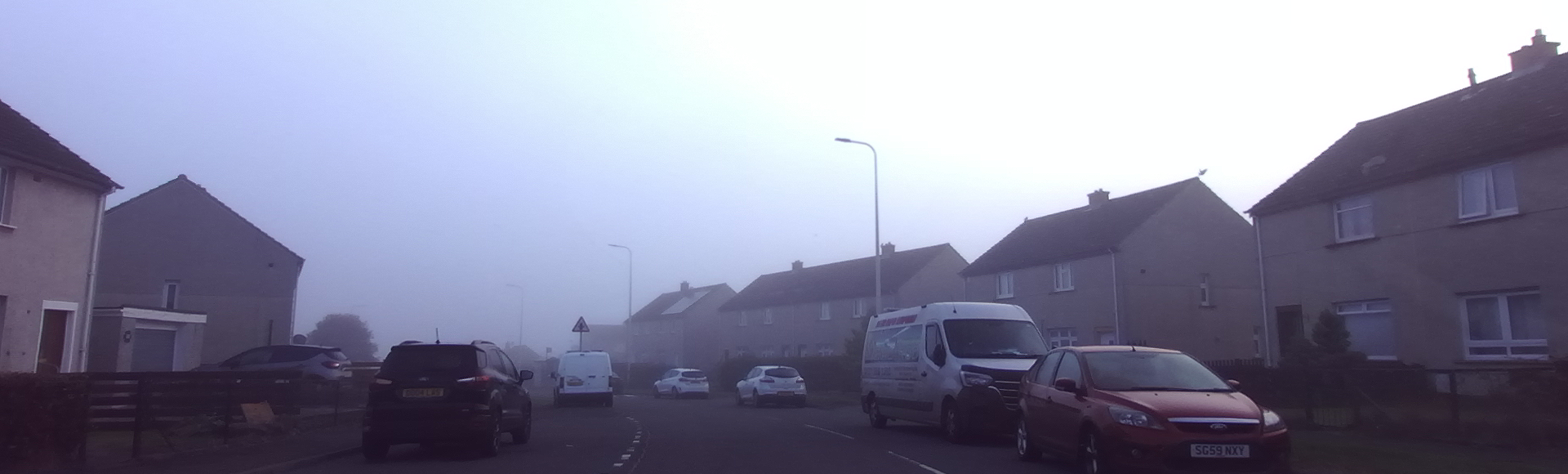}};
                \begin{scope}[x={(image.south east)},y={(image.north west)}]
                    \coordinate (ref) at (.0, .0);
                    \spy on ($(ref) + (1.73, -0.8)$) in node[below left = 0em and 0em of image.north east, line width=0.1mm];
                \end{scope}
            \end{tikzpicture} &
            \begin{tikzpicture}[node distance = 0cm, anchor=north west, inner sep = 0pt, spy using outlines={rectangle, yellow, magnification=2, every spy on node/.append style={ultra thin}, width=0.05\linewidth, height=0.05\linewidth}]
                \node [anchor=north west,inner sep=-1] (image) at (0,0) {\includegraphics[cfbox=magenta 10pt 0pt, width=.485\linewidth]{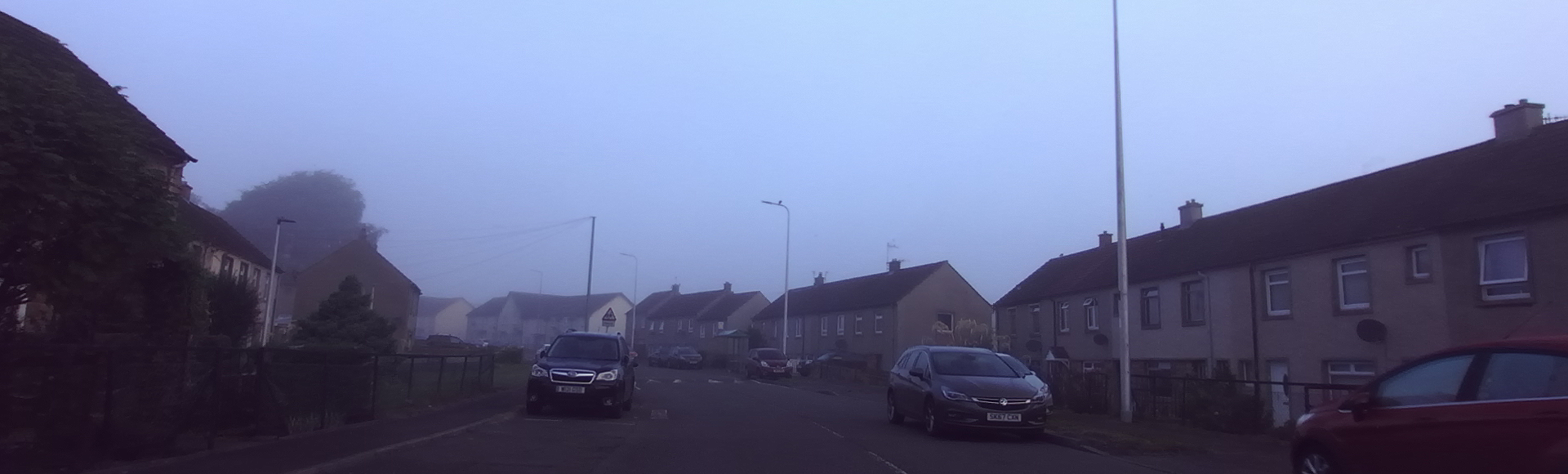}};
                \begin{scope}[x={(image.south east)},y={(image.north west)}]
                    \coordinate (ref) at (.0, .0);
                    \spy on ($(ref) + (1.74, -0.79)$) in node[below left = 0em and 0em of image.north east, line width=0.1mm];
                \end{scope}
            \end{tikzpicture} \\
        \end{tabular}
    \end{minipage}

    \setlength{\abovecaptionskip}{5pt}
    \caption{Evaluating the intra-sequence consistency of $\beta$ on SDIRF.
    We show the estimate of $\beta$ vs frame of two foggy sequences in which the fog demonstrates a noticeable spatial variation in its density.
    (a) The vehicle traverses an area where the fog suddenly thickens.
    (b) The vehicle traverses an area where the fog gradually weakens.
    Shown at the bottom are the foggy images of the frames which are indicated by the vertical cursors in the corresponding plot above.
    Each image is bordered by the colour that matches the corresponding cursor's.
    See the close-up of yellow squares to better visually compare the fog density.
    We observe that our method is the only one that is able to respond adaptively to changes in fog density through updating the estimated values of $\beta$.
    These results also add to the evidence that our assumption of local homogeneity of fog is still valid when the fog is spatially variant.
    }
    \setlength{\belowcaptionskip}{-10pt}
    \label{fig:beta_vs_frame}
\end{figure*}

\subsubsection{Error Metrics vs Visibility}
We investigate how the fog parameter estimation performance varies with visibility.
\RefFig \ref{fig:RMSE_vs_vis} plots the relative RMSE of $\beta$ and $A$ against visibility.

We observe:
a) Our method consistently excels by a large margin in both estimates of $\beta$ and $A$ for all visibility levels tested;
b) Both Li's and Li's modified demonstrate a downward trend in the relative RMSE of $\beta$ as visibility increases.
By comparing the histogram in \RefFig \ref{fig:beta_hist}(c) with that in \RefFig \ref{fig:beta_hist}(b), we infer that as visibility increases, the number of estimates of $\beta$ to build a histogram becomes larger, which in turn improves the performance of the statistics-based estimation method used by the two baseline methods;
c) All methods witness an upward trend in the relative RMSE of $A$ as visibility increases, which is expected because images will appear to be less fog-obscured as visibility increases.

\subsection{Evaluation on Real Data}
For real foggy data from SDIRF, we focus on \emph{qualitative} evaluation because it is not possible to obtain the ground truth values of the fog parameters for real foggy images taken in an open, uncontrolled environment.

\subsubsection{Scattering Coefficient Estimation}
We examine both \emph{inter-sequence} consistency and \emph{intra-sequence} consistency between the perceptual density of the fog and the estimated $\beta$.

Firstly, in \RefFig \ref{fig:thin_vs_thick_beta_hist} we show the normalised histograms of $\beta$ estimated by various methods of the eight foggy sequences whose first left frames are shown in \RefFig \ref{fig:sample_sdirf_images}.
The key observation is that our method demonstrates the best inter-sequence consistency between the visual appearance of the foggy images and the estimated $\beta$.
In addition, the mean visibility value computed according to \eqref{eq:vis} using the mean $\beta$ value estimated by our method is perceptually more sensible than the rest.
See our supplementary material for more results.

Secondly, in \RefFig \ref{fig:beta_vs_frame} we plot the estimated $\beta$ vs frame of two foggy sequences in which the fog demonstrates a noticeable spatial variation in its density.
The key observation is that our method, being the only one that is capable of responding adaptively to spatial variation in fog density, demonstrates the best intra-sequence consistency between the visual appearance of the foggy images and the estimated $\beta$.

\subsubsection{Atmospheric Light Estimation}
\begin{figure*}
    \centering
    \setlength\tabcolsep{1pt} 
    \renewcommand{\arraystretch}{0.5}

    \begin{minipage}{0.25\textwidth}
        \centering
        \setlength\tabcolsep{1pt} 
        \renewcommand{\arraystretch}{0.5}
        \begin{tabular}{ccccc}
            \multicolumn{5}{c}{\tiny{(a)}} \\

            \multicolumn{5}{c}{\includegraphics[width=.98\linewidth]{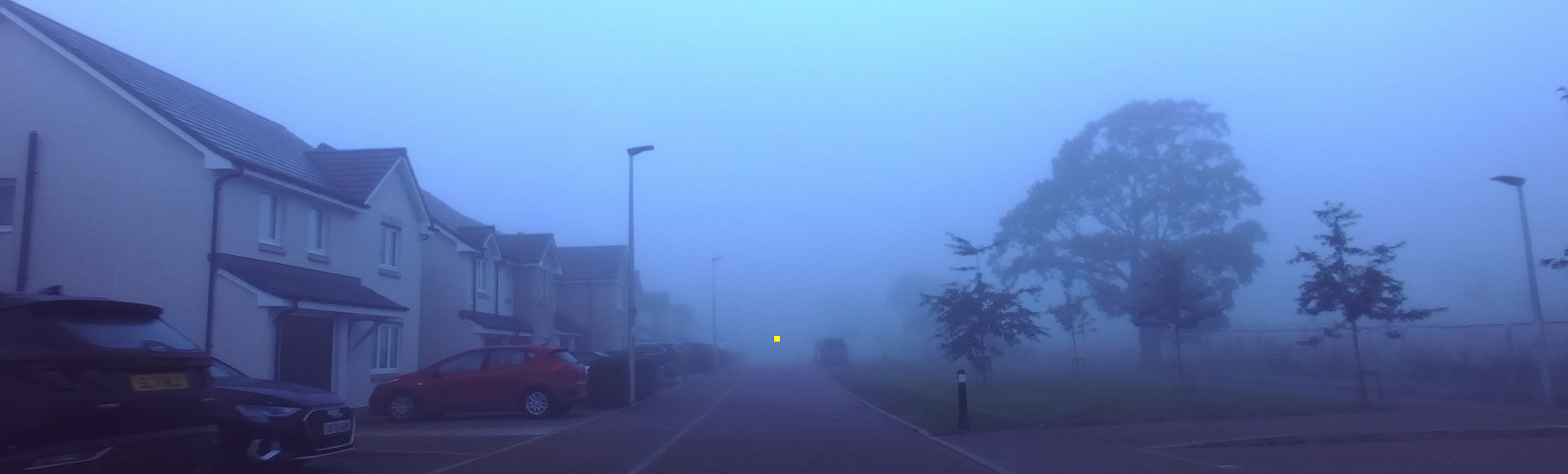}} \\

            \begin{overpic}[width=0.208\linewidth]{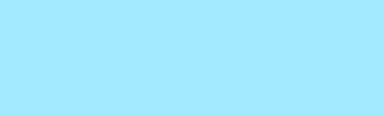}
            \put (23,8.5) {\tiny 146.36}
            \end{overpic} &
            \begin{overpic}[width=0.208\linewidth]{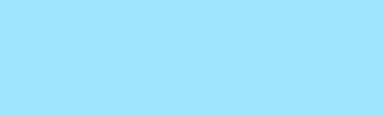}
            \put (23,8.5) {\tiny 139.03}
            \end{overpic} &
            \begin{overpic}[width=0.208\linewidth]{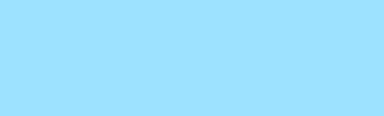}
            \put (23,8.5) {\tiny 138.17}
            \end{overpic} &
            \begin{overpic}[width=0.208\linewidth]{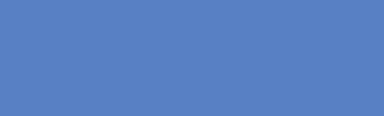}
            \put (34.5,8.5) {\tiny 8.11}
            \end{overpic} &
            \includegraphics[trim={0cm 0cm 8.14cm 0cm}, clip, width=0.083\linewidth]{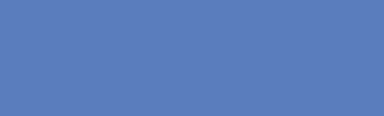} \\

            \tiny{Li's \cite{li2015simultaneous}} &
            \tiny{Li's mod} &
            \tiny{Berman's \cite{berman2016non}} &
            \tiny{Ours} &
            \tiny{GT}
            
        \end{tabular}
    \end{minipage}%
    \begin{minipage}{0.25\textwidth}
        \centering
        \setlength\tabcolsep{1pt} 
        \renewcommand{\arraystretch}{0.5}
        \begin{tabular}{ccccc}
            \multicolumn{5}{c}{\tiny{(b)}} \\
 
            \multicolumn{5}{c}{\includegraphics[width=.98\linewidth]{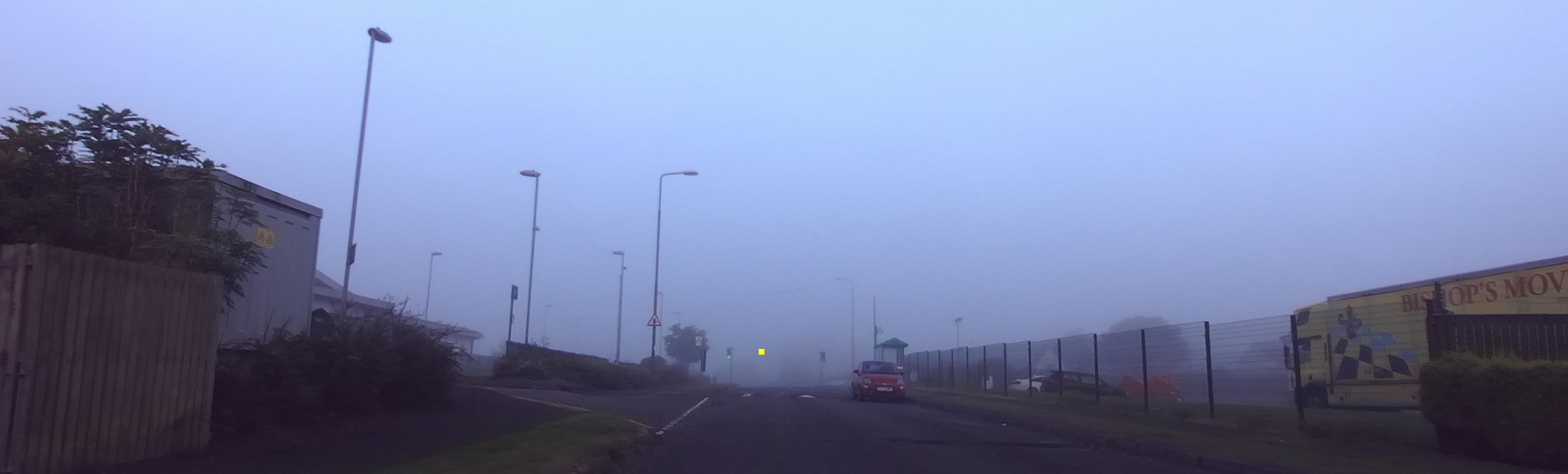}} \\

            \begin{overpic}[width=0.208\linewidth]{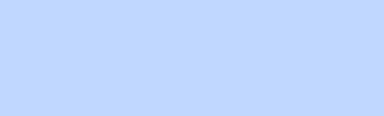}
            \put (23,8.5) {\tiny 144.10}
            \end{overpic} &
            \begin{overpic}[width=0.208\linewidth]{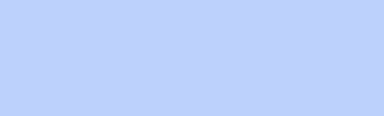}
            \put (23,8.5) {\tiny 136.04}
            \end{overpic} &
            \begin{overpic}[width=0.208\linewidth]{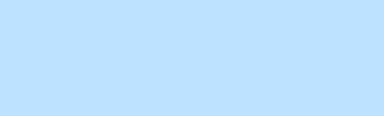}
            \put (23,8.5) {\tiny 148.83}
            \end{overpic} &
            \begin{overpic}[width=0.208\linewidth]{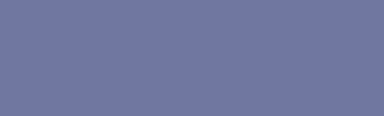}
            \put (29,8.5) {\tiny 14.01}
            \end{overpic} &
            \includegraphics[trim={0cm 0cm 8.14cm 0cm}, clip, width=0.083\linewidth]{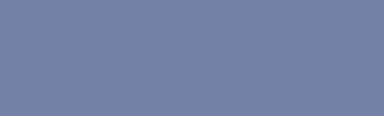} \\

            \tiny{Li's \cite{li2015simultaneous}} &
            \tiny{Li's mod} &
            \tiny{Berman's \cite{berman2016non}} &
            \tiny{Ours} &
            \tiny{GT}
            
        \end{tabular}
    \end{minipage}%
    \begin{minipage}{0.25\textwidth}
        \centering
        \setlength\tabcolsep{1pt} 
        \renewcommand{\arraystretch}{0.5}
        \begin{tabular}{ccccc}
            \multicolumn{5}{c}{\tiny{(c)}} \\

            \multicolumn{5}{c}{\includegraphics[width=.98\linewidth]{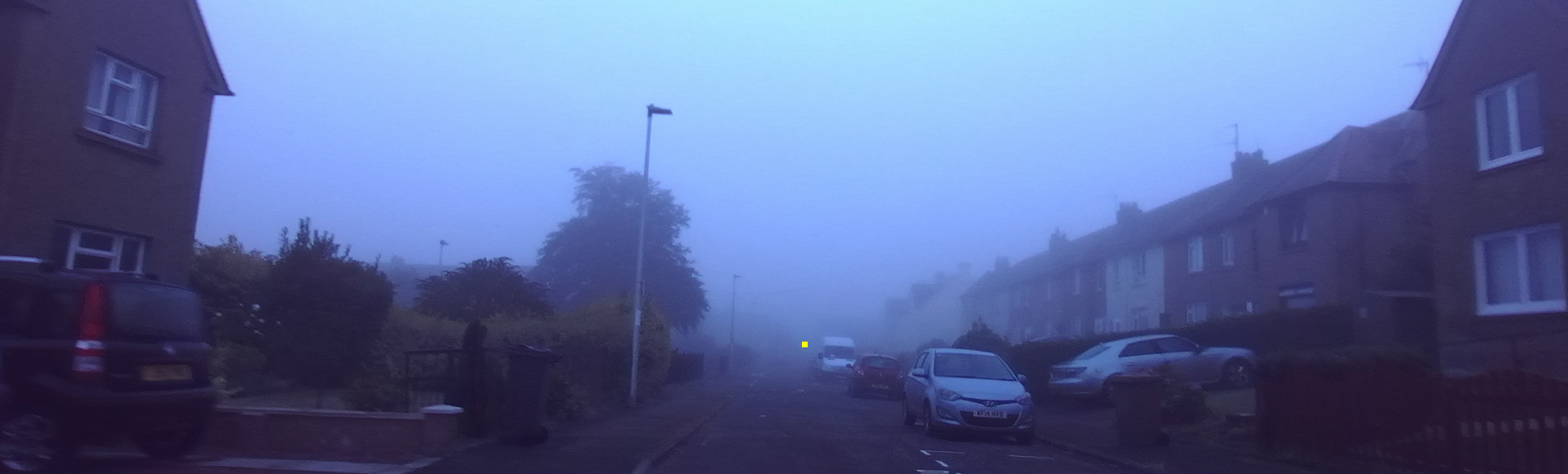}} \\

            \begin{overpic}[width=0.208\linewidth]{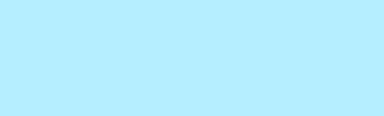}
            \put (23,8.5) {\tiny 171.93}
            \end{overpic} &
            \begin{overpic}[width=0.208\linewidth]{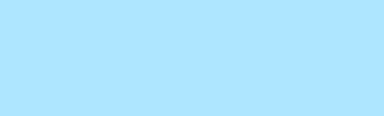}
            \put (23,8.5) {\tiny 164.31}
            \end{overpic} &
            \begin{overpic}[width=0.208\linewidth]{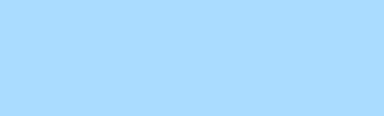}
            \put (23,8.5) {\tiny 155.19}
            \end{overpic} &
            \begin{overpic}[width=0.208\linewidth]{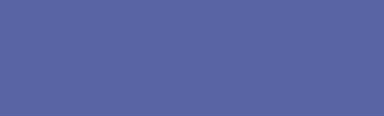}
            \put (29,8.5) {\tiny 20.55}
            \end{overpic} &
            \includegraphics[trim={0cm 0cm 8.14cm 0cm}, clip, width=0.083\linewidth]{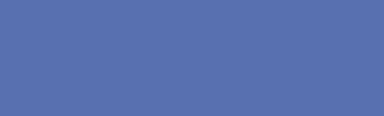} \\

            \tiny{Li's \cite{li2015simultaneous}} &
            \tiny{Li's mod} &
            \tiny{Berman's \cite{berman2016non}} &
            \tiny{Ours} &
            \tiny{GT}
            
        \end{tabular}
    \end{minipage}%
    \begin{minipage}{0.25\textwidth}
        \centering
        \setlength\tabcolsep{1pt} 
        \renewcommand{\arraystretch}{0.5}
        \begin{tabular}{ccccc}
            \multicolumn{5}{c}{\tiny{(d)}} \\

            \multicolumn{5}{c}{\includegraphics[width=.98\linewidth]{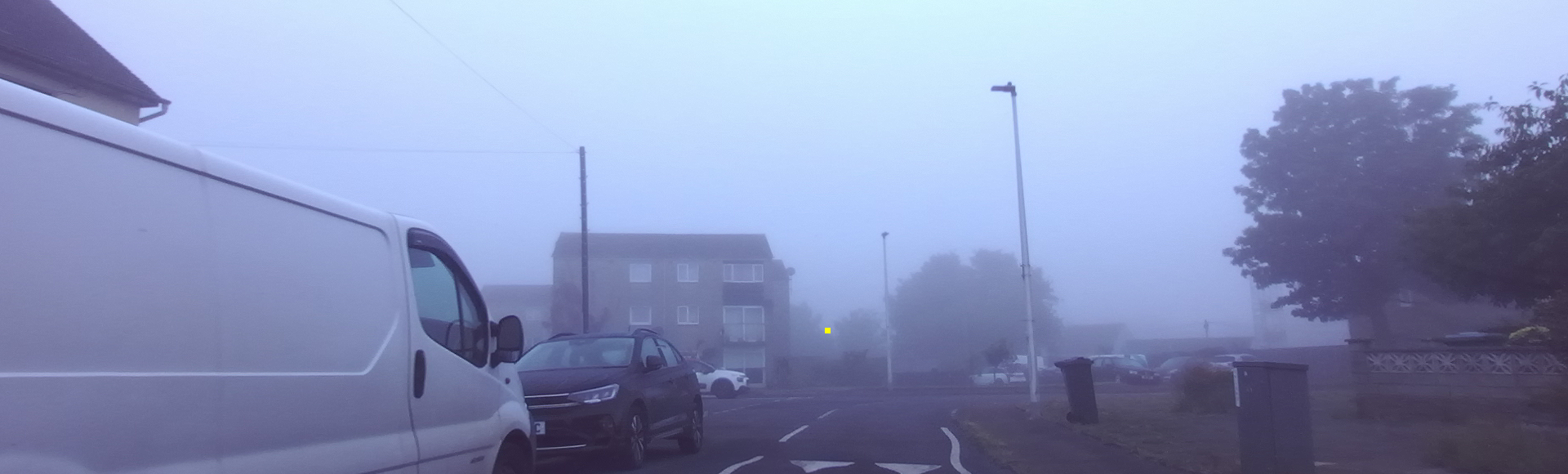}} \\

            \begin{overpic}[width=0.208\linewidth]{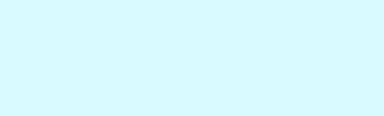}
            \put (23,8.5) {\tiny 147.22}
            \end{overpic} &
            \begin{overpic}[width=0.208\linewidth]{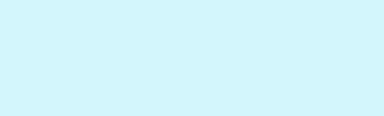}
            \put (23,8.5) {\tiny 139.20}
            \end{overpic} &
            \begin{overpic}[width=0.208\linewidth]{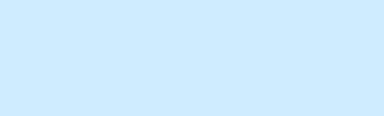}
            \put (23,8.5) {\tiny 131.04}
            \end{overpic} &
            \begin{overpic}[width=0.208\linewidth]{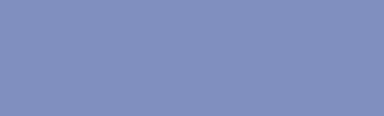}
            \put (34.5,8.5) {\tiny 7.84}
            \end{overpic} &
            \includegraphics[trim={0cm 0cm 8.14cm 0cm}, clip, width=0.083\linewidth]{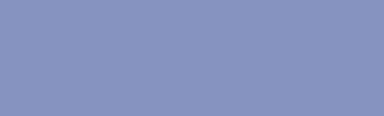} \\

            \tiny{Li's \cite{li2015simultaneous}} &
            \tiny{Li's mod} &
            \tiny{Berman's \cite{berman2016non}} &
            \tiny{Ours} &
            \tiny{GT}
            
        \end{tabular}
    \end{minipage}

    \setlength{\abovecaptionskip}{2.5pt}
    \caption{Evaluating the accuracy of the estimated $A$ on SDIRF.
    The small rectangles at the bottom are painted the colours of $A$ estimated by various methods.
    We also show $A$'s pseudo-ground truth colour, which is extracted by visually examining each foggy image then manually selecting a pixel just above the horizon in the central area (see the little yellow square in each foggy image).
    Each rectangle is overlaid with the Euclidean distance from the corresponding estimate to the pseudo-ground truth.
    Note that the real foggy images are typically very different from the simulated ones shown in \RefFig \ref{fig:synthetic_foggy_images} in a way that the sky region is not of a uniform colour, which is particularly the case of a foggy image taken at dawn.
    We infer that this phenomenon causes competitive methods to fail as they all estimate the atmospheric light from a single image.
    The results demonstrate that only our method is able to accurately unveil the atmospheric light.
    In addition, our method is more robust to changes in the atmospheric light.
    }
    \setlength{\belowcaptionskip}{-20pt}
    \label{fig:compare_atmos}
\end{figure*}

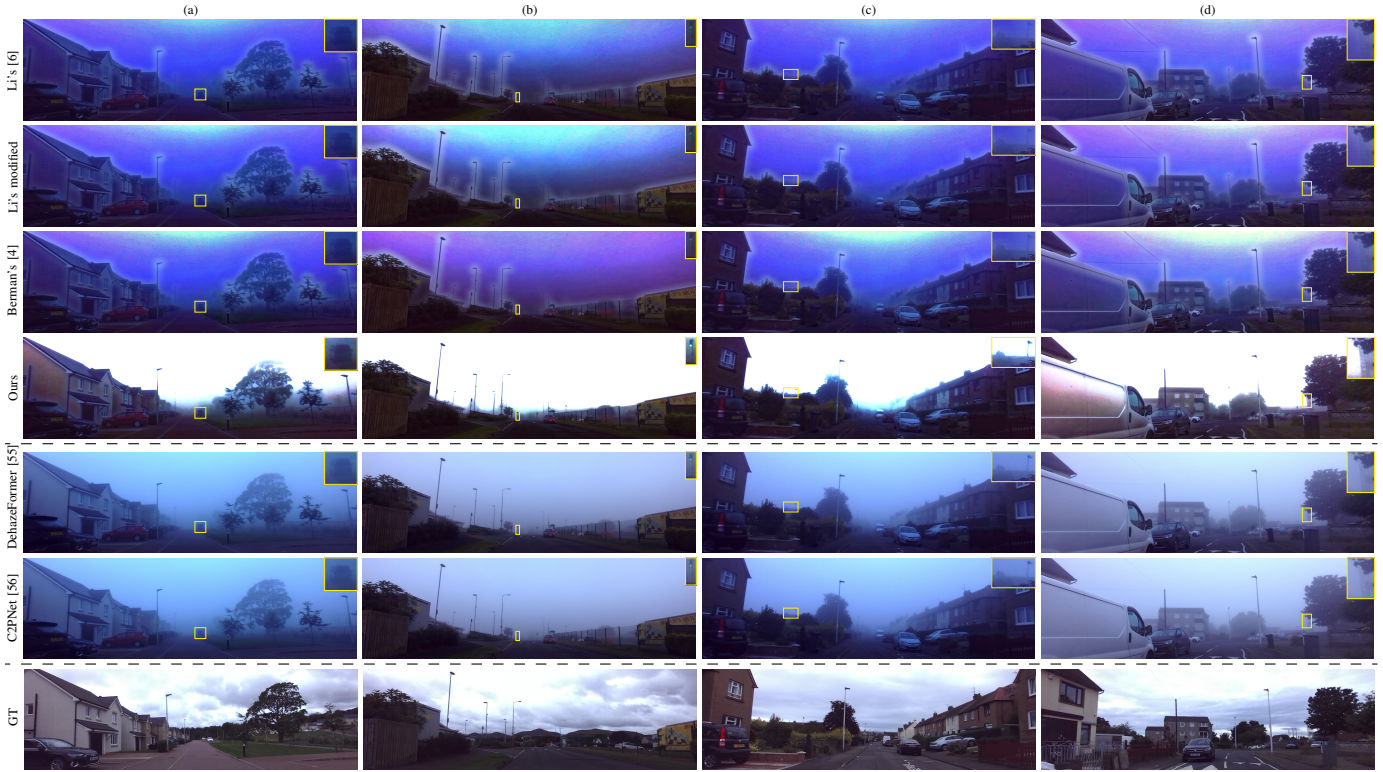
\begin{figure*}
    \centering
    \setlength\tabcolsep{1pt} 
    \renewcommand{\arraystretch}{0.5}
    \begin{tabular}{ccccc}
        \centering
        
        &
        \tiny{(a)} &
        \tiny{(b)} &
        \tiny{(c)} &
        \tiny{(d)} \\

        \rotatebox[origin=lc]{90}{\; \ \tiny{Li's \cite{li2015simultaneous}}} &
        \begin{tikzpicture}[node distance = 0cm, anchor=north west, inner sep = 0pt, spy using outlines={rectangle, yellow, magnification=3, every spy on node/.append style={ultra thin}, width=0.024\textwidth, height=0.024\textwidth}]
            \node [anchor=north west,inner sep=0] (image) at (0,0) {\includegraphics[width=0.244\textwidth]{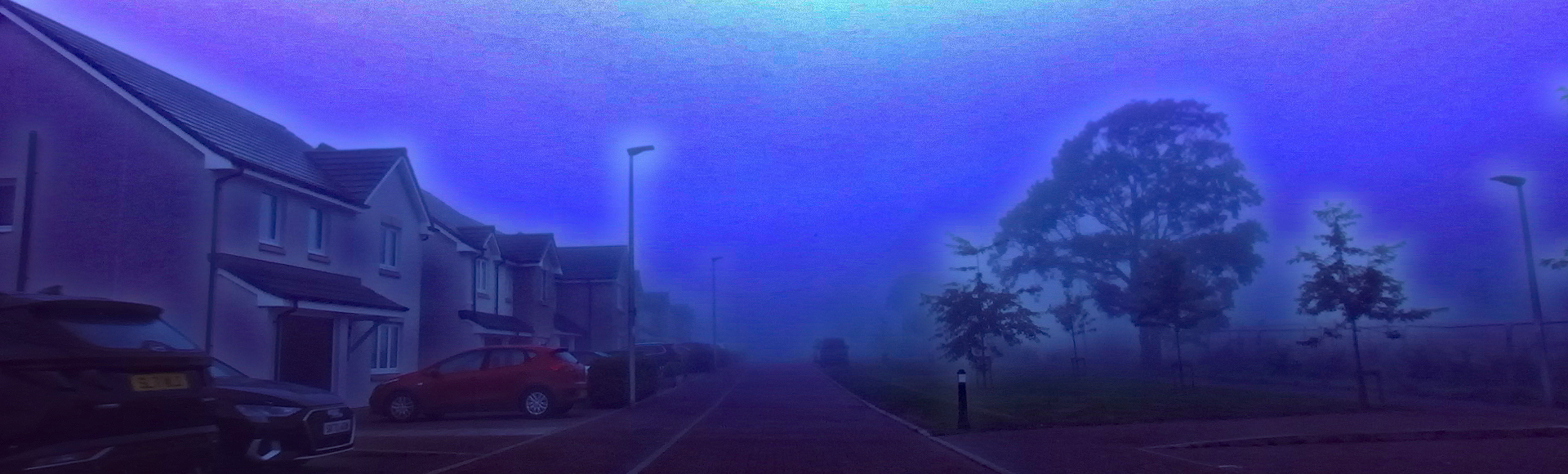}};
            \begin{scope}[x={(image.south east)},y={(image.north west)}]
                \coordinate (ref) at (.0, .0);
                \spy on ($(ref) + (2.28, -0.93)$) in node[below left = 0em and 0em of image.north east, line width=0.1mm];
            \end{scope}
        \end{tikzpicture} &
        \begin{tikzpicture}[node distance = 0cm, anchor=north west, inner sep = 0pt, spy using outlines={rectangle, yellow, magnification=3, every spy on node/.append style={ultra thin}, width=0.008\textwidth, height=0.020\textwidth}]
            \node [anchor=north west,inner sep=0] (image) at (0,0) {\includegraphics[width=0.244\textwidth]{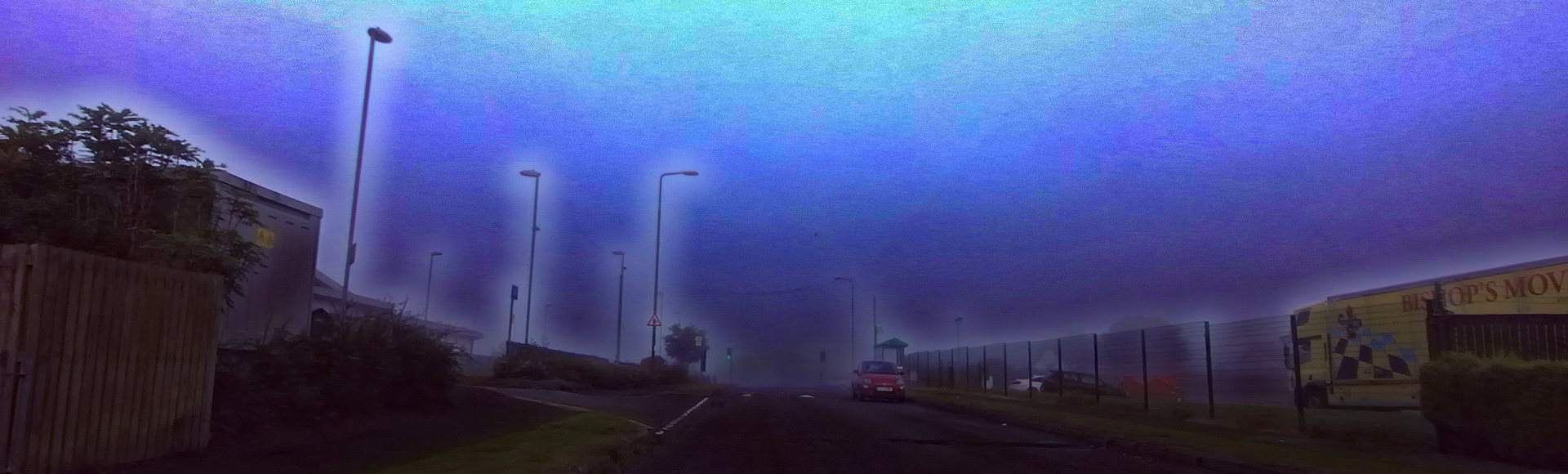}};
            \begin{scope}[x={(image.south east)},y={(image.north west)}]
                \coordinate (ref) at (.0, .0);
                \spy on ($(ref) + (2.04, -0.98)$) in node[below left = 0em and 0em of image.north east, line width=0.1mm];
            \end{scope}
        \end{tikzpicture} &
        \begin{tikzpicture}[node distance = 0cm, anchor=north west, inner sep = 0pt, spy using outlines={rectangle, yellow, magnification=3, every spy on node/.append style={ultra thin}, width=0.032\textwidth, height=0.022\textwidth}]
            \node [anchor=north west,inner sep=0] (image) at (0,0) {\includegraphics[width=0.244\textwidth]{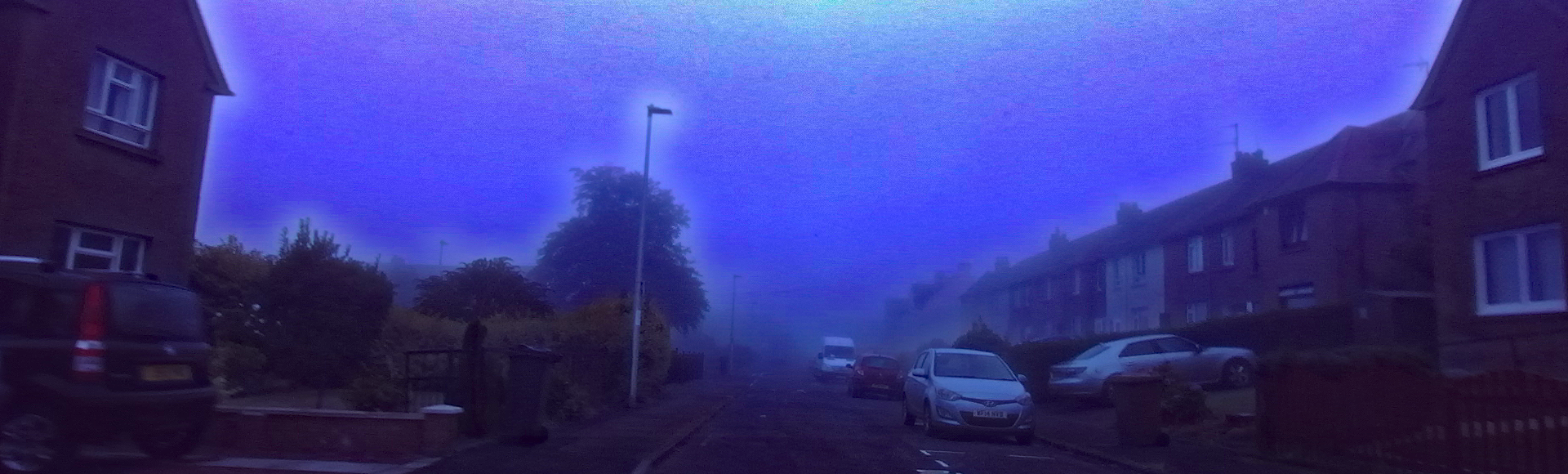}};
            \begin{scope}[x={(image.south east)},y={(image.north west)}]
                \coordinate (ref) at (.0, .0);
                \spy on ($(ref) + (1.09, -0.67)$) in node[below left = 0em and 0em of image.north east, line width=0.1mm];
            \end{scope}
        \end{tikzpicture} &
        \begin{tikzpicture}[node distance = 0cm, anchor=north west, inner sep = 0pt, spy using outlines={rectangle, yellow, magnification=3, every spy on node/.append style={ultra thin}, width=0.020\textwidth, height=0.030\textwidth}]
            \node [anchor=north west,inner sep=0] (image) at (0,0) {\includegraphics[width=0.244\textwidth]{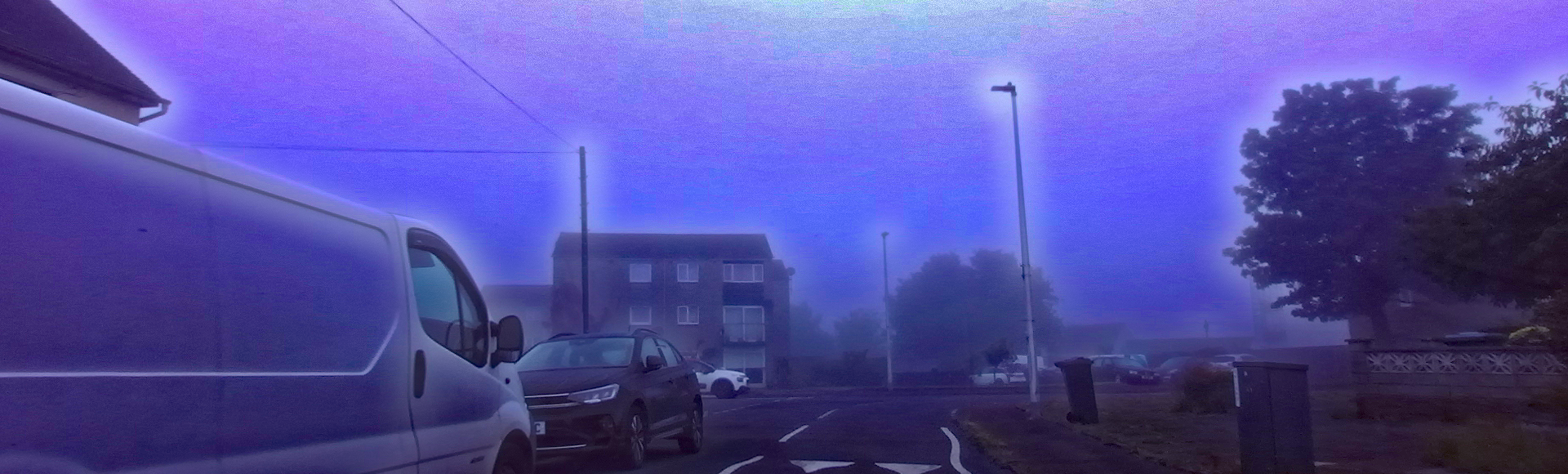}};
            \begin{scope}[x={(image.south east)},y={(image.north west)}]
                \coordinate (ref) at (.0, .0);
                \spy on ($(ref) + (3.48, -0.75)$) in node[below left = 0em and 0em of image.north east, line width=0.1mm];
            \end{scope}
        \end{tikzpicture} \\

        \rotatebox[origin=lc]{90}{\tiny{\; Li's modified}} &
        \begin{tikzpicture}[node distance = 0cm, anchor=north west, inner sep = 0pt, spy using outlines={rectangle, yellow, magnification=3, every spy on node/.append style={ultra thin}, width=0.024\textwidth, height=0.024\textwidth}]
            \node [anchor=north west,inner sep=0] (image) at (0,0) {\includegraphics[width=0.244\textwidth]{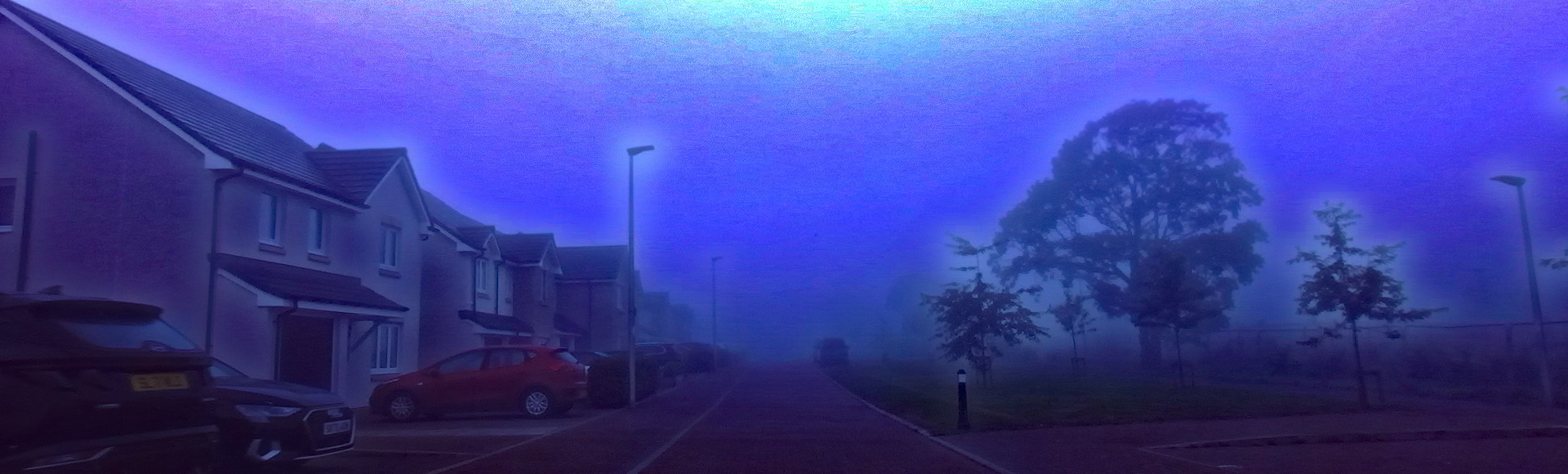}};
            \begin{scope}[x={(image.south east)},y={(image.north west)}]
                \coordinate (ref) at (.0, .0);
                \spy on ($(ref) + (2.28, -0.93)$) in node[below left = 0em and 0em of image.north east, line width=0.1mm];
            \end{scope}
        \end{tikzpicture} &
        \begin{tikzpicture}[node distance = 0cm, anchor=north west, inner sep = 0pt, spy using outlines={rectangle, yellow, magnification=3, every spy on node/.append style={ultra thin}, width=0.008\textwidth, height=0.020\textwidth}]
            \node [anchor=north west,inner sep=0] (image) at (0,0) {\includegraphics[width=0.244\textwidth]{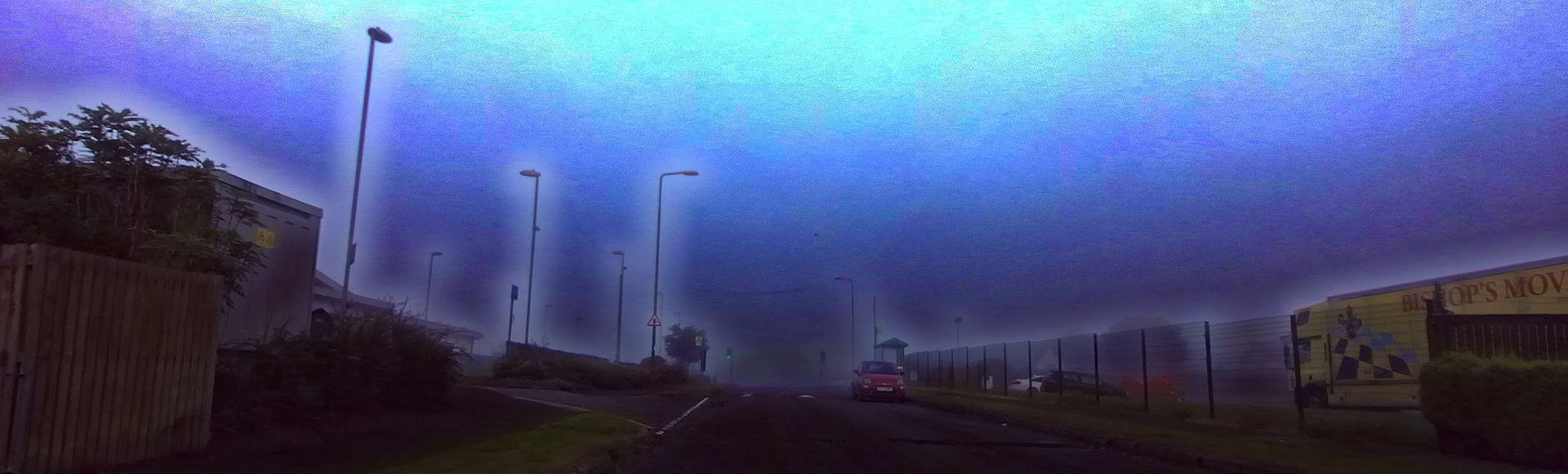}};
            \begin{scope}[x={(image.south east)},y={(image.north west)}]
                \coordinate (ref) at (.0, .0);
                \spy on ($(ref) + (2.04, -0.98)$) in node[below left = 0em and 0em of image.north east, line width=0.1mm];
            \end{scope}
        \end{tikzpicture} &
        \begin{tikzpicture}[node distance = 0cm, anchor=north west, inner sep = 0pt, spy using outlines={rectangle, yellow, magnification=3, every spy on node/.append style={ultra thin}, width=0.032\textwidth, height=0.022\textwidth}]
            \node [anchor=north west,inner sep=0] (image) at (0,0) {\includegraphics[width=0.244\textwidth]{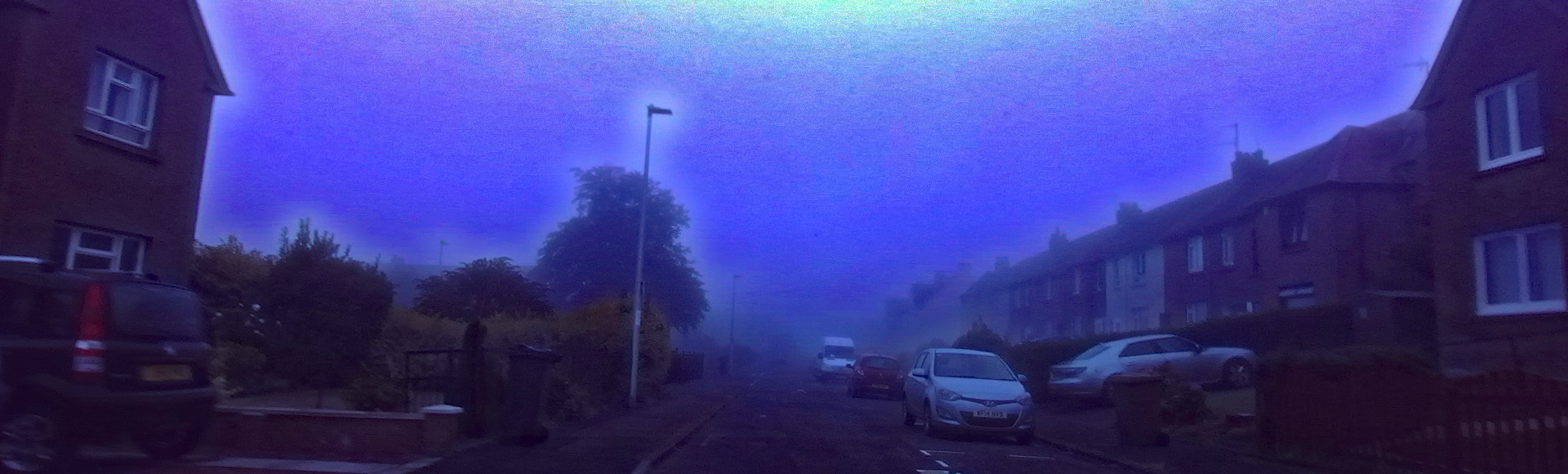}};
            \begin{scope}[x={(image.south east)},y={(image.north west)}]
                \coordinate (ref) at (.0, .0);
                \spy on ($(ref) + (1.09, -0.67)$) in node[below left = 0em and 0em of image.north east, line width=0.1mm];
            \end{scope}
        \end{tikzpicture} &
        \begin{tikzpicture}[node distance = 0cm, anchor=north west, inner sep = 0pt, spy using outlines={rectangle, yellow, magnification=3, every spy on node/.append style={ultra thin}, width=0.020\textwidth, height=0.030\textwidth}]
            \node [anchor=north west,inner sep=0] (image) at (0,0) {\includegraphics[width=0.244\textwidth]{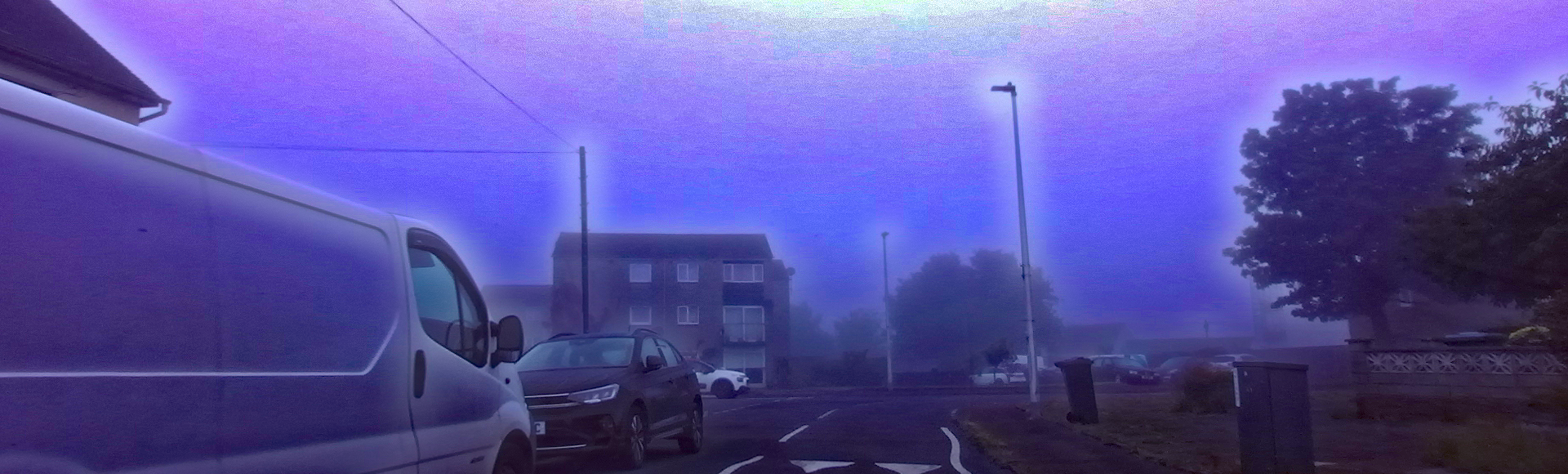}};
            \begin{scope}[x={(image.south east)},y={(image.north west)}]
                \coordinate (ref) at (.0, .0);
                \spy on ($(ref) + (3.48, -0.75)$) in node[below left = 0em and 0em of image.north east, line width=0.1mm];
            \end{scope}
        \end{tikzpicture} \\

        \rotatebox[origin=lc]{90}{\tiny{\quad Berman's \cite{berman2016non}}} &
        \begin{tikzpicture}[node distance = 0cm, anchor=north west, inner sep = 0pt, spy using outlines={rectangle, yellow, magnification=3, every spy on node/.append style={ultra thin}, width=0.024\textwidth, height=0.024\textwidth}]
            \node [anchor=north west,inner sep=0] (image) at (0,0) {\includegraphics[width=0.244\textwidth]{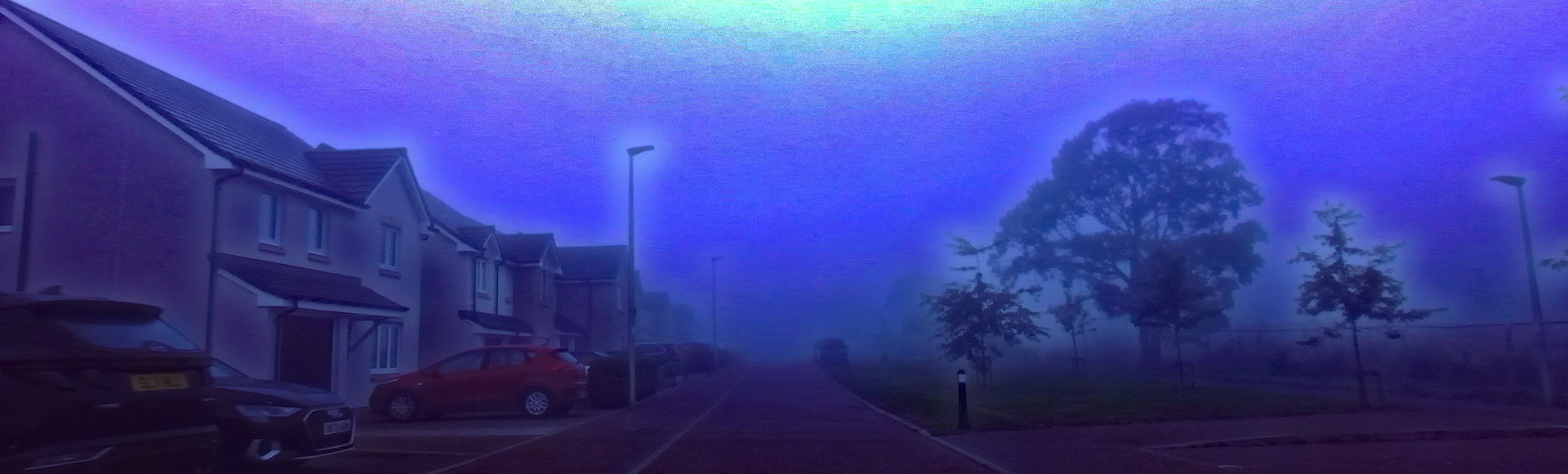}};
            \begin{scope}[x={(image.south east)},y={(image.north west)}]
                \coordinate (ref) at (.0, .0);
                \spy on ($(ref) + (2.28, -0.93)$) in node[below left = 0em and 0em of image.north east, line width=0.1mm];
            \end{scope}
        \end{tikzpicture} &
        \begin{tikzpicture}[node distance = 0cm, anchor=north west, inner sep = 0pt, spy using outlines={rectangle, yellow, magnification=3, every spy on node/.append style={ultra thin}, width=0.008\textwidth, height=0.020\textwidth}]
            \node [anchor=north west,inner sep=0] (image) at (0,0) {\includegraphics[width=0.244\textwidth]{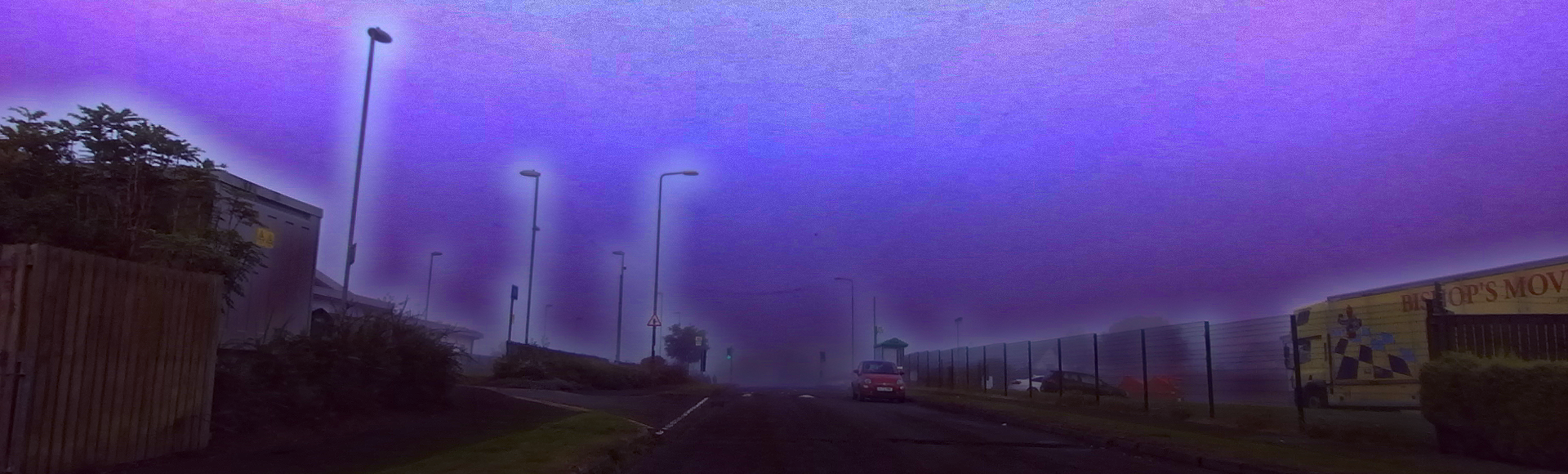}};
            \begin{scope}[x={(image.south east)},y={(image.north west)}]
                \coordinate (ref) at (.0, .0);
                \spy on ($(ref) + (2.04, -0.98)$) in node[below left = 0em and 0em of image.north east, line width=0.1mm];
            \end{scope}
        \end{tikzpicture} &
        \begin{tikzpicture}[node distance = 0cm, anchor=north west, inner sep = 0pt, spy using outlines={rectangle, yellow, magnification=3, every spy on node/.append style={ultra thin}, width=0.032\textwidth, height=0.022\textwidth}]
            \node [anchor=north west,inner sep=0] (image) at (0,0) {\includegraphics[width=0.244\textwidth]{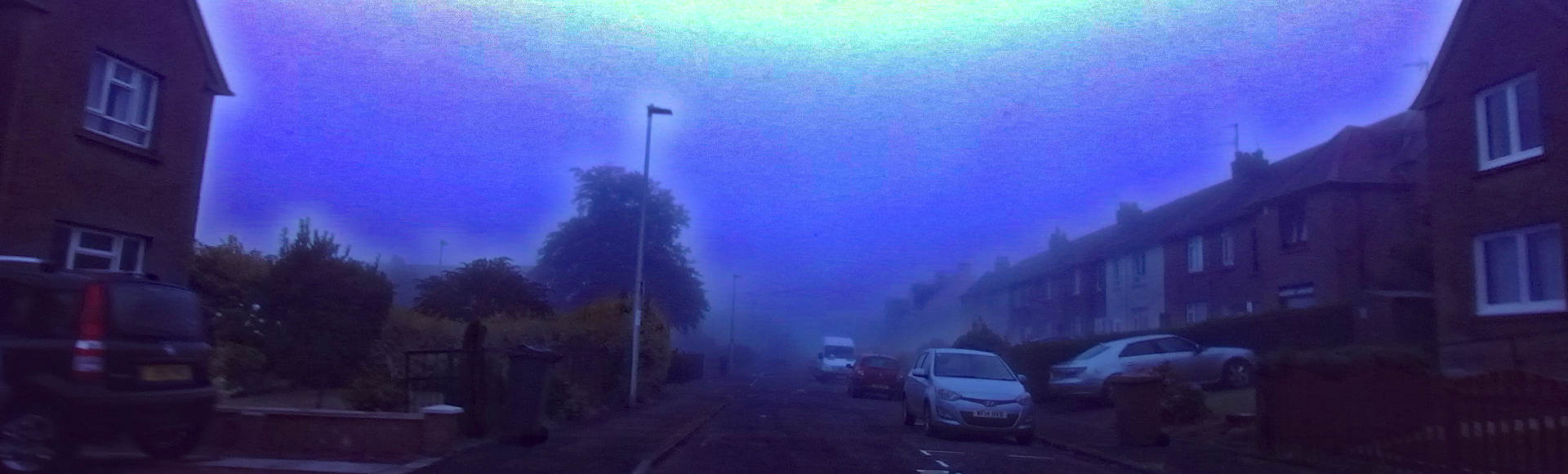}};
            \begin{scope}[x={(image.south east)},y={(image.north west)}]
                \coordinate (ref) at (.0, .0);
                \spy on ($(ref) + (1.09, -0.67)$) in node[below left = 0em and 0em of image.north east, line width=0.1mm];
            \end{scope}
        \end{tikzpicture} &
        \begin{tikzpicture}[node distance = 0cm, anchor=north west, inner sep = 0pt, spy using outlines={rectangle, yellow, magnification=3, every spy on node/.append style={ultra thin}, width=0.020\textwidth, height=0.030\textwidth}]
            \node [anchor=north west,inner sep=0] (image) at (0,0) {\includegraphics[width=0.244\textwidth]{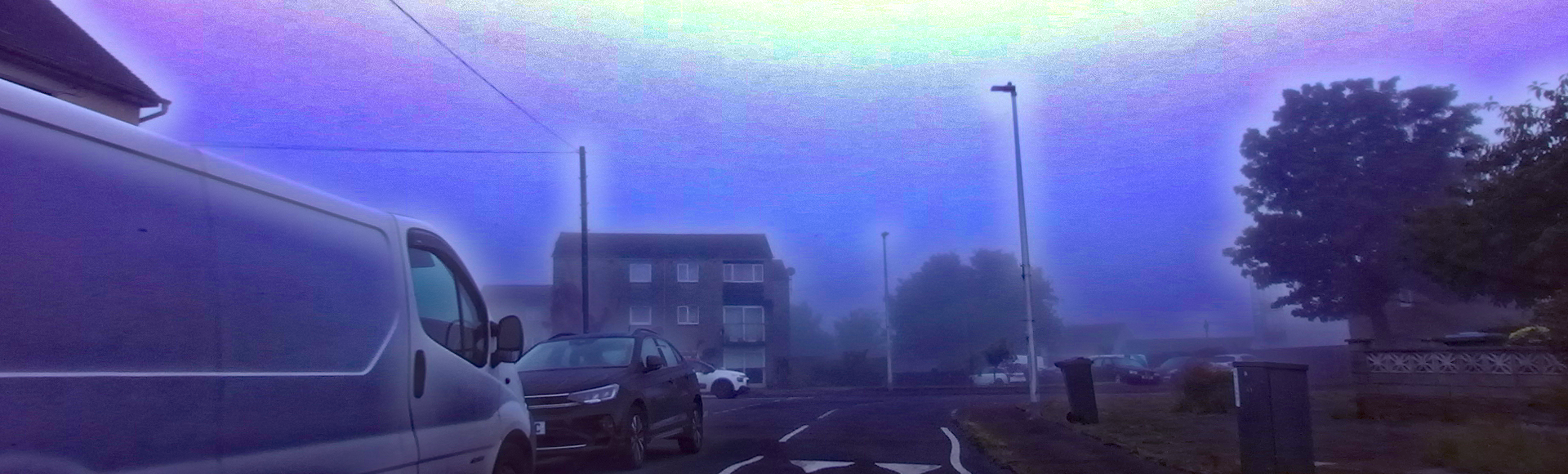}};
            \begin{scope}[x={(image.south east)},y={(image.north west)}]
                \coordinate (ref) at (.0, .0);
                \spy on ($(ref) + (3.48, -0.75)$) in node[below left = 0em and 0em of image.north east, line width=0.1mm];
            \end{scope}
        \end{tikzpicture} \\

        \rotatebox[origin=lc]{90}{\; \; \tiny{Ours}} &
        \begin{tikzpicture}[node distance = 0cm, anchor=north west, inner sep = 0pt, spy using outlines={rectangle, yellow, magnification=3, every spy on node/.append style={ultra thin}, width=0.024\textwidth, height=0.024\textwidth}]
            \node [anchor=north west,inner sep=0] (image) at (0,0) {\includegraphics[width=0.244\textwidth]{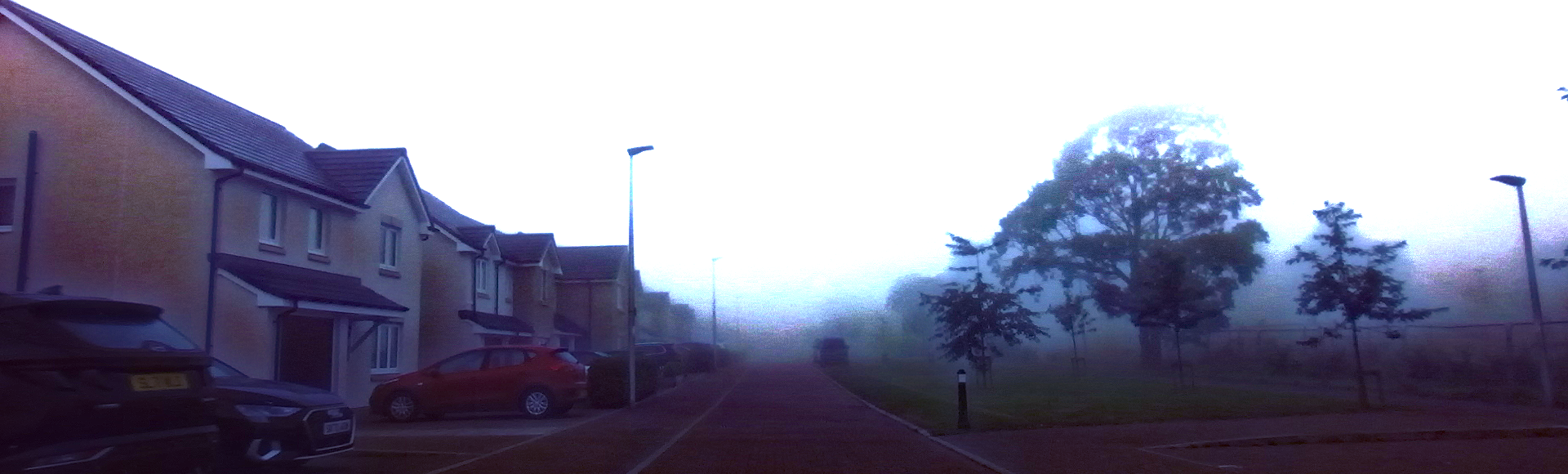}};
            \begin{scope}[x={(image.south east)},y={(image.north west)}]
                \coordinate (ref) at (.0, .0);
                \spy on ($(ref) + (2.28, -0.93)$) in node[below left = 0em and 0em of image.north east, line width=0.1mm];
            \end{scope}
        \end{tikzpicture} &
        \begin{tikzpicture}[node distance = 0cm, anchor=north west, inner sep = 0pt, spy using outlines={rectangle, yellow, magnification=3, every spy on node/.append style={ultra thin}, width=0.008\textwidth, height=0.020\textwidth}]
            \node [anchor=north west,inner sep=0] (image) at (0,0) {\includegraphics[width=0.244\textwidth]{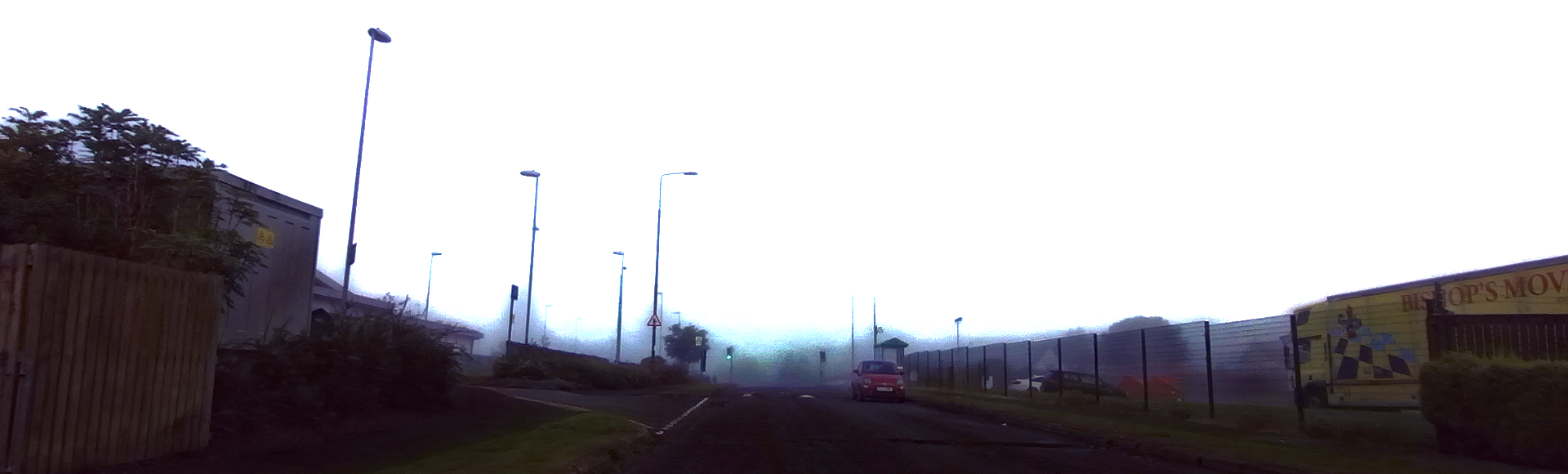}};
            \begin{scope}[x={(image.south east)},y={(image.north west)}]
                \coordinate (ref) at (.0, .0);
                \spy on ($(ref) + (2.04, -0.98)$) in node[below left = 0em and 0em of image.north east, line width=0.1mm];
            \end{scope}
        \end{tikzpicture} &
        \begin{tikzpicture}[node distance = 0cm, anchor=north west, inner sep = 0pt, spy using outlines={rectangle, yellow, magnification=3, every spy on node/.append style={ultra thin}, width=0.032\textwidth, height=0.022\textwidth}]
            \node [anchor=north west,inner sep=0] (image) at (0,0) {\includegraphics[width=0.244\textwidth]{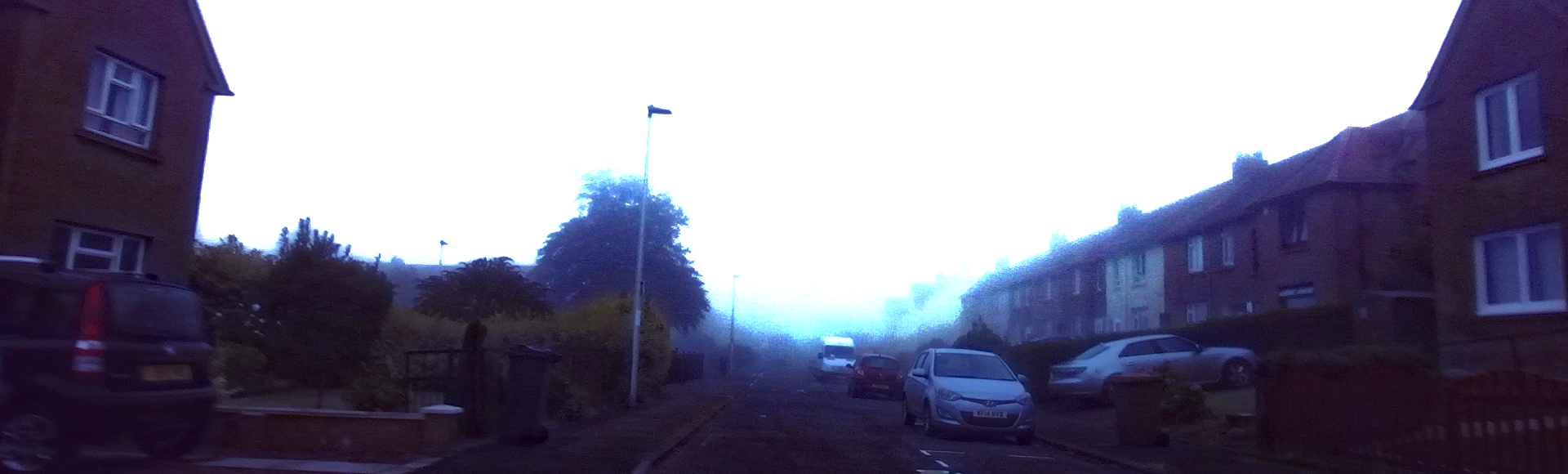}};
            \begin{scope}[x={(image.south east)},y={(image.north west)}]
                \coordinate (ref) at (.0, .0);
                \spy on ($(ref) + (1.09, -0.67)$) in node[below left = 0em and 0em of image.north east, line width=0.1mm];
            \end{scope}
        \end{tikzpicture} &
        \begin{tikzpicture}[node distance = 0cm, anchor=north west, inner sep = 0pt, spy using outlines={rectangle, yellow, magnification=3, every spy on node/.append style={ultra thin}, width=0.020\textwidth, height=0.030\textwidth}]
            \node [anchor=north west,inner sep=0] (image) at (0,0) {\includegraphics[width=0.244\textwidth]{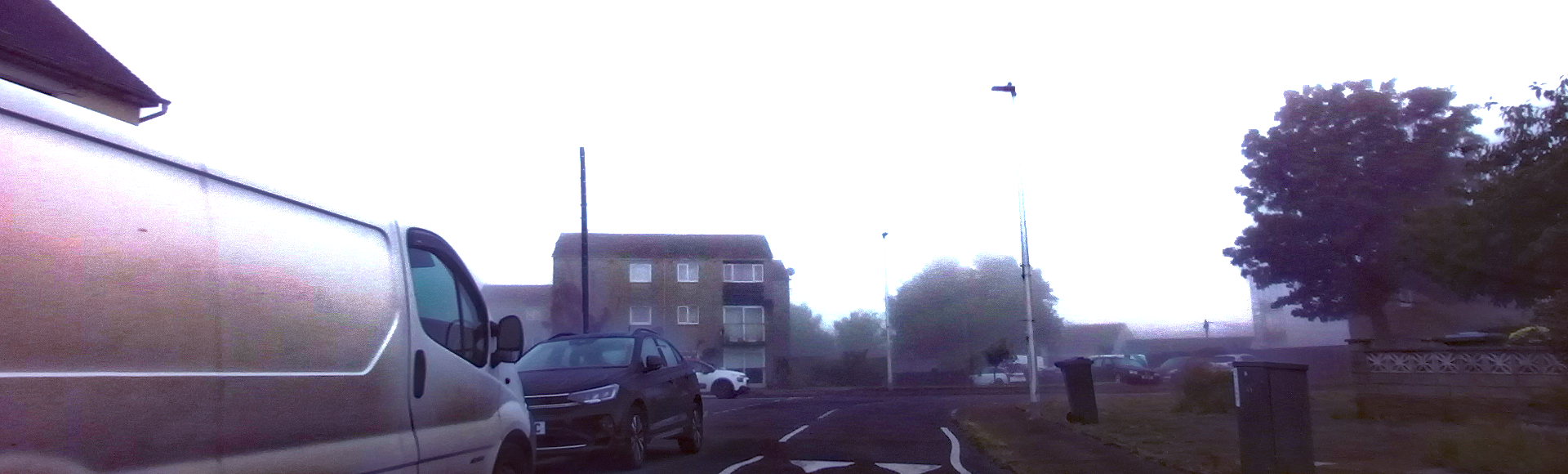}};
            \begin{scope}[x={(image.south east)},y={(image.north west)}]
                \coordinate (ref) at (.0, .0);
                \spy on ($(ref) + (3.48, -0.75)$) in node[below left = 0em and 0em of image.north east, line width=0.1mm];
            \end{scope}
        \end{tikzpicture} \\
    
        \hdashline

        \rotatebox[origin=lc]{90}{\tiny{DehazeFormer \cite{song2023vision}}} &
        \begin{tikzpicture}[node distance = 0cm, anchor=north west, inner sep = 0pt, spy using outlines={rectangle, yellow, magnification=3, every spy on node/.append style={ultra thin}, width=0.024\textwidth, height=0.024\textwidth}]
            \node [anchor=north west,inner sep=0] (image) at (0,0) {\includegraphics[width=0.244\textwidth]{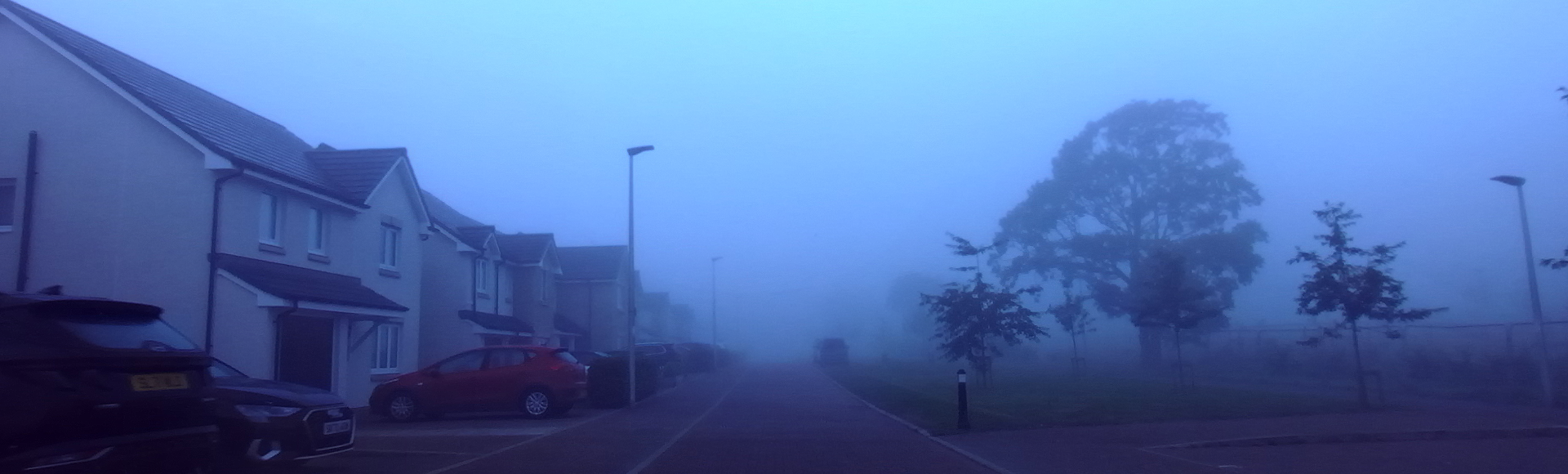}};
            \begin{scope}[x={(image.south east)},y={(image.north west)}]
                \coordinate (ref) at (.0, .0);
                \spy on ($(ref) + (2.28, -0.93)$) in node[below left = 0em and 0em of image.north east, line width=0.1mm];
            \end{scope}
        \end{tikzpicture} &
        \begin{tikzpicture}[node distance = 0cm, anchor=north west, inner sep = 0pt, spy using outlines={rectangle, yellow, magnification=3, every spy on node/.append style={ultra thin}, width=0.008\textwidth, height=0.020\textwidth}]
            \node [anchor=north west,inner sep=0] (image) at (0,0) {\includegraphics[width=0.244\textwidth]{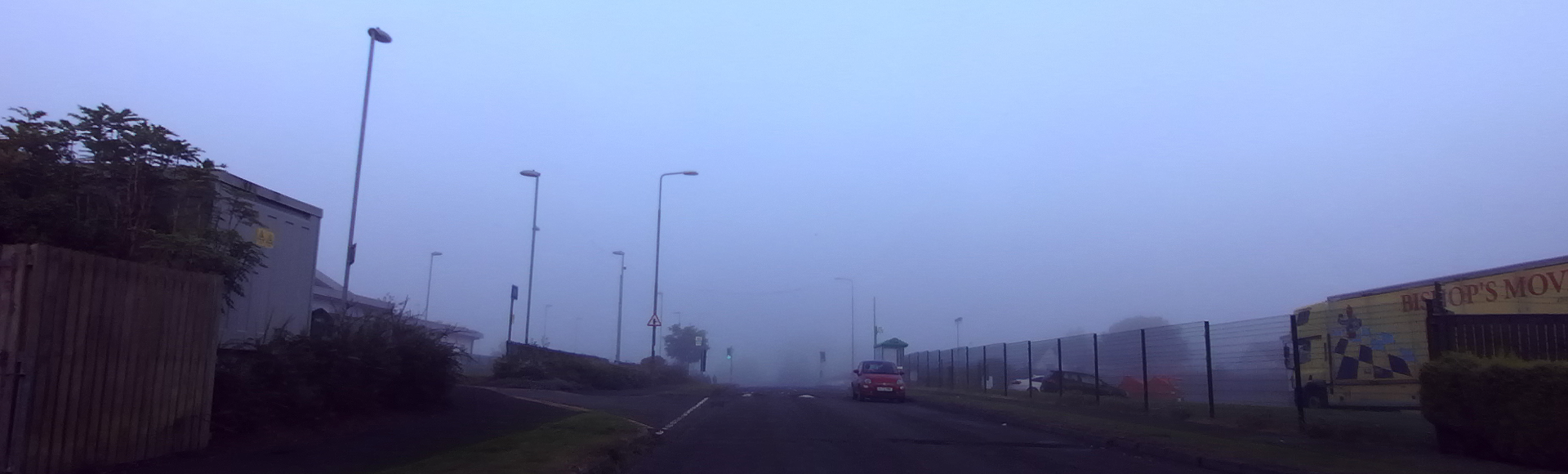}};
            \begin{scope}[x={(image.south east)},y={(image.north west)}]
                \coordinate (ref) at (.0, .0);
                \spy on ($(ref) + (2.04, -0.98)$) in node[below left = 0em and 0em of image.north east, line width=0.1mm];
            \end{scope}
        \end{tikzpicture} &
        \begin{tikzpicture}[node distance = 0cm, anchor=north west, inner sep = 0pt, spy using outlines={rectangle, yellow, magnification=3, every spy on node/.append style={ultra thin}, width=0.032\textwidth, height=0.022\textwidth}]
            \node [anchor=north west,inner sep=0] (image) at (0,0) {\includegraphics[width=0.244\textwidth]{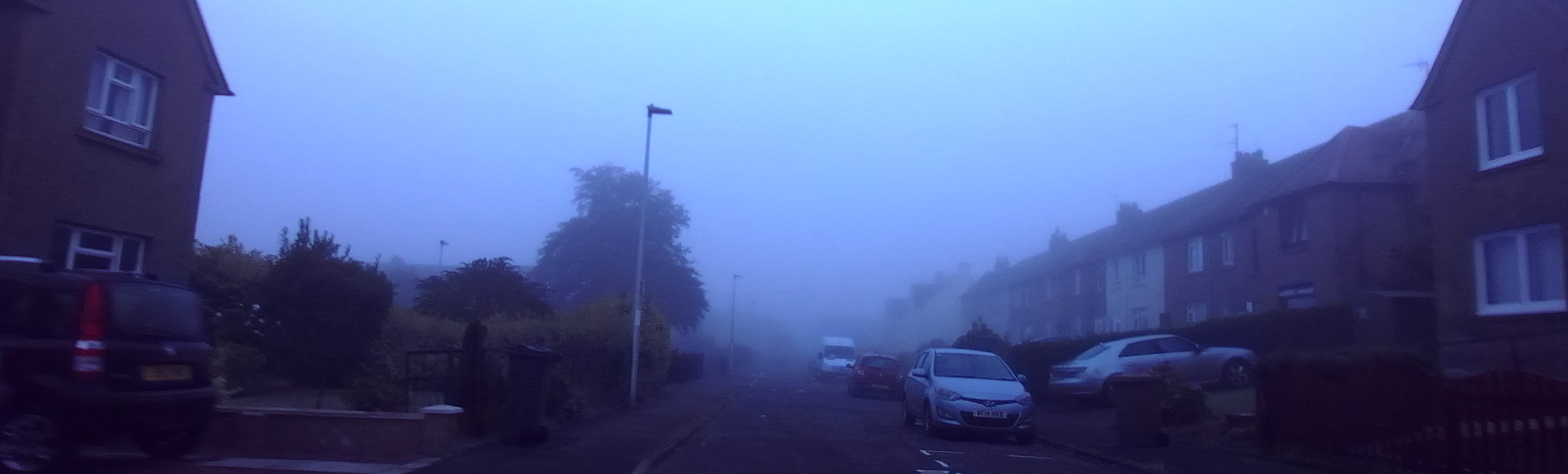}};
            \begin{scope}[x={(image.south east)},y={(image.north west)}]
                \coordinate (ref) at (.0, .0);
                \spy on ($(ref) + (1.09, -0.67)$) in node[below left = 0em and 0em of image.north east, line width=0.1mm];
            \end{scope}
        \end{tikzpicture} &
        \begin{tikzpicture}[node distance = 0cm, anchor=north west, inner sep = 0pt, spy using outlines={rectangle, yellow, magnification=3, every spy on node/.append style={ultra thin}, width=0.020\textwidth, height=0.030\textwidth}]
            \node [anchor=north west,inner sep=0] (image) at (0,0) {\includegraphics[width=0.244\textwidth]{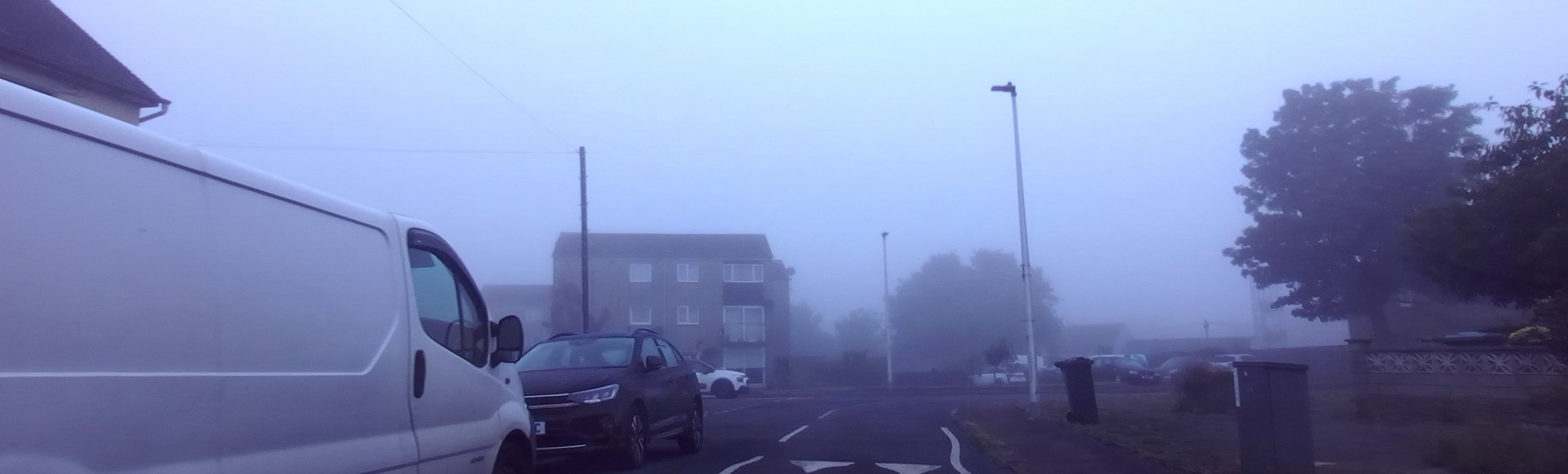}};
            \begin{scope}[x={(image.south east)},y={(image.north west)}]
                \coordinate (ref) at (.0, .0);
                \spy on ($(ref) + (3.48, -0.75)$) in node[below left = 0em and 0em of image.north east, line width=0.1mm];
            \end{scope}
        \end{tikzpicture} \\

        \rotatebox[origin=lc]{90}{\tiny{\quad C2PNet \cite{zheng2023curricular}}} &
        \begin{tikzpicture}[node distance = 0cm, anchor=north west, inner sep = 0pt, spy using outlines={rectangle, yellow, magnification=3, every spy on node/.append style={ultra thin}, width=0.024\textwidth, height=0.024\textwidth}]
            \node [anchor=north west,inner sep=0] (image) at (0,0) {\includegraphics[width=0.244\textwidth]{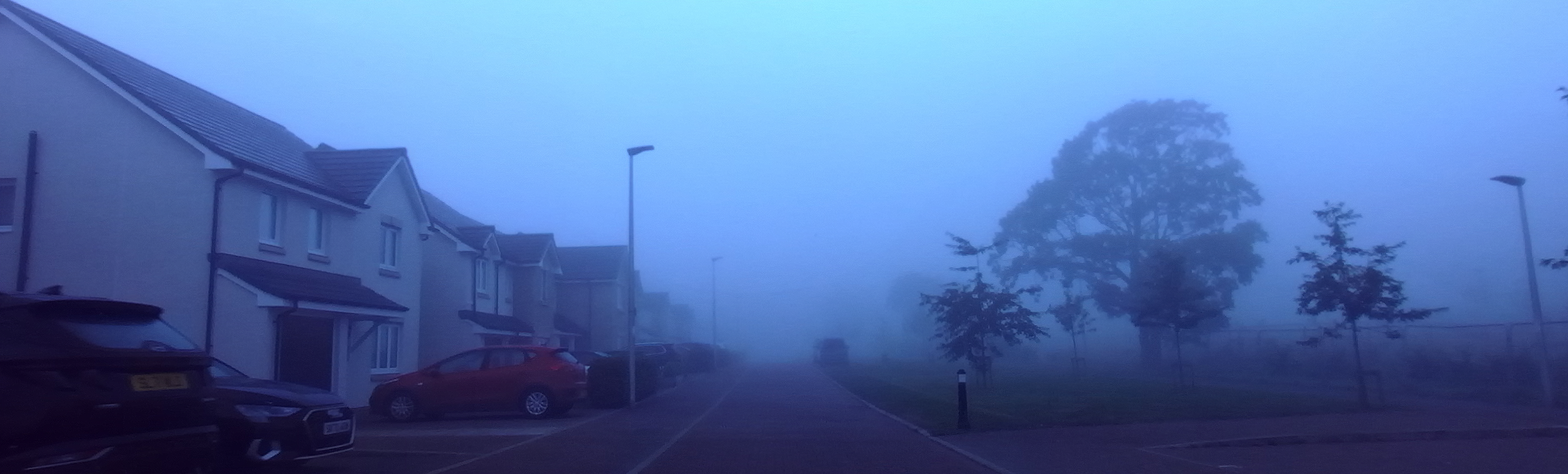}};
            \begin{scope}[x={(image.south east)},y={(image.north west)}]
                \coordinate (ref) at (.0, .0);
                \spy on ($(ref) + (2.28, -0.93)$) in node[below left = 0em and 0em of image.north east, line width=0.1mm];
            \end{scope}
        \end{tikzpicture} &
        \begin{tikzpicture}[node distance = 0cm, anchor=north west, inner sep = 0pt, spy using outlines={rectangle, yellow, magnification=3, every spy on node/.append style={ultra thin}, width=0.008\textwidth, height=0.020\textwidth}]
            \node [anchor=north west,inner sep=0] (image) at (0,0) {\includegraphics[width=0.244\textwidth]{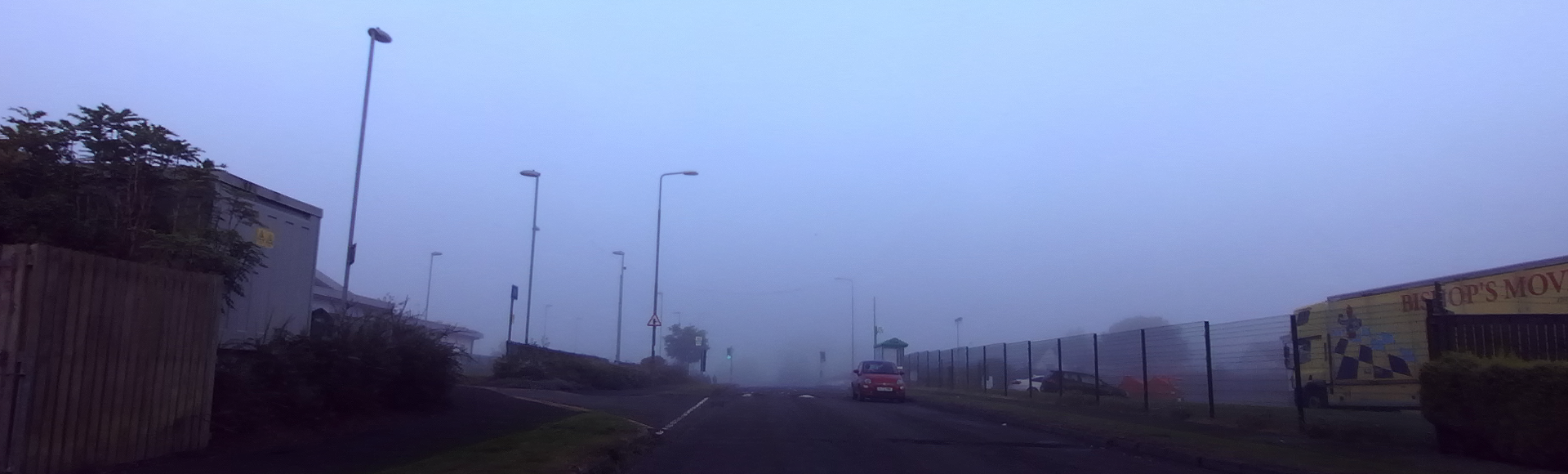}};
            \begin{scope}[x={(image.south east)},y={(image.north west)}]
                \coordinate (ref) at (.0, .0);
                \spy on ($(ref) + (2.04, -0.98)$) in node[below left = 0em and 0em of image.north east, line width=0.1mm];
            \end{scope}
        \end{tikzpicture} &
        \begin{tikzpicture}[node distance = 0cm, anchor=north west, inner sep = 0pt, spy using outlines={rectangle, yellow, magnification=3, every spy on node/.append style={ultra thin}, width=0.032\textwidth, height=0.022\textwidth}]
            \node [anchor=north west,inner sep=0] (image) at (0,0) {\includegraphics[width=0.244\textwidth]{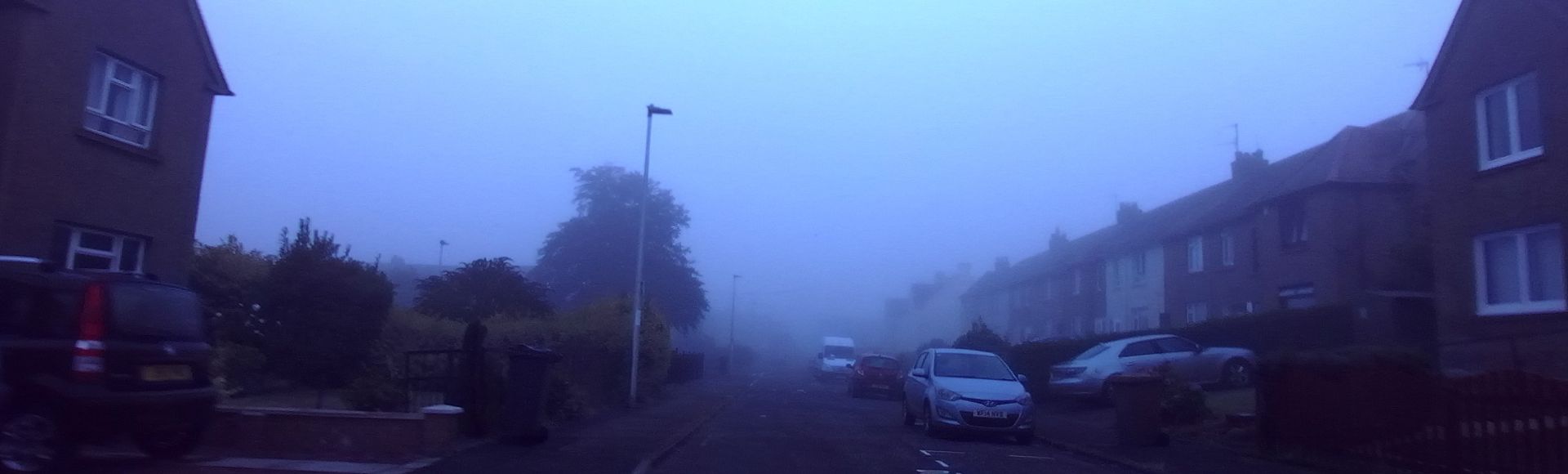}};
            \begin{scope}[x={(image.south east)},y={(image.north west)}]
                \coordinate (ref) at (.0, .0);
                \spy on ($(ref) + (1.09, -0.67)$) in node[below left = 0em and 0em of image.north east, line width=0.1mm];
            \end{scope}
        \end{tikzpicture} &
        \begin{tikzpicture}[node distance = 0cm, anchor=north west, inner sep = 0pt, spy using outlines={rectangle, yellow, magnification=3, every spy on node/.append style={ultra thin}, width=0.020\textwidth, height=0.030\textwidth}]
            \node [anchor=north west,inner sep=0] (image) at (0,0) {\includegraphics[width=0.244\textwidth]{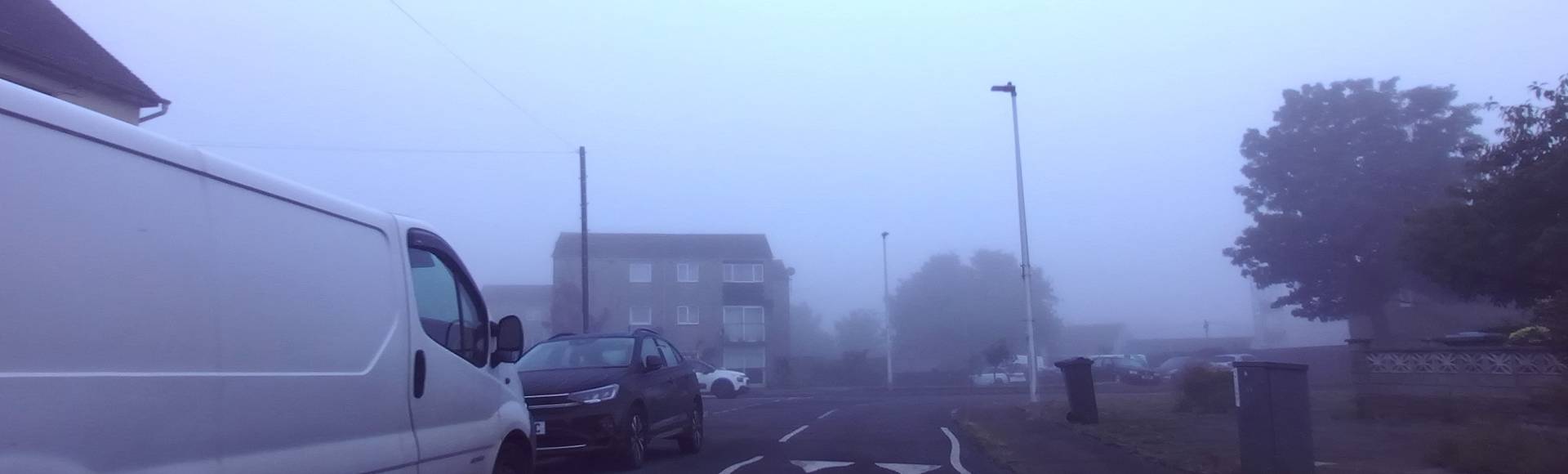}};
            \begin{scope}[x={(image.south east)},y={(image.north west)}]
                \coordinate (ref) at (.0, .0);
                \spy on ($(ref) + (3.48, -0.75)$) in node[below left = 0em and 0em of image.north east, line width=0.1mm];
            \end{scope}
        \end{tikzpicture} \\

        \hdashline
        \addlinespace[0.4ex]

        \rotatebox[origin=lc]{90}{\; \, \ \tiny{GT}} &
        \includegraphics[width=0.244\textwidth]{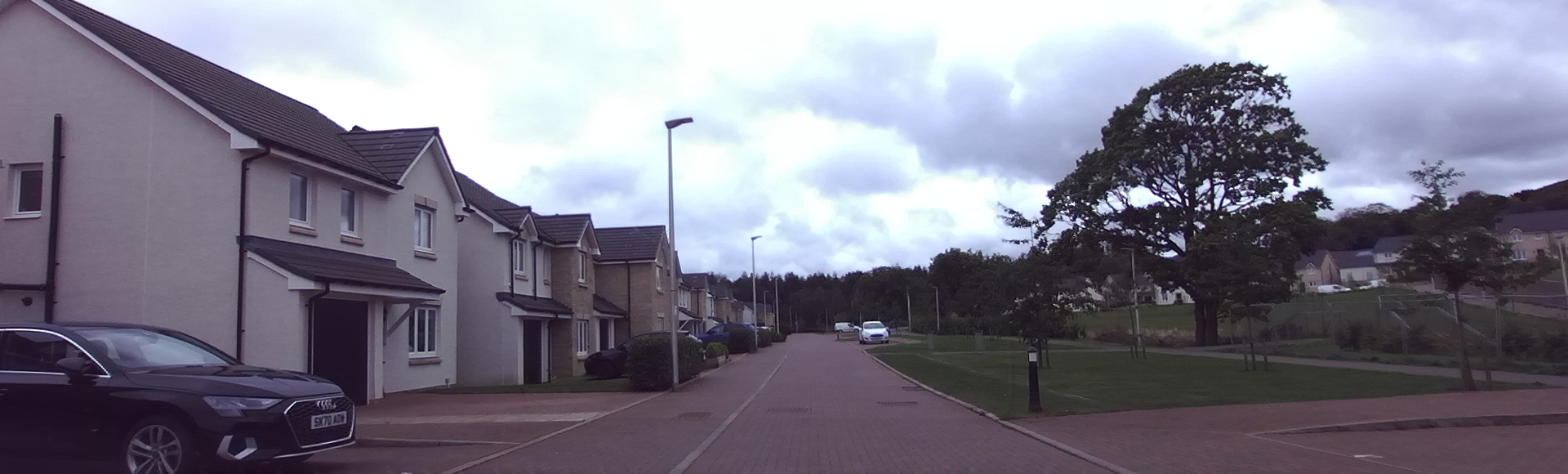} &
        \includegraphics[width=0.244\textwidth]{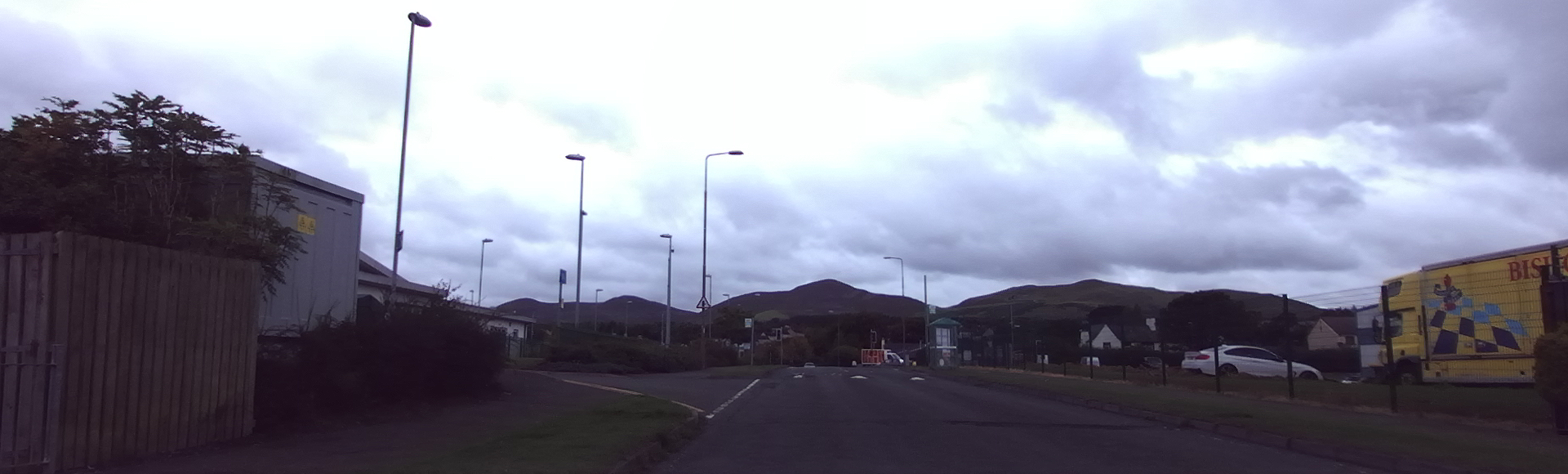} &
        \includegraphics[width=0.244\textwidth]{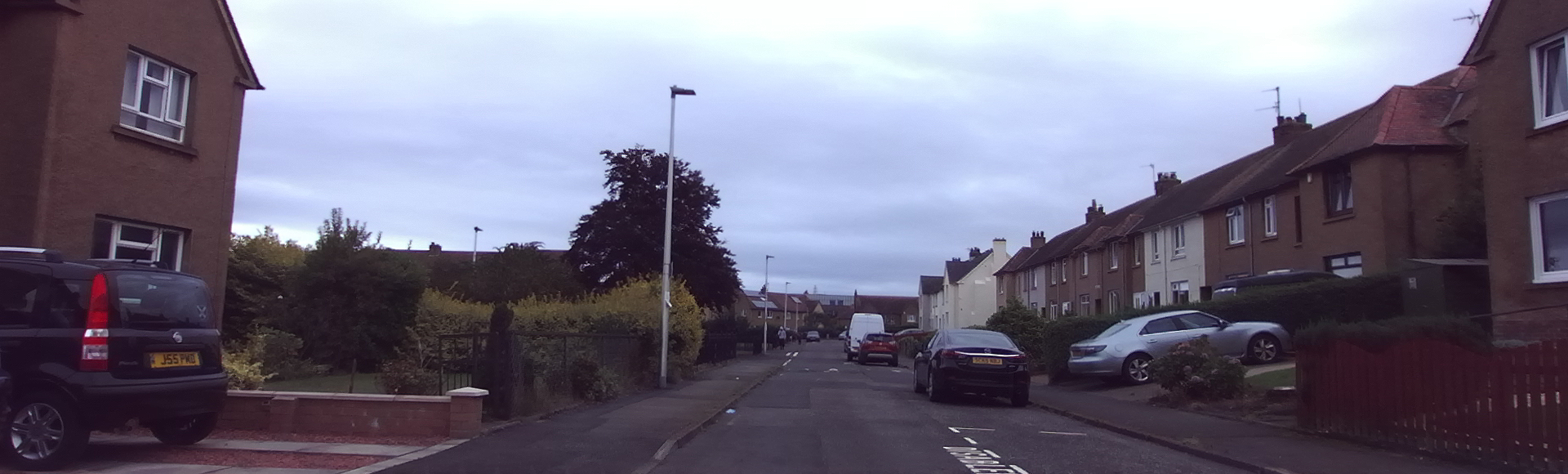} &
        \includegraphics[width=0.244\textwidth]{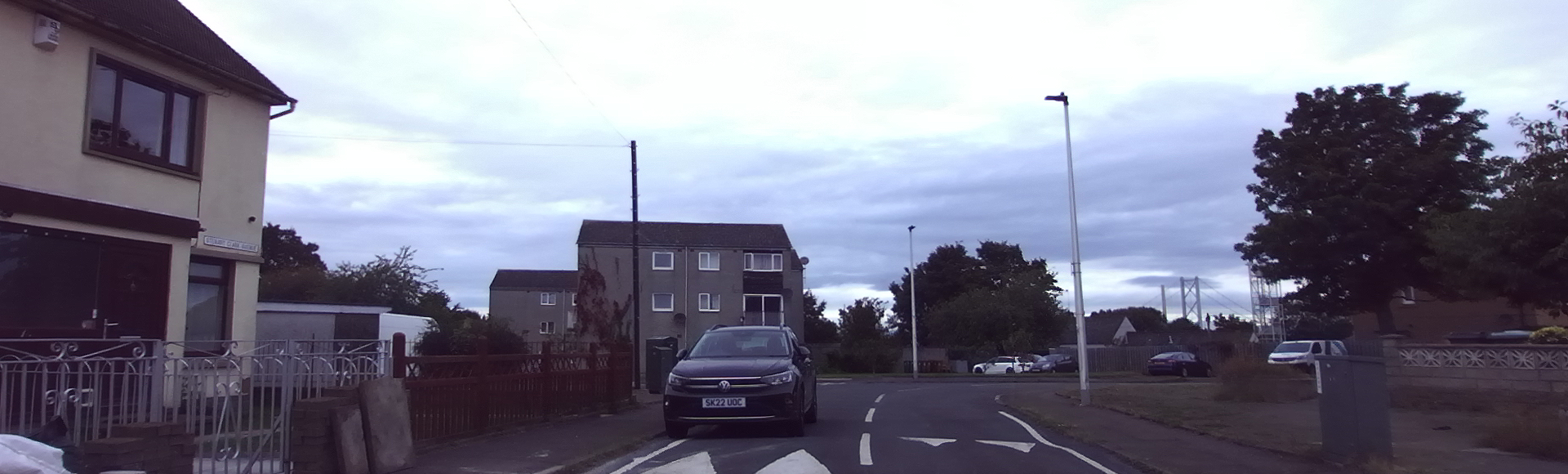}
        
    \end{tabular}

    \setlength{\abovecaptionskip}{2.5pt}
    \caption{Evaluating the atmospheric light estimated by various methods (top four rows) on SDIRF by using it to perform defogging following \cite{he2010single}.
    The input foggy images are shown in the corresponding columns of \RefFig \ref{fig:compare_atmos}.
    The corresponding clear images recorded in overcast weather are shown in the last row to serve as pseudo-ground truth.
    We observe:
    a) The defogged images which use the atmospheric light estimated by our method appear to be more accurate in colour compared to others with minimal visual artefacts;
    b) Using the atmospheric light estimated by our method, fog on distant objects seems to be better removed (see the close-up of yellow squares).
    In addition, we investigate how two state-of-the-art, end-to-end deep learning-based defogging methods, DehazeFormer \cite{song2023vision} and C2PNet \cite{zheng2023curricular}, perform on the same foggy images (in the intensity domain, as their networks were trained on intensity images).
    The results are shown in the middle two rows between the dashed lines.
    The key observation is that for both methods their defogging effect is barely visible.
    }
    \setlength{\belowcaptionskip}{-20pt}
    \label{fig:compare_defogged_with_different_atmos}
\end{figure*}

For real foggy images, atmospheric light is no longer monochrome.
Therefore, we apply our method to each colour channel.

Firstly, we visually compare the atmospheric light estimated by various methods.
After obtaining the estimate of $L_{\infty}$ of each colour channel by each method, for visualisation purposes we map it back to pixel intensity $A$ by applying $g^{-1}$ so that its colour can be illustrated.
Sample results are shown in \RefFig \ref{fig:compare_atmos}.
The key observation is that the atmospheric light estimated by our method is the closest to the colour of the horizon, i.e., the most fog-opaque region in a foggy image.
We also observe that our method is more robust to changes in the atmospheric light than competing ones.

Next, we take a step further by visually comparing the defogging results using $L_{\infty}$ estimated by various methods.
To facilitate a fair comparison between all methods such that the only difference is $L_{\infty}$, we follow \cite{he2010single} to estimate $t$ and perform defogging.
Again, for visualisation purposes, we apply $g^{-1}$ to the defogging results, and the final defogged images are shown in the top four rows in \RefFig \ref{fig:compare_defogged_with_different_atmos}.
The clear images recorded in overcast weather are shown in the last row as pseudo-ground truth.
The key observation is that using the atmospheric light estimated by our method yields defogged images that are perceptually superior to those produced by other methods.

In addition, we evaluate two state-of-the-art, end-to-end deep learning-based defogging methods, DehazeFormer \cite{song2023vision} and C2PNet \cite{zheng2023curricular}, on the same foggy images (in the intensity domain, as their networks were trained on intensity images).
The results are shown in the middle two rows between the dashed lines in \RefFig \ref{fig:compare_defogged_with_different_atmos}.
Compared with the foggy images in \RefFig \ref{fig:compare_atmos}, their defogging effect is barely visible.

See our supplementary material for more results, including comparisons of the estimated atmospheric light and the defogged images obtained in the radiance and intensity domains.

\subsection{Additional Experiments}
We report two additional experiments to demonstrate
a) our results on $\beta$'s wavelength dependence align with what was reported from previous physics experiments;
b) the non-linearity introduced by gamma correction cannot be ignored when estimating $\beta$. 

\subsubsection{Scattering Coefficient's Wavelength Dependence}
In this experiment, we apply our method to each colour channel (RGB) as well as to the grayscale image independently, and investigate how $\beta$ varies with wavelength.

To this end, we examine all $\beta$ estimates from a total of $34457$ frames evaluated on all foggy sequences of SDIRF.
We categorise the $\beta$ values obtained from each frame into the following three cases:
a) $\beta_{\text{R}} > \beta_{\text{G}} > \beta_{\text{B}}$ (i.e., $\beta$ strictly increases with wavelength);
b) $\beta_{\text{R}} < \beta_{\text{G}} < \beta_{\text{B}}$ (i.e., $\beta$ strictly decreases with wavelength);
c) Otherwise.
We investigate how these three cases are distributed as the grayscale scattering coefficient $\beta_{\text{grayscale}}$, which measures the mean visibility, changes for both Li's modified method and our method.
The results are are illustrated in \RefFig \ref{fig:beta_wavelength_dependence} as normalised histograms [(a) and (c)], and as cumulative distribution curves [(b) and (d)].

The key observation is that the results of our method align with \cite[Figure 6.12]{mccartney1976optics}, which shows that at lower visibility (i.e., a larger $\beta$) the relative attenuation of different colour channels tends to increase with wavelength (i.e., $\beta_{\text{R}} > \beta_{\text{G}} > \beta_{\text{B}}$), whereas at higher visibility (i.e., a smaller $\beta$) it tends to decrease with wavelength (i.e., $\beta_{\text{R}} < \beta_{\text{G}} < \beta_{\text{B}}$).
In contrast, the results of Li's modified method do not show such trends.

\subsubsection{Gamma Correction}    \label{subsubsec:gamma_correction}
\begin{figure}[!t]
    \centering
    \includegraphics[trim={0.41cm 0.38cm 0.4cm .4cm},clip,width=\linewidth]{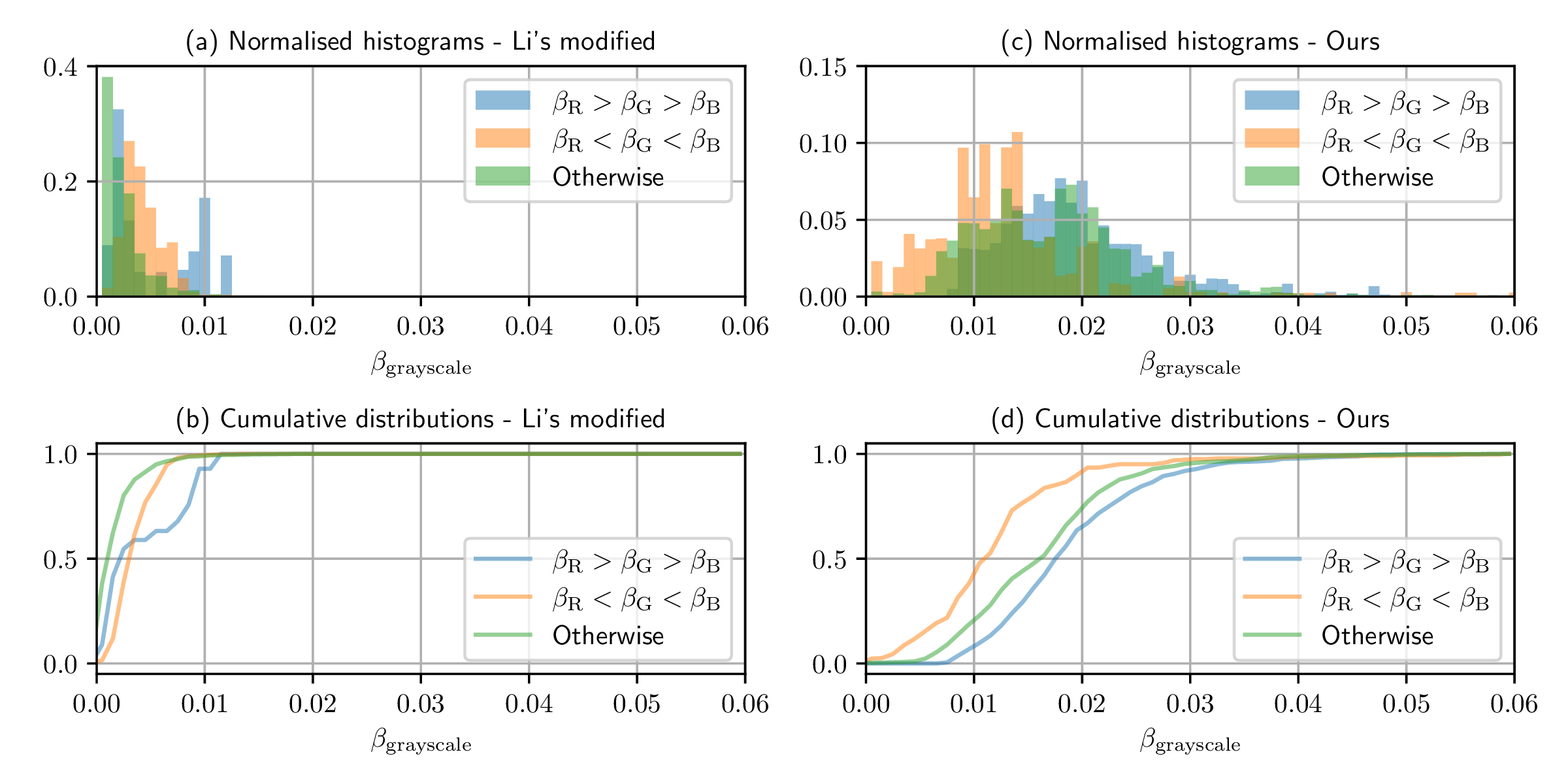}
    \setlength{\abovecaptionskip}{-15pt}
    \caption{Investigating $\beta$'s wavelength dependence on all foggy sequences in SDIRF.
    The results of Li's modified method are shown in (a) and (b), and the results of our method are shown in (c) and (d).
    From (c) and (d) we observe that the case $\beta_{\text{R}} > \beta_{\text{G}} > \beta_{\text{B}}$ tends to happen at a larger $\beta$ value, whereas the case $\beta_{\text{R}} < \beta_{\text{G}} < \beta_{\text{B}}$ tends to happen at a smaller $\beta$ value.
    The remaining cases tend to happen at intermediate $\beta$ values.
    The above observations are in line with \cite[Figure 6.12]{mccartney1976optics}.
    However, they cannot be made from (a) or (b). 
    }
    \setlength{\belowcaptionskip}{-10pt}
    \label{fig:beta_wavelength_dependence}
\end{figure}

We investigate the effect of the non-linearity caused by gamma correction on the estimate of $\beta$.

\begin{figure}[!t]
    \centering
    \includegraphics[trim={0.38cm 0.38cm 0.4cm .4cm},clip,width=\linewidth]{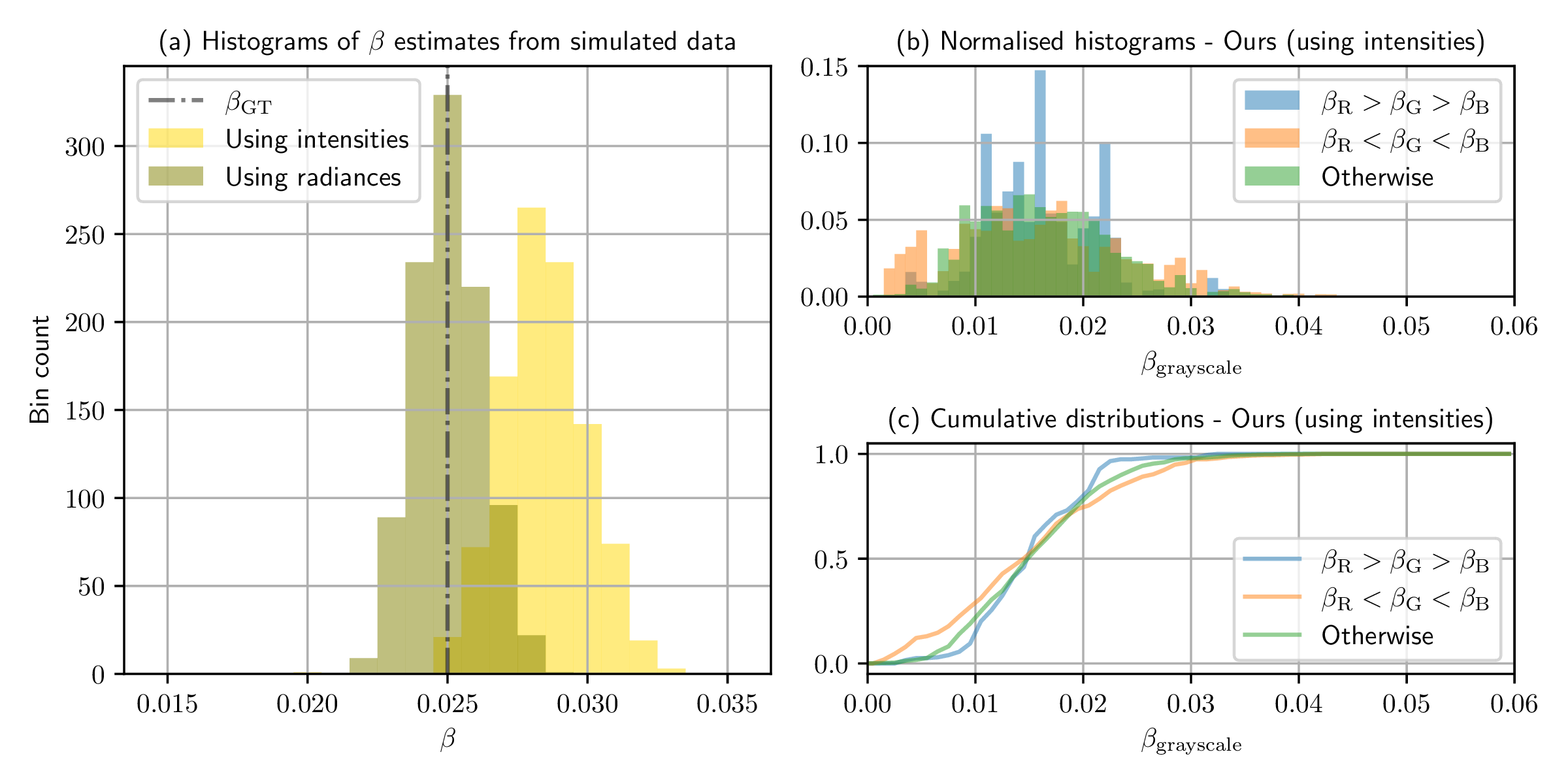}
    \setlength{\abovecaptionskip}{-15pt}
    \caption{
    (a) Investigating how the gamma correction affects the estimate of $\beta$ using simulated data.
    We observe that using intensities rather than radiances overestimates $\beta$.
    Comparing (b) and (c) with their counterparts [\RefFigs \ref{fig:beta_wavelength_dependence}(c) and \ref{fig:beta_wavelength_dependence}(d)], we can see that using intensities rather than radiances makes both the histograms and the cumulative distribution curves significantly overlap.
    }
    \setlength{\belowcaptionskip}{-10pt}
    \label{fig:beta_affected_by_gamma_correction}
\end{figure}

Firstly, we conduct the following experiment using simulated data.
We generate clean data consisting of the radiances of a number of landmarks observed from a range of distances according to \eqref{eq:asm_radiance} [i.e., drawing samples from the dotted lines shown in \RefFig \ref{fig:method}(b)] with ground truth $\beta_{\text{GT}} = 0.025$, which is then corrupted with random Gaussian noise.
We then apply $g^{-1}$ (with $\gamma>1$ in accordance with our photometric calibration results) to convert the radiance data to intensity data.
Next, we use our proposed method to estimate two $\beta$ values, one from the radiance data and one from the intensity data.
The experiment is repeated $1000$ times.
Due to random noise, each instance leads to slightly different values for $\beta$.
We plot their histogram in \RefFig \ref{fig:beta_affected_by_gamma_correction}(a).
We observe that using intensities we tend to overestimate $\beta$, whereas using radiances the estimates seem to be unbiased.
In fact, estimating $\beta$ using intensity \emph{always} yields estimates larger than using radiance.
Our experiment also reveals that a) the direction of the bias depends on whether $\gamma>1$ or $\gamma<1$; b) the amount of the bias increases as $\gamma$ deviates from 1.
See our supplementary material for more details of the experiment and the results.

Finally, in \RefFigs \ref{fig:beta_affected_by_gamma_correction}(b) and \ref{fig:beta_affected_by_gamma_correction}(c) we show the counterparts of \RefFigs \ref{fig:beta_wavelength_dependence}(c) and \ref{fig:beta_wavelength_dependence}(d) but using intensities instead of radiances.
We observe that this time the histograms and the cumulative distribution curves significantly overlap, which adds to the evidence that the atmospheric scattering model should be applied to radiances rather than to intensities.

\section{Conclusion}    \label{sec:conclusion}
We presented an optimisation-based method for estimating the parameters of fog.
While prior methods adopt a sequential estimation strategy that is prone to error propagation, our method simultaneously estimates the parameters by solving a minimisation problem.
Extensive experiments show that our method outperforms prior ones on synthetic data both qualitatively and quantitatively, and on real data qualitatively from various aspects.
Our method has the potential to be plugged into an existing feature-based visual SLAM/odometry system as an add-on module for its deployment in fog.
In addition, we have introduced SDIRF, a dataset consisting of high-quality, consecutive stereo foggy images of real road scenes under a variety of visibility conditions.
SDIRF also provides calibrated photometric parameters, which makes it photometrically ready to apply the atmospheric scattering model, as well as counterpart clear images taken in overcast weather of the same routes, which will be useful for companion work in image defogging and depth reconstruction.
All of the above features together make SDIRF a first-of-its-kind dataset for the study of visual perception for autonomous driving in fog.

In the future, we will investigate how to improve the resilience of our method when the underlying visual SLAM system struggles to generate accurate distance and/or intensity (hence radiance) information, which is a limitation of our current method.
Our experimental results in \RefSec \ref{subsubsec:synthetic_quantitative_results} suggest that these situations arise in countryside scenes with very sparse features or when the ego-vehicle is surrounded by other vehicles moving at similar speed.
To this end, we will consider the following two approaches:
a) to more tightly couple our method with a visual SLAM system by jointly optimising the fog parameters, the camera's poses, and the landmark's 3D positions;
b) to integrate our method into a visual-inertial SLAM system (e.g., ORB-SLAM3 \cite{campos2021orb}) that is inherently more robust in the presence of fog.

\section*{Acknowledgements}
The authors thank Aongus McCarthy for offering help and equipment for the photometric calibration.
This work is supported in part by U.K.’s Engineering and Physical Sciences Research Council (EPSRC) under Grant EP/S023208/1.

\bibliographystyle{IEEEtran}
\bibliography{references}

\newpage

\begin{IEEEbiography}[{\includegraphics[width=1in,height=1.25in,clip,keepaspectratio]{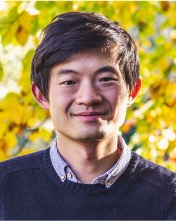}}]{Yining Ding}
received the B.Eng. (Hons.) degree in electronics and electrical engineering from the University of Edinburgh, Edinburgh, U.K., in 2013, the M.Sc. degree in communications and signal processing from Imperial College London, London, U.K., in 2014, and the Ph.D. degree in robotics and autonomous systems from the Edinburgh Centre for Robotics, Edinburgh, U.K., in 2025.
He worked as a design engineer on ultrasound signal processing for industrial metrology applications at Renishaw plc, U.K., between 2014 and 2020.
His research focuses on robust visual perception in fog, including fog parameter estimation, depth reconstruction, and image defogging.
\end{IEEEbiography}

\vspace{11pt}

\begin{IEEEbiography}[{\includegraphics[width=1in,height=1.25in,clip,keepaspectratio]{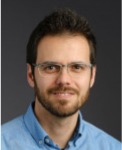}}]{João F. C. Mota}
received the M.Sc. and Ph.D. degrees in electrical and computer engineering from the Technical University of Lisbon, Lisbon, Portugal, in 2008 and 2013, respectively, and the Ph.D. degree in electrical and computer engineering from Carnegie Mellon University, Pittsburgh, PA, USA, in 2013.
He is currently an Assistant Professor of Signal and Image Processing with Heriot-Watt University, Edinburgh, U.K.
His research interests include theoretical and practical aspects of high-dimensional data processing, inverse problems, optimization theory, machine learning, data science, and distributed information processing and control.
Dr. Mota was a recipient of the 2015 IEEE Signal Processing Society Young Author Best Paper Award and is currently Associate Editor for IEEE Transactions on Signal Processing.
\end{IEEEbiography}

\vspace{11pt}

\begin{IEEEbiography}[{\includegraphics[width=1in,height=1.25in,clip,keepaspectratio]{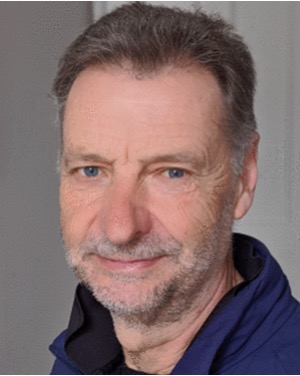}}]{Andrew Michael Wallace}
received the B.Sc. and Ph.D. degrees from the University of Edinburgh, Edinburgh, U.K., in 1972 and 1975, respectively.
He was an Emeritus Professor of Signal and Image Processing with Heriot-Watt University, Edinburgh, U.K.
His research interests included LiDAR and 3D vision, image and signal processing, and accelerated computing.
He had authored or coauthored extensively and had secured funding from EPSRC, the EU and other sponsors.
He was a Chartered Engineer and a fellow of the Institute of Engineering Technology.
\end{IEEEbiography}

\vspace{11pt}

\begin{IEEEbiography}[{\includegraphics[width=1in,height=1.25in,clip,keepaspectratio]{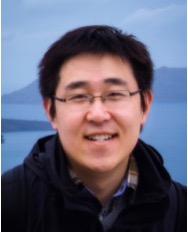}}]{Sen Wang}
received the Ph.D. degree in robotics from the University of Essex, Colchester, U.K., in 2015.
He is currently an Associate Professor with the Sense Robotics Lab, Imperial College London, London, U.K.
His research interests include robot perception and autonomy using probabilistic and learning approaches, especially autonomous navigation, robotic vision, SLAM, and robot learning. He has served as an Associate Editor for IEEE Transactions on Robotics, IEEE Transactions on Automation Science and Engineering and IEEE Robotics and Automation Letters.
\end{IEEEbiography}

\vfill

\end{document}